\documentclass[preprint, 11pt, authoryear, sort]{elsarticle}

\usepackage[utf8]{inputenc}
\usepackage{amsmath,amssymb,amsfonts}
\usepackage{algorithm,algpseudocode}
\usepackage{graphicx}
\usepackage{multirow}
\usepackage{textcomp}
\usepackage{stfloats}
\usepackage{setspace}
\usepackage{cite}
\usepackage{booktabs}
\usepackage{url}
\usepackage[colorlinks=true,urlcolor=blue,linkcolor=blue,citecolor=blue]{hyperref}
\usepackage{orcidlink}
\usepackage[a4paper,
            bindingoffset=0.2in,
            left=0.5in,
            right=0.5in,
            top=0.5in,
            bottom=0.5in,
            footskip=.25in]{geometry}
\newcommand{\startparent}{%
  \setcounter{equation}{0}%
  \xdef\theparentequation{\arabic{parentequation}}%
}
\DeclareMathOperator*{\argmax}{arg\,max}

\journal{European Journal of Operational Research}

\begin{document}

\begin{frontmatter}


\title{A Two-Stage Reactive Auction Framework for the Multi-Depot Rural Postman Problem with Dynamic Vehicle Failures}

\author[1]{Eashwar Sathyamurthy}
\ead{eashwar@umd.edu}

\affiliation[1]{organization={Department of Mechanical Engineering, University of Maryland},
                city={College Park}, 
                state={Maryland},
                postcode={20740}, 
                country={USA}}

\author[2]{Jeffrey W. Herrmann}

\author[1]{Shapour Azarm}


\affiliation[2]{organization={Department of Mechanical Engineering, 
The Catholic University of America},
                city={Washington},
                state={D.C.},
                postcode={20064}, 
                country={USA}}

\cortext[cor1]{Corresponding author}

\begin{abstract}
Although unmanned vehicle fleets offer efficiency in transportation, logistics and inspection, their susceptibility to failures poses a significant challenge to mission continuity. We study the Multi-Depot Rural Postman Problem with Rechargeable and Reusable Vehicles (MD-RPP-RRV) with vehicle failures, where unmanned rechargeable vehicles placed at multiple depots with capacity constraints may fail while serving arc-based demands. To address unexpected vehicle breakdowns during operation, we propose a two-stage real-time rescheduling framework. First, a centralized auction quickly generates a feasible rescheduling solution; for this stage, we derive a theoretical additive bound that establishes an analytical guarantee on the worst-case rescheduling penalty. Second, a peer auction refines this baseline through a problem-specific magnetic field router for local schedule repair, utilizing parameters calibrated via sensitivity analysis to ensure controlled computational growth. We benchmark this approach against a simulated annealing metaheuristic to evaluate solution quality and execution speed. Experimental results on 257 diverse failure scenarios demonstrate that the framework achieves an average runtime reduction of over 95\% relative to the metaheuristic baseline, cutting rescheduling times from hours to seconds while maintaining high solution quality. The two-stage framework excels on large-scale instances, surpassing the centralized auction in nearly 80\% of scenarios with an average solution improvement exceeding 12\%. Moreover, it outperforms the simulated annealing mean and best results in 59\% and 28\% of scenarios, respectively, offering the robust speed-quality trade-off required for real-time mission continuity.
\end{abstract}



\begin{keyword}

 Transportation and Logistics \sep Auctions \sep Simulated Annealing \sep  Rural postman problem \sep Vehicle failures   

\end{keyword}

\end{frontmatter}



\section{Introduction}

Unmanned battery-operated rechargeable vehicles, including Unmanned Aerial Vehicles (UAVs) and Unmanned Ground Vehicles (UGVs), are increasingly deployed for inspection of infrastructural networks in rural environments. In practice, these systems are applied to scenarios such as power transmission line inspection \citep{xing2023autonomous}, oil and gas pipeline monitoring \citep{karkoub2020gas}, and rural last-mile parcel delivery \citep{alverhed2024autonomous}. These applications mirror the Rural Postman Problem (RPP) \citep{eiselt1995arc}, requiring the traversal of specific path segments (edges) for service, maintenance, or inspection.

While unmanned systems offer potential for cost-effectiveness and operational efficiency \citep{uav-app1, uav-app2}, their deployment introduces distinct challenges not typically encountered with traditional manned vehicles. Unlike manned fleets, which generally exhibit high reliability and extended operational ranges, unmanned vehicles are constrained by limited battery capacities and a significantly higher susceptibility to stochastic failures. Consequently, shifting the focus from manned to unmanned fleets fundamentally alters the characterization of the routing problem, necessitating new strategies to ensure transportation reliability and performance.

The failure rate for drones, for example, is approximately 1 in 1,000 flight hours, two orders of magnitude higher than commercial aviation's 1 in 100,000 flight hours, and sophisticated UAV systems can face an overall failure rate of 25\% \citep{motivation}. Consequently, routing models must account for frequent interruptions and the need for dynamic rescheduling, constraints that are less critical in manned vehicle logistics. Recognizing the inherently high failure rates of unmanned systems is central to the motivation for this work. Such failures can lead to considerable delays and disruptions within the transportation network, emphasizing the need for strategies that improve the reliability and robustness of vehicle operations.

Although many preventive maintenance approaches have been proposed to increase the reliability of unmanned vehicles \citep{motivation, motivation1}, any failure during a mission requires adapting the operational plan to account for the loss of the vehicle. This paper, therefore, focuses on effectively managing and mitigating the impact of vehicle failures in transportation-oriented routing problems, specifically addressing the challenges of rerouting and task reallocation to ensure efficient mission completion despite unexpected vehicle breakdowns.

We formally study the Multi-Depot Rural Postman Problem with Rechargeable and Reusable Vehicles (MD-RPP-RRV), with the vehicles having limited capacity (operation time) but can be recharged and reused for multiple trips from multiple depots to traverse a subset of required edges in a weighted undirected connected graph. The objective is to minimize the mission time or the maximum trip time, defined as the maximum time spent by any individual vehicle to complete its assigned trips, subject to the constraint that all required edges are traversed. Extending our previous work which assumed no vehicle failures \citep{md-rpp-rrv}, this study considers the possibility of stochastic vehicle failures during the mission. \par

The MD-RPP-RRV is NP-hard as it generalizes the RPP \citep{rpp_nphard} with multiple depots and capacity constraints. The introduction of stochastic vehicle failures further increases this complexity by transforming the static routing problem into a dynamic one, necessitating algorithms that can adjust routes in real-time to unpredictable disruptions. To address these challenges, we propose a reactive rescheduling approach that To address these challenges, we propose a reactive rescheduling approach that balances solution quality with the speed required to restore mission feasibility in real-time. We introduce a two-stage framework that operates without prior knowledge of failure timing or location. The first stage utilizes a centralized auction algorithm,which maintains a global view of all vehicle states and enables rapid, efficient task reallocation. This phase reassigns trips from failed vehicles to active agents to minimize the mission time increment. Subsequently, in the second stage, a peer auction employing a new magnetic field router refines this initial solution through in-depth local schedule repair. To rigorously evaluate our framework, we benchmark its performance against Simulated Annealing (SA) metaheuristic and compare solution quality and execution times. 

The main contributions of this paper are the following:
\begin{enumerate}

    \item A two-stage reactive framework that integrates a centralized auction for rapid task reallocation with a peer auction utilizing a novel magnetic field router for local schedule repair. This approach advances the literature by reformulating the MD-RPP-RRV as a dynamic variant of the Generalized Assignment Problem \citep{gap}. It specifically addresses gaps in existing methods \citep{cenauc3, cenauc5, cenauc6} by enabling the assignment of multiple failed trips to single agents and adapting to fluctuating fleet sizes during the mission, not addressed by existing auction-based arc routing methods.

    \item A theoretical analysis deriving a worst-case additive performance bound for the centralized auction. Unlike standard competitive ratios, which can be unstable in dynamic routing contexts, this bound explicitly characterizes the rescheduling penalty as an unavoidable additive cost driven by the vehicle's battery capacity and recharge time. This establishes an analytical guarantee on the maximum deviation of the mission time from an offline optimal solution with perfect failure foreknowledge.

    \item Comprehensive experimental validation across 257 failure scenarios distinguishes this work through a rigorous sensitivity analysis that identifies tractable peer-auction parameter regimes that prevent computational blowup. We benchmark the framework against a reactive simulated annealing metaheuristic \citep{md-rpp-rrv}, demonstrating that the proposed approach reduces average runtime by over 95\%, cutting computational time from hours to minutes or seconds. This efficiency yields a pragmatic trade-off for mission continuity, outperforming the average metaheuristic results and maintaining solution quality within 8\% of the best-known metaheuristic result.

\end{enumerate}

The remainder of this paper is organized as follows:
Section \ref{sec:literature-review} presents a literature review of related works. Section \ref{sec:prob-desc} provides the operational assumptions and formally defines the MD-RPP-RRV. Section \ref{sec:approach} details the proposed two-stage reactive framework comprising of centralized and peer auction. Section \ref{sec:results} presents the experimental evaluation, including a rigorous sensitivity analysis of algorithmic parameters and a comparative benchmark against a reactive simulated annealing metaheuristic across 257 failure scenarios. The section also derives a theoretical performance bound for the proposed centralized auction. Finally, Section \ref{sec:conclusion} concludes the paper.

\section{Literature Review}
\label{sec:literature-review}

The MD-RPP-RRV involving vehicle failures extends the classical Multi-Depot Rural Postman Problem (MD-RPP) \citep{md-rpp, md-rpp1, md-rpp2} by incorporating multiple trips and vehicle failure uncertainties. While vehicle failures have not been extensively studied in the specific context of the RPP, related research exists in its variants, such as the Capacitated Arc Routing Problem (CARP) \citep{carp, carp1, carp2, carp4} and the Vehicle Routing Problem (VRP) \citep{vrp, vrp1, vrp5}. This review examines relevant literature across dynamic routing and auction algorithms to identify key research gaps, specifically focusing on the lack of efficient, real-time approaches for handling multiple vehicle failures in multi-depot scenarios without relying on auxiliary resources.

\subsection{Dynamic Routing with Vehicle Failures}
The management of vehicle failures falls under the broader research area of dynamic routing, where operations must adapt to real-time disruptions.

In the domain of Arc Routing, the Dynamic Capacitated Arc Routing Problem (DCARP) extends standard CARP by incorporating dynamic changes in demand \citep{dcarp-demand}, service cost \citep{dcarp-service-cost}, and vehicle availability \citep{dcarp-availability}. Liu et al \citep{dcarp} addressed a multi-depot DCARP considering fluctuations in vehicle availability and demand, proposing a memetic algorithm with a split scheme to minimize travel distance. While this approach effectively updates routes upon interruptions, it relies on the assumption that failed vehicles are repaired and reused. This overlooks the transport and repair time critical for time-sensitive missions where immediate reuse is impossible. Similarly, Licht et al. \citep{rarp} introduced the Rescheduling Arc Routing Problem (R-ARP) and proposed a Mixed Integer Linear Programming (MILP) based local repair strategy that solves a reduced subproblem for single-depot operations under random single-vehicle failure. Their method minimizes disruption costs by locally modifying the routes of nearby vehicles that remain operational. However, this approach is limited to single-vehicle failures and uncapacitated vehicles; extending such MILP-based local repair to multi-depot settings with capacitated vehicles and multiple concurrent failures may incur significant computational overhead.

In the domain of node routing (VRP), the problem is often referred to as the Vehicle Rescheduling Problem (VRSP). Li et al. \citep{brp, vrsp} addressed the Single Depot Vehicle Rescheduling Problem using sequential and parallel auction algorithms to dynamically reassign trips. While their results demonstrated that parallel auctions significantly improved computational efficiency for large instances, their node-based model optimizes travel between discrete locations, whereas the MD-RPP-RRV imposes the distinct constraint of continuously traversing specific edges for service. Li et al. \citep{vrsp1} further proposed a Lagrangian heuristic to minimize schedule deviations, which proved effective for late-trip failures but less so for early-stage breakdowns. Additionally, Mu et al. \citep{vrp-vb} utilized tabu search to minimize disruption costs in the VRSP. A critical limitation of this study, however, is the assumption that an extra vehicle is always available at the depot to handle breakdowns. This reliance on redundant resources leaves a significant gap in developing robust solutions for lean fleets where no backup vehicles are available and tasks must be redistributed among the existing active fleet.

\subsection{Auction Algorithms for Task Reallocation}
Since the MD-RPP-RRV with failures can be formulated as a variant of the Generalized Assignment Problem (GAP) \citep{gap}, auction algorithms \citep{aucalg, aucalg1, aucalg2} offer a promising mechanism for task allocation. These approaches generally fall into centralized or decentralized categories.

Decentralized (distributed) auction algorithms \citep{disauc1, disauc2, disauc3, disauc4, disauc5} rely on peer-to-peer interactions, making them robust to communication failures. However, they are prone to converging on suboptimal, locally optimal decisions. In the context of the MD-RPP-RRV, where multiple simultaneous failures require tight global coordination to manage battery constraints, the inherent suboptimality and slower convergence of pure decentralized methods render them less suitable for rapid recovery.

Conversely, centralized auctions \citep{fullcenauc1, fullcenauc2, fullcenauc3} employ a central auctioneer to maintain a global view of vehicle states, enabling quicker decision-making and global optimization of mission time. Whisle computationally demanding for NP-hard allocation problems \citep{cenauc1}, heuristic approaches such as combinatorial \citep{cenauc2, cenauc3} and greedy auctions \citep{cenauc3} can significantly reduce the computational burden. However, existing centralized auction methods typically address static assignments or "one task per agent" scenarios. They lack mechanisms to handle the complex dynamics of the MD-RPP-RRV, specifically the requirement to assign multiple failed trips to a single agent and to dynamically reallocate these trips among a fluctuating fleet of active vehicles.

\subsection{Summary of Contributions vs. Existing Literature}
To clearly distinguish this study from the existing state of the art, Table \ref{tab:lit_comparison} summarizes the differences in terms of problem characteristics and solution methodology.

\begin{table}[ht]
\caption{Comparison of this paper with related literature}
\label{tab:lit_comparison}
\centering
\resizebox{\textwidth}{!}{%
\begin{tabular}{lccccc}
\hline
\textbf{Reference} & \textbf{Problem Type} & \textbf{Multi-Depot} & \textbf{Rechargeable} & \textbf{Failures} & \textbf{Methodology} \\ \hline
\citep{md-rpp2} & MD-RPP & Yes & No & None & Heuristics \\
\citep{dcarp} & DCARP & Yes & No & Failures (Reuse) & Memetic Algorithm \\
\citep{rarp} & R-ARP & No & No & Single Failure & Local MILP \\
\citep{vrsp} & VRSP (Node) & No & No & Failures & Auction \\
\citep{md-rpp-rrv} & MD-RPP-RRV & Yes & Yes & None & Greedy, SA \\ \hline
\textbf{This Paper} & \textbf{MD-RPP-RRV} & \textbf{Yes} & \textbf{Yes} & \textbf{Multiple (Random)} & \textbf{Centralized + Peer Auction} \\ \hline
\end{tabular}%
}
\end{table}

As illustrated in Table \ref{tab:lit_comparison}, this paper addresses the specific intersection of multi-Depot, rechargeable, and reusable arc routing under stochastic multiple failures. It overcomes the limitations of previous works by proposing a two-stage auction framework that does not rely on spare vehicles or infinite repair capabilities, providing a robust solution for autonomous fleets with capacity constraints.

\section{Problem Description}
\label{sec:prob-desc}

This section formally describes the MD-RPP-RRV subject to stochastic vehicle failures. In our previous study \citep{md-rpp-rrv}, we presented a MILP formulation for the static version of this problem, which assumed reliable vehicle operations. In the present work, we extend that framework to a dynamic environment where multiple vehicles may fail unexpectedly, with failure times and locations unknown \textit{a priori}. To ensure this manuscript remains self-contained while avoiding redundancy, we do not reproduce the full static MILP formulation here. Instead, we adopt the same problem definitions and symbolic notation used in the previous study, augmenting them with the necessary descriptions and constraints to characterize the stochastic nature of vehicle failures and the resulting dynamic fleet reduction.

\subsection{Assumptions}
\label{assum}
The operational framework is defined by the following assumptions:
\begin{enumerate}
    \item All vehicles are homogeneous, possessing identical battery capacity, recharge time, and constant uniform travel speed.
    \item A required edge is considered successfully traversed only if a vehicle traverses it and subsequently completes the trip by reaching a depot node. If a failure occurs after traversal but before the vehicle returns to a depot, the edge is marked as un-serviced and must be traversed by another vehicle. This assumption models scenarios involving high-bandwidth data acquisition (e.g., LiDAR, high-resolution imaging) in rural environments where real-time wireless transmission is infeasible, necessitating physical data retrieval at the depot.
    \item Each required edge can be traversed by any vehicle without restrictions on vehicle-route assignments.
    \item Vehicle failure can occur during a trip (edge traversal) but not during recharging processes at a depot.
    \item At least one vehicle remains functional (i.e., the set of active vehicles is non-empty).  
    (If no vehicle remains functional, then there is no rescheduling problem to solve.)
    \item Vehicle failures are detected and communicated to the central system and other vehicles immediately upon occurrence. This assumes the use of standard low-bandwidth telemetry ``heartbeat'' signals. Unlike high-bandwidth sensor data which necessitates physical retrieval (Assumption 2), status flags require negligible bandwidth and are reliably transmitted over long-range networks, where signal termination indicates immediate failure.
    \item Vehicle failures are instantaneous events rather than gradual performance degradations.
\end{enumerate}

\subsection{Problem Inputs and Objective}
\label{sec:inputs-obj}
An instance of the MD-RPP-RRV includes a fleet of vehicles that must perform a set of tasks.
The fleet consists of $K$ homogeneous vehicles. Each vehicle operates under strict capacity constraints, specifically a maximum battery capacity $C$ expressed as the maximum allowable operational time per trip. Vehicles are reusable and can perform multiple trips. Between consecutive trips, a vehicle must undergo a fixed recharge time $R_T$ to replenish its battery at a depot.

The vehicles travel along an undirected weighted connected graph $G = (N, E, T)$. The set $N$ represents the nodes in the network, and $E$ denotes the set of edges that connect these nodes. 
The vehicles must use these edges to travel between nodes.  
The edges in $E_u \subseteq E$ represent the required edges or tasks that must be serviced at least once. The fleet operates from a set of depot nodes $N_d \subseteq N$ where vehicles can start, stop, and recharge. 
The set of weights $T = \{ t(i,j): (i,j) \in E \}$, where $t(i,j)$ corresponds to the time required to traverse edge $(i,j)$. This traversal time is calculated as $l(i,j)/S$, where $l(i,j)$ is the length of the edge and $S$ is the constant uniform speed of the vehicles.

There is a set of vehicles $F \subset K$ that will experience a breakdown. Each vehicle $k \in F$ has a  failure time $f_k$. While these parameters $F$ and $f_k$ exist within the problem environment, their values are unknown to the router at the start of the mission.  Each one is revealed only when the failure event occurs. 
After its failure at time $f_k$, the vehicle ceases all operations and cannot continue its assigned route or return to the depot.

The primary objective is to determine a routing schedule such that every required edge in $E_u$ is successfully traversed by a vehicle that returns to a depot. The specific goal is to minimize the mission time, denoted as $\beta$. A valid solution must ensure flow conservation, meaning vehicles must start and end trips at depots and satisfy demand constraints such that all functional required edges are covered. Furthermore, the solution must strictly satisfy the capacity limit so that no trip exceeds duration $C$ and that appropriate recharge intervals $R_T$ are inserted between trips. The dynamic nature of the problem necessitates that the routing schedule be adjusted in real-time to ensure these objectives and constraints are met despite the disruptions that the unexpected failures cause.

Given the NP-hard complexity of the underlying RPP and the stochasticity introduced by dynamic failures, solving this problem requires efficient algorithmic strategies capable of rapid rescheduling. The following section details our proposed two-stage framework designed to address these computational challenges.

\begin{table}[t!]
\centering
\caption{Nomenclature}
\label{tab:nomenclature}
\renewcommand{\arraystretch}{1.1}
\resizebox{14cm}{!}{
\begin{tabular}{ll}
    \hline
    \textbf{Variable} & \textbf{Description} \\
    \hline
    $C$ & Maximum vehicle battery capacity (time units) \\
    $E_u$ & Set of required edges to be serviced; $E_u \subseteq E$ \\
    $F$ & Subset of vehicles that experience failure; $F \subset \{1, ..., K\}$ \\
    $f_k$ & Failure time of vehicle $k \in F$ \\
    $G = (N, E, T)$ & Network graph with nodes, edges, and edge weights \\
    $K$ & Total number of vehicles \\
    $N_d$ & Set of depot nodes; $N_d \subseteq N$ \\
    $R_T$ & Fixed time required to recharge a vehicle at a depot \\
    \hline
    \multicolumn{2}{l}{\textit{State Variables}} \\
    $n_k$ & Current node position of vehicle $k$ in graph $G$\\
    $P_k$ & Route for vehicle $k$ (sequence of trips) \\
    $S_k$ & Status of vehicle $k$ (True = Active, False = Failed) \\
    $t$ & Current simulation time \\
    $t_m$ & Mission time; $t_m = \max_{k} y_k$ \\
    $y_k$ & Arrival time of vehicle $k$ at the depot after its last scheduled trip \\
    \hline
    \multicolumn{2}{l}{\textit{Centralized Auction Parameters}} \\
    $\Delta r$ & Increment step for search radius expansion \\
    $M_F$ & Dictionary mapping failed trips to their unserviced required edges \\
    $r_i$ & Initial search radius for finding candidate vehicles \\
    $\tau_f$ & A specific failed trip being auctioned \\
    \hline
    \multicolumn{2}{l}{\textit{Peer Auction \& Router Parameters}} \\
    $i_{anch}$ & Index of the last visited depot (anchor) in a trip \\
    $i_{lock}$ & Index of the last serviced required edge (lock) in a trip \\
    $L$ & Computational budget (max transactions per receiver) \\
    $R$ & Maximum peer auction rounds \\
    $S(u,v)$ & Convex scoring function value for moving from $u$ to $v$ \\
    $W$ & Maximum window size for trip combinations in Peer Auction \\
    $w_{cap}$ & Capacity usage ratio ($t_{cur} / C$) used in scoring \\
    $\mathcal{X}$ & Set of proposed transactions (swaps/relocations) \\
    \hline
\end{tabular}}
\end{table}

\section{Proposed Approach}
\label{sec:approach}
This section presents the proposed two stage reactive framework, includes pseudocode for the key procedures, and demonstrates it with a small instance of the MD-RPP-RRV with vehicle failures. 
Table \ref{tab:nomenclature} provides the nomenclature used in the subsequent sections.

\subsection{MD-RPP-RRV Routes and Rescheduling Example}
\label{routetypes}
To aid the reader's understanding of the proposed framework, this section presents an illustrative example demonstrating a feasible MD-RPP-RRV solution, the impact of a dynamic vehicle failure, and the subsequent reactive rescheduling process. A feasible solution specifies a route for each vehicle. A route includes one or more trips, each beginning at a depot and ending at the same or a different depot. The capacity constraint limits the length of a trip. Although the primary objective of each vehicle is to traverse required edges, a vehicle might need to make a trip without covering any required edges, or end a trip at a different depot to reposition itself for subsequent tasks.

\begin{figure}[b!]
    \centering
    \includegraphics[width=\textwidth]{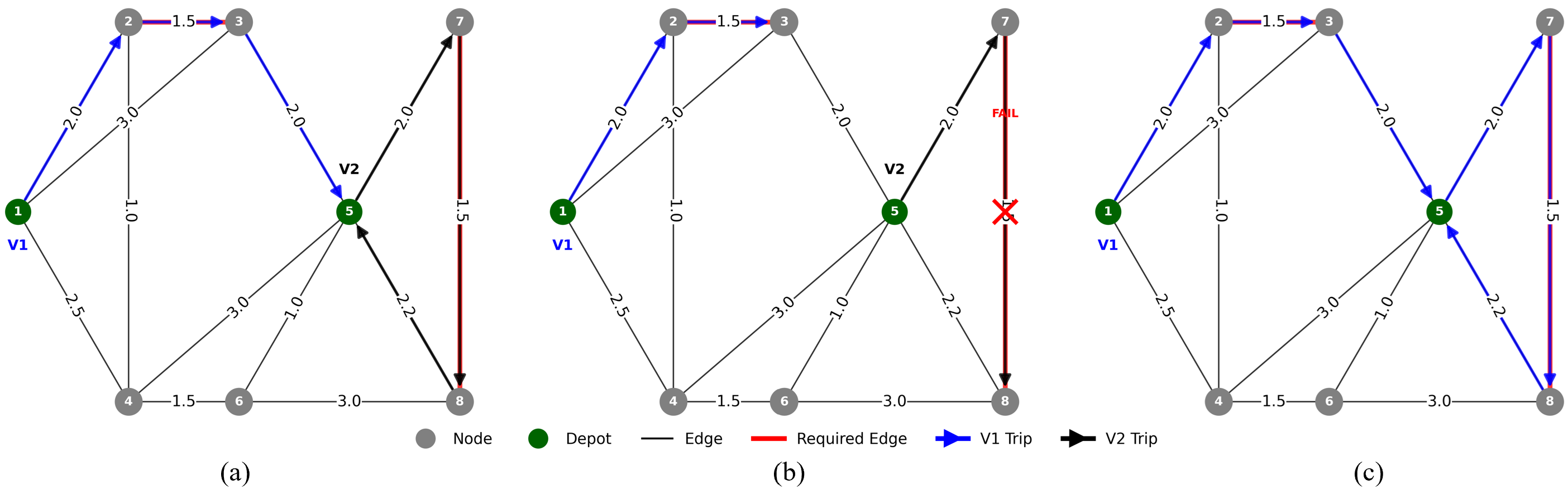}
    \caption{\textrm{ (a) Initial two-vehicle plan, (b) dynamic failure of $V_2$ on a required edge, and (c) rescheduled $V_1$ route covering the unserved required edge.}
    \label{case2}}
\end{figure}

Consider an instance with an undirected graph $G$ that has 8 nodes, 13 edges, 2 depot nodes (Nodes 1 and 5), and 2 required edges ($(2,3)$ and $(7,8)$), shown in Figure \ref{case2}{\color{blue}(a)}. There are two vehicles: $V_1$ starts at the depot at Node 1, and $V_2$ starts at the depot at Node 5. The vehicle capacity is $C = 7$ time units, and the recharge time is $R_T = 1.1$ time units.

The initial plan assigns one trip to each vehicle. $V_1$ performs the trip $\{1-2-3-5\}$, which covers the required edge $(2,3)$ and ends at Node 5. The total duration of this trip is $2.0 + 1.5 + 2.0 = 5.5$ time units, which is feasible ($5.5 \le C$). Simultaneously, $V_2$ is assigned the trip $\{5-7-8-5\}$, which covers the required edge $(7,8)$ and returns to Node 5. This trip takes $2.0 + 1.5 + 2.2 = 5.7$ time units, which is also feasible.

Figure \ref{case2}{\color{blue}(b)} illustrates a dynamic failure scenario where $V_2$ fails while traversing the required edge $(7,8)$. This failure renders $V_2$'s remaining route infeasible and leaves the required edge $(7,8)$ unserviced.

Figure \ref{case2}{\color{blue}(c)} shows the updated route for $V_1$ after reactive rescheduling. $V_1$ becomes responsible for the unserved required edge $(7,8)$ originally assigned to $V_2$. $V_1$ completes its first trip at Node 5 at $t=5.5$. It then recharges at Node 5, which takes $1.1$ time units. Finally, $V_1$ executes the second trip $\{5-7-8-5\}$ to traverse the required edge $(7,8)$, taking an additional $5.7$ time units. Thus, the updated route is $P_1 = \{ 1-2-3-5, 5-7-8-5 \}$. The vehicle finishes its route at $y_1 = 5.5 + 1.1 + 5.7 = 12.3$. The new mission time is $t_m = 12.3$ time units.

\subsection{Overview of the Reactive Framework}
\label{sec:flowchart_desc}

\begin{figure}[t!]
    \centering
    \includegraphics[width=0.6\textwidth]{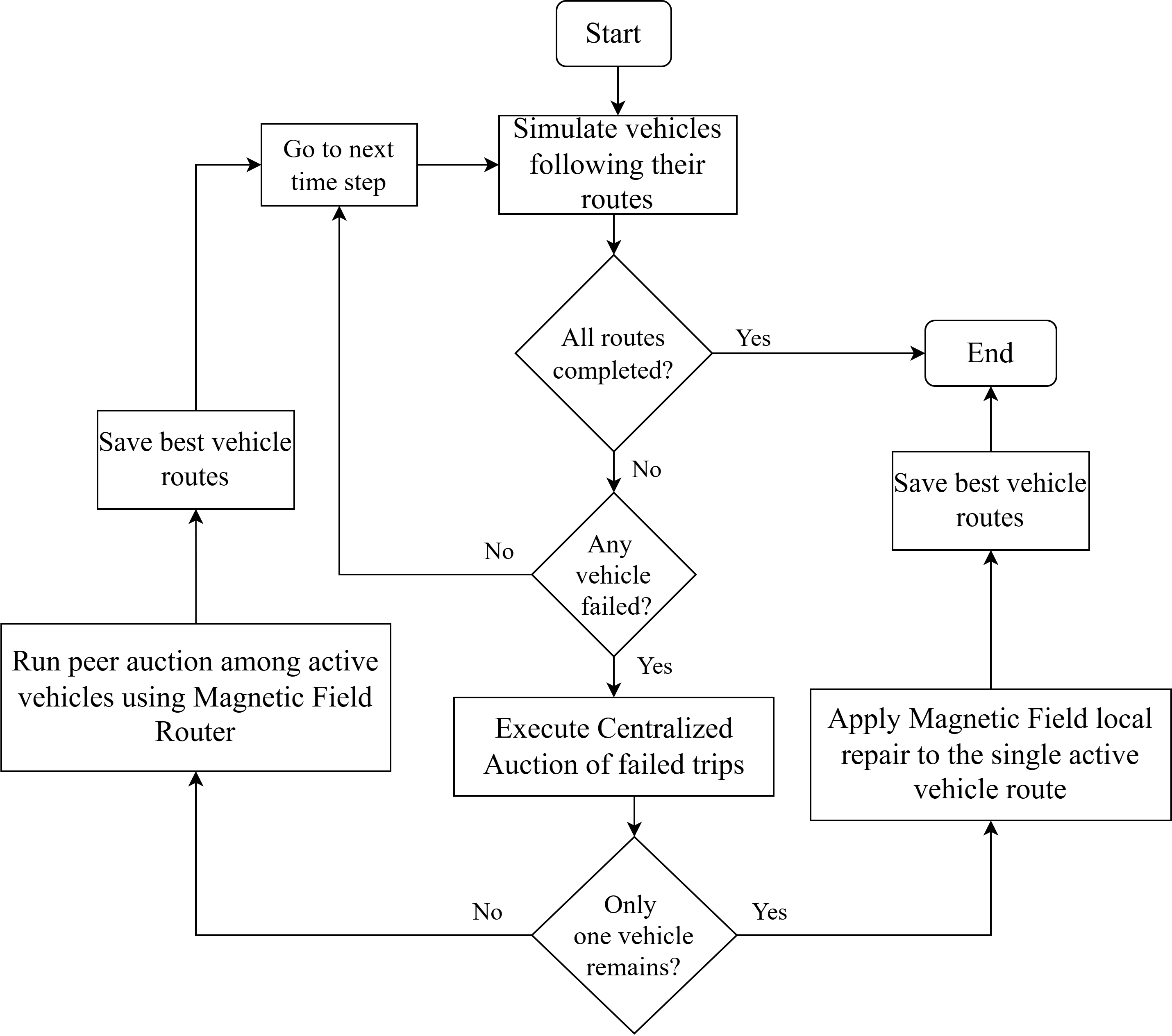}
    \caption{Flowchart of the proposed reactive framework. The process begins with failure detection and task filtering, followed by a rapid Centralized Auction for feasibility and a Peer Auction refinement phase for solution quality.}
    \label{fig:framework_flowchart}
\end{figure}

The overall decision flow of the proposed reactive framework is visualized in Figure \ref{fig:framework_flowchart}. The process begins by using a simulated annealing algorithm to generate initial routes, which have no vehicle failures. 
The vehicles then begin to follow their routes.
As they progress, the system continuously monitors the fleet for failure events, which occur at the times $f_k$, $k \in F$. Upon detection of a failure, the framework first isolates the specific trip interrupted by the breakdown.

To minimize computational overhead, the algorithm reviews the failed vehicle's remaining schedule to identify the trips that contain unserviced required edges.  (Purely logistical trips, such as repositioning movements between depots without traversing required edges, are discarded.) The remaining critical trips are aggregated into an auction pool. If this pool is non-empty, the two-stage rescheduling protocol is activated.

The first stage executes the \textsc{Centralized Auction} procedure (described in Section \ref{comb-auction}). This mechanism prioritizes operational speed and feasibility, rapidly reallocating the failed trips to active vehicles to restore immediate service continuity. Once a valid schedule is established, the framework transitions to the refinement stage. If multiple vehicles remain active, the \textsc{Peer Auction} procedure (described in Section \ref{peer_auction}) is triggered to balance workloads through negotiation. 
If only one vehicle remains, the peer auction is bypassed, and the \textsc{Magnetic Router} (described in Section \ref{mag_router_desc}) is applied directly to the lone vehicle to optimize the greedy insertions made by the centralized auction. This adaptive approach ensures that the system reacts instantaneously to disruptions, due to vehicle failure, and improves the fleet's mission time regardless of fleet size.

\begin{algorithm}[t!]
\fontsize{11}{14.5}\selectfont
\caption{MD-RPP-RRV with Vehicle Failures}
\label{simulation}
\begin{algorithmic}[1]
\Procedure{Simulation}{$G, K, F, C, R_T, N_d, n_K, r_i, \Delta r, R, W, L$}
\State $P_K, y_K \leftarrow \Call{SA}{G, N_d, K, n_K, y_K, E_u, R_T, C}$
\State $t_m \leftarrow \underset{k =  1, \ldots, K}{\max}y_k, \; S_k \leftarrow \text{True}, \; \forall k \in K$
\State $D_d \leftarrow \Call{DepottoDepotRoutes}{G, N_d, R_T, C}$
\State Initialize dictionary $M_F \leftarrow \emptyset$    
\For{$t = 0 \to t_m$}
    \ForAll{$k = 1, \ldots, K$}
        \If{$t = f_k, k \in F$ and $S_k = \text{True}$}  
            \State $S_k \leftarrow \text{False}$
            \State $i \leftarrow \Call{tripindex}{G, P_k, R_T, t}$ 
            \State $\tau_f \leftarrow \emptyset, e_f \leftarrow \emptyset$
            \For{$j = i \to$ len($P_k$)}      
                    \State $e_f \leftarrow \Call{requiredtrip}{E_u, P_k[j]}$
                    \If{$e_f \neq \emptyset$}
                        \State $\tau_f \leftarrow P_k[j]$      
                        \State$M_F[\tau_f] \leftarrow e_f$
                    \EndIf
            \EndFor
        \EndIf
    \EndFor
    \If{$M_F \neq \emptyset$}
        \State $P_K, y_K \leftarrow \Call{CentralizedAuction}{G, M_F, R_T, D_d, t, K, P_K, S_K, y_K, r_i, \Delta r}$ 
        \State $P_K \leftarrow \Call{PeerAuction}{G, K, C, R_T, P_K, y_K, N_d, S_K, t, R, W, L}$ 
        \State $t_m \leftarrow \Call{Missiontime}{G, P_K, R_T}$  
    \EndIf
\EndFor
\State \Return $P_K$
\EndProcedure
\end{algorithmic}
\end{algorithm}

\subsection{MD-RPP-RRV with Vehicle Failures}

Solving an instance of the MD-RPP-RRV with vehicle failures requires a dynamic approach: generating an initial set of routes assuming perfect vehicle reliability, and then rescheduling reactively when failures occur. Algorithm \ref{simulation} outlines this simulation process.

In this study, we determined the vehicles' initial routes using a simulated annealing algorithm (Line 2). We then simulated the mission by iterating time $t$ from 0 to the mission completion $t_m$ (Line 6), constantly monitoring the fleet status. Note that, for any vehicle $k$, its route $P_k$ has len($P_k$) trips; $P_k[j]$ denotes the $j$-th trip. Also, $P_K$ is the set of all routes,  $y_K$ is the set of all arrival times, and $S_K$ is the set of all vehicle status.

When a vehicle $k \in F$ fails at time $f_k$ (Line 8), the algorithm must identify precisely which tasks in the failed vehicle's schedule require reallocation.
This identification is handled by two high-level helper procedures:
\begin{enumerate}
    \item \textsc{TripIndex} (Supplementary Materials Algorithm 2): This procedure (Line 10) iterates through the schedule for failed vehicle $k$, summing the duration of completed trips and recharge intervals. It returns the specific index $i$ of the trip that vehicle $k$ was traversing at time $f_k$.
    \item \textsc{RequiredTrip}: After the failure point is identified, the algorithm iterates through all subsequent trips in the schedule for vehicle $k$ (Lines 12-18). 
    Because there might be some repositioning trips, a \textsc{RequiredTrip} check filters the schedule. It inspects the edges within a trip and returns the set of required edges $e_f$ that have not yet been serviced.
\end{enumerate}

Only trips containing unserviced required edges are added to the auction dictionary $M_F$ (Line 16). This filtering step is crucial as it eliminates the computational overhead of auctioning purely logistical trips. Finally, if $M_F$ is non-empty, the two-stage rescheduling framework is triggered: first, the \textsc{CentralizedAuction} procedure generates a feasible reallocation of these trips to active vehicles, followed immediately by the \textsc{PeerAuction} procedure, which refines the solution through local improvements (Lines 22-23).

\subsection{Centralized Auction (CA) Procedure}
\label{comb-auction}

The \textsc{CentralizedAuction} procedure (Algorithm \ref{auction_procedure}) rapidly reallocates failed trips $\tau_f$ from the set $M_F$ to active, non-failed vehicles. The allocation strategy is greedy, with the aim of minimizing the immediate increase in the overall mission time $t_m$. The procedure iterates through each failed trip (Line 5) and executes a three-step process for each: Candidate Search, Bidding, and Assignment.

\begin{algorithm}[t!]
\fontsize{11}{14.5}\selectfont
\caption{Centralized Auction Procedure}
\label{auction_procedure}
\begin{algorithmic}[1]
\Procedure{CentralizedAuction}{$G, M_F, R_T, D_d, t, K, P_K, S_K, y_K, r_i, \Delta r$}
    \While{$M_F \neq \emptyset$}
    \State Initialize $B_{min} \leftarrow \infty, k^{*} \leftarrow \emptyset, R_b \leftarrow \emptyset, \tau_f^{*} \leftarrow \emptyset$
    \State $t_m \leftarrow \Call{Missiontime}{G, P_K, R_T}$  
    \ForAll{$\tau_f, e_f \in M_F$}
        \State Initialize $r \leftarrow r_i, K_r \leftarrow \emptyset$
        \While{$K_r = \emptyset$}
            \State $K_r \leftarrow \Call{Search}{G, P_K, R_T, \tau_f, r, K, S_K, t}$
            \State $r \leftarrow r + \Delta r$
        \EndWhile
        \ForAll{vehicle $k \in K_r$}
            \State $bid, R_{k}\leftarrow \Call{CalcBid}{G, D_d, \tau_f, t, e_f, k, P_k, t_m}$ 
            \If{$bid < B_{min}$}
                \State $B_{min} \leftarrow bid; R_{b} \leftarrow R_k; k^{*} \leftarrow k; \tau_f^{*} \leftarrow \tau_f$
            \EndIf
        \EndFor
        \EndFor
        \State $P_{k^{*}} \leftarrow R_b$, Update $y_{k^{*}}$
        \State Remove $\tau_f^{*}, M_F[\tau_f^{*}]$ from $M_F$
    \EndWhile
    \State \Return $P_K, y_K$
\EndProcedure
\end{algorithmic}
\end{algorithm}

\begin{figure}[b!]
\centering
\includegraphics[width=\textwidth]{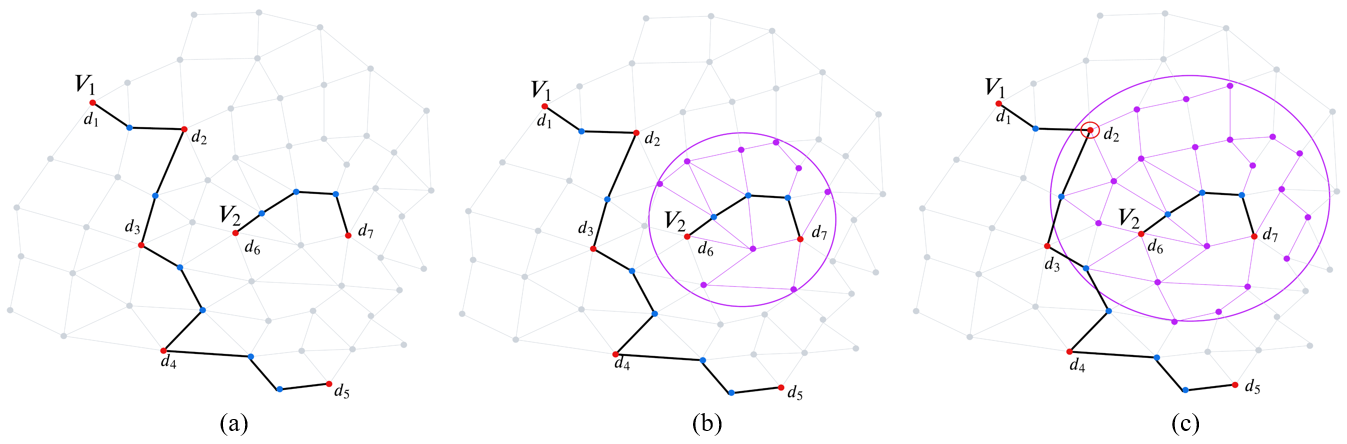}
\caption{\textrm{MD-RPP-RRV Candidate Search procedure: (a) Initial setup. (b) First search iteration with small radius yields no results. (c) Second iteration with expanded radius identifies Vehicle 1 as a candidate.}}
\label{search}
\end{figure}

\textbf{1. Candidate Search:}
To keep the algorithm efficient, the algorithm avoids evaluating the entire fleet for every reallocation task. Instead, it employs a \textsc{Search} procedure (Supplementary Materials' Algorithm 1) (Line 8) to filter for nearby candidate vehicles. As illustrated in Figure \ref{search}, the search begins with a small radius $r_i$ centered on the depots associated with the failed trip. The procedure iterates through all active vehicles, checking the shortest path distance between their scheduled depot stops and the failed trip's location. If the distance to either the start or end depot of the failed trip falls within the current radius $r$, the vehicle is added to the candidate set $K_r$. If the set $K_r$ remains empty after checking all vehicles, the radius is incrementally expanded by $\Delta r$ (Figure \ref{search}(b)-(c)) (Lines 7-10). This iterative expansion ensures that only nearby vehicles are considered in the bidding, significantly optimizing the process.

\textbf{2. Bidding:}
Once a non-empty set of candidate vehicles $K_r$ is identified, each candidate calculates a bid representing the ``cost'' of incorporating the failed trip $\tau_f$ into its schedule (Line 12). This process is handled by a \textsc{CalcBid} procedure (Supplementary Materials Algorithm 3), which determines the optimal insertion point within the candidate's existing route $P_k$. The procedure iterates through every future depot stop $d_r$ in the vehicle's remaining schedule to simulate a potential insertion. To rigorously maintain flow conservation, the insertion logic adapts based on the position of $d_r$ within the current route.

\begin{figure}[t!]
\centering
\includegraphics[width=0.65\textwidth]{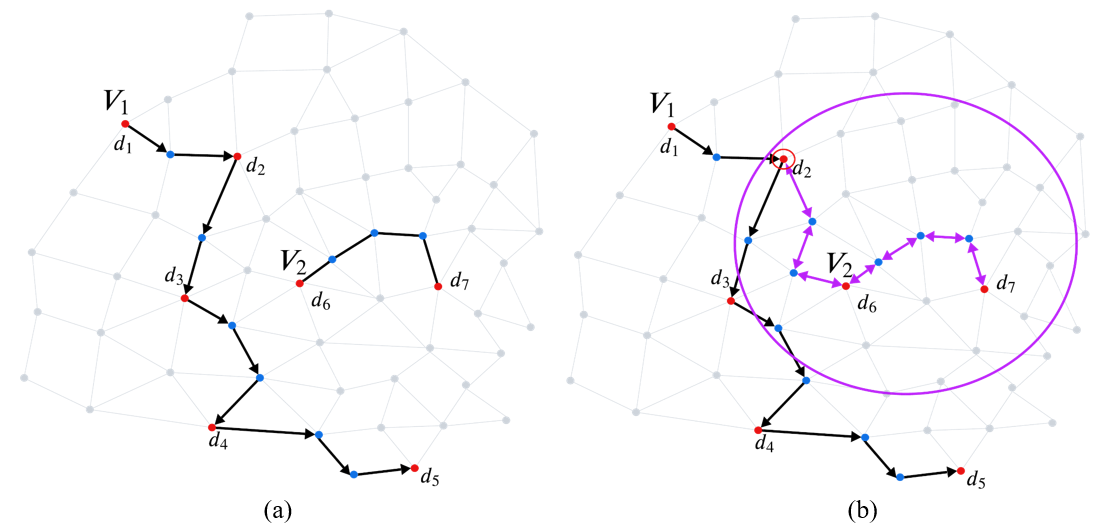}
\caption{Example of Trip Insertion: (a) Initial routes of $V_1$ and $V_2$. (b) $V_2$'s failed trip is inserted into $V_1$'s route at depot $d_2$, creating a valid sub-tour.}
\label{fig:inserttrip_example}
\end{figure}

When the insertion point corresponds to an intermediate depot, as depicted in Figure \ref{fig:inserttrip_example}, the algorithm constructs a detour cycle: the vehicle travels from $d_r$ to the start of the failed trip $\tau_f$, completes the service traversal, and returns to $d_r$ to resume its original path. Conversely, if $d_r$ represents the final depot of the route, the constraint is relaxed; the vehicle simply extends its mission to cover $\tau_f$ without the necessity of returning to $d_r$, provided the chosen direction (forward or reverse traversal of $\tau_f$) minimizes total travel time. The bid returned corresponds to the minimum resulting increase in the total mission time found across all feasible insertion points.

\textbf{3. Assignment:}
The auctioneer compares the bids received from all candidates in $K_r$ and identifies the vehicle $k^*$ offering the lowest bid $B_{min}$ (Lines 11-16). This corresponds to the vehicle capable of absorbing the failed trip with minimal increment to mission time. The failed trip $\tau_f$ is assigned to vehicle $k^*$, and its route $P_k$ is permanently updated to the optimal configuration $R_b$ determined during the bidding phase. The assigned trip $\tau^{*}_f$ is then removed from $M_F$, and the loop continues until all failed trips and respective failed required edges are reallocated.

The centralized auction prioritizes speed and feasibility, rapidly generating a valid schedule by greedily inserting entire failed trips into active routes. However, this failure trip insertion strategy inevitably introduces inefficiencies; even optimal insertions lead to increase in mission time by at least $C + R_T$. 
Although this approach establishes a critical feasible baseline, it lacks the granularity to optimize the internal route structure. To mitigate these detours and enhance solution quality, the following section introduces the peer auction, which employs a local repair mechanism to refine the baseline schedule.

\begin{algorithm}[t!]
\fontsize{11}{14.5}\selectfont
\caption{Peer Auction (Refinement Phase)}
\label{peer_auction_pseudo}
\begin{algorithmic}[1]
\Procedure{PeerAuction}{$G, K, C, R_T, P_K, y_K, N_d, S_K, t, R, W, L$}
    \State $r_{cnt} \leftarrow 0$, $k_{imp} \leftarrow \text{True}$
    \While{$k_{imp}$ \textbf{and} $r_{cnt} < R$}
        \State $k_{imp} \leftarrow \text{False}$, $r_{cnt} \leftarrow r_{cnt} + 1$
        \State $t_m \leftarrow \Call{MissionTime}{G, P_K, R_T}$
        
        \State $D \leftarrow \underset{k = 1, \ldots, K}{\argmax} \; y_k$ \Comment{Donor}
        \State $K_R \leftarrow \operatorname{sort}_{y_k \uparrow} \big( \{\, k \in K \mid S_k = \text{True} \land k \neq D \,\} \big)$

        \State $\mathcal{X} \leftarrow \emptyset$ 
        
        \State $T_D \leftarrow \Call{GenerateTripCombinations}{P_D, t, W}$
        \ForAll{$r \in K_R$}
            \State $T_r \leftarrow \Call{GenerateTripCombinations}{P_r, t, W}$
            \State $\mathcal{X} \leftarrow \mathcal{X} \cup \Call{BuildTransactions}{D, r, T_D, T_r, L}$
        \EndFor
        
        \ForAll{$(D, r, t_D, t_r) \in \mathcal{X}$}
            \State $E'_{D}, E'_{r} \leftarrow \Call{ExchangeEdges}{P_D, P_r, t_D, t_r}$
            \State $P'_D \gets \Call{MagneticRouter}{G, E'_D, N_d, \textsc{start}(P_D[t_D[1]]), C}$ 
            \State $P'_r \gets \Call{MagneticRouter}{G, E'_r, N_d, \textsc{start}(P_r[t_r[1]]), C}$ 
            
            \State $t_{new} \leftarrow \max(\Call{RouteTime}{P'_{D}}, \Call{RouteTime}{P'_{r}})$
            \If{$t_{new} < t_m$} 
                \State $P_{D} \leftarrow P'_{D}; P_{r} \leftarrow P'_{r}; y_D \leftarrow \Call{RouteTime}{P_{D}}; y_r \leftarrow \Call{RouteTime}{P_{r}}$
                \State $k_{imp} \leftarrow \text{True}$
                \State \textbf{break} 
            \EndIf
        \EndFor
    \EndWhile
    \State \Return $P_K, y_K$
\EndProcedure
\end{algorithmic}
\end{algorithm}

\subsection{Peer Auction (PA) Procedure}
\label{peer_auction}

The \textsc{PeerAuction} procedure (Algorithm \ref{peer_auction_pseudo}) iteratively improves the feasible solution generated by the centralized auction. 
Although the initial solution ensures mission completion, it often results in imbalanced routes due to the greedy nature of the centralized assignment. The peer auction addresses this by facilitating cooperative transactions between active vehicles to redistribute required edges and reduce the mission time $t_m$.

The algorithm operates in rounds, continuing as long as an improvement is found or until a maximum round limit $R$ is reached (Line 3). At the start of each round, the current mission time $t_m$ is calculated (Line 5). The algorithm then identifies the donor vehicle $D$ which is the vehicle with the maximum route time i.e., the one that completes its route last (Line 6). The other active, non-failed vehicles are designated as candidate receivers (the set $K_R$), sorted by their route times (Line 7). This strategy targets the donor vehicle that defines the mission time and attempts to move required edges to the underutilized vehicles with $y_k < y_D$.

\begin{algorithm}[b!]
\fontsize{11}{14.5}\selectfont
\caption{Generate Trip Combinations}
\label{trip_comb}
\begin{algorithmic}[1]
\Procedure{GenerateTripCombinations}{$P_k, t, W$}
    \State $I \leftarrow \Call{GetFutureTripIndices}{P_k, t}$
    \State $C_t \leftarrow \emptyset$
    \State $m \leftarrow \min(W, |I|)$
    \For{$len \leftarrow 1 \textbf{ to } m$}
        \State $C_t \leftarrow C_t \cup \{ I[j : j + len] \mid 0 \le j \le |I| - len \}$
    \EndFor
    \State \Return $C_t$
\EndProcedure
\end{algorithmic}
\end{algorithm}

\subsubsection{Transaction Generation}
To explore the search space efficiently, the algorithm generates a set of candidate transactions $\mathcal{X}$. First, the \textsc{GenerateTripCombinations} procedure (Algorithm \ref{trip_comb}) constructs sets of trip combinations $T_D$ for the donor (Algorithm \ref{peer_auction_pseudo}, Line 9). This procedure identifies all future trip indices (after time $t$) $I$ that can be peer auctioned (Algorithm \ref{trip_comb}, Line 2) and generates all contiguous sub-segments of length up to $W$ (Algorithm \ref{trip_comb}, Lines 4-6). For example, if a vehicle has future trips indexed $\{3, 4, 5\}$ and $W=2$, the procedure generates the combinations $\{[3], [4], [5], [3, 4], [4, 5]\}$. This allows the algorithm to consider moving single trips or blocks of adjacent trips.

\begin{algorithm}[t!]
\fontsize{11}{14.5}\selectfont
\caption{Build Transactions}
\label{build_transac}
\begin{algorithmic}[1]
\Procedure{BuildTransactions}{$d, r, T_d, T_r, L$}
    \State $\mathcal{X} \leftarrow \emptyset$ 
    \ForAll{$t_d \in T_d$}
        \State $\mathcal{X} \leftarrow \mathcal{X} \cup \{ (d, r, t_d, \emptyset) \}$ \Comment{Relocation}
        
        \ForAll{$t_r \in T_r$}
            \If{$|\mathcal{X}| \ge L$} \State \textbf{return} $\mathcal{X}$ \EndIf
            \State $\mathcal{X} \leftarrow \mathcal{X} \cup \{ (d, r, t_d, t_r) \}$ \Comment{Swap}
        \EndFor
    \EndFor
    \State \Return $\mathcal{X}$
\EndProcedure
\end{algorithmic}
\end{algorithm}

The algorithm then iterates through each receiver $r$ (Algorithm \ref{peer_auction_pseudo}, Line 10), generating a similar set of trip combinations $T_r$ using \textsc{GenerateTripCombinations} (Line 11). The \textsc{BuildTransactions} procedure (Algorithm \ref{build_transac}) is then called to create specific proposals from these combinations (Algorithm \ref{peer_auction_pseudo}, Line 12). This procedure considers two types of moves to generate combinations: relocation, where the set of required edges associated with a combination of future trip indexes $t_d$ is moved entirely from the donor to the receiver vehicle (Algorithm \ref{build_transac}, Line 4), and swap, where the sets of required edges associated with combinations of future trip indexes $t_d$ and $t_r$ are exchanged between the donor and receiver vehicles (Algorithm \ref{build_transac}, Line 8). To maintain real-time responsiveness, we impose a computational budget $L$ on the number of transactions evaluated per receiver vehicle. This limit prevents combinatorial explosion while ensuring the most promising local moves are considered.

\subsubsection{Evaluation and Reconstruction}
The core evaluation occurs in lines 14-22 of Algorithm \ref{peer_auction_pseudo}. For each proposed transaction, the algorithm executes a virtual trade. First, the \textsc{ExchangeEdges} procedure reassigns the set of required edges associated with the transaction between the donor and receiver vehicles (Line 15). Crucially, the routes for both vehicles are then completely reconstructed from scratch using the magnetic field router (\textsc{MagneticRouter}, Lines 16-17).  Note that the \textsc{start} function returns the depot node at which a trip starts.

Unlike simple insertion heuristics, the magnetic field router (described in Section \ref{mag_router_desc}) rebuilds the route trip-by-trip for both donor and receiver vehicles based on the new required edges assignment obtained from future trip combinations of $t_d$ and $t_r$ respectively. The new local mission time $t_{new}$ is calculated as the maximum of the reconstructed route times (Line 18). If $t_{new}$ is strictly less than the current mission time $t_m$ (Line 19), the transaction is accepted immediately. The vehicle routes are updated, and the flag $k_{imp}$ is set to true (Line 21), indicating that an improvement in mission time was achieved. This triggers an immediate exit from the inner loop, skipping remaining transactions since the vehicle states have changed. The $k_{imp}$ flag subsequently controls the outer loop (Line 3); if a round completes without finding any improving transaction ($k_{imp}$ remains false), the algorithm terminates.

\subsubsection{Magnetic Field Router}
\label{mag_router_desc}

The \textsc{MagneticRouter} procedure (Algorithm \ref{mag_router}) is a local repair algorithm that rebuilds feasible vehicle routes from scratch given a set of required edges $E_{req}$. The algorithm initializes an empty global route $P$ and iteratively constructs individual trips using the \textsc{ConstructTrip} procedure until all required edges are serviced (Algorithm \ref{mag_router}, Lines 2-10). This sequential construction ensures that capacity and route continuity constraints are strictly respected while maximizing the coverage of required edges in each trip.

\begin{algorithm}[t!]
\fontsize{11}{14.5}\selectfont
\caption{Magnetic Field Router}
\label{mag_router}
\begin{algorithmic}[1]
\Procedure{MagneticRouter}{$G, E_{req}, N_d, u_{start}, C$}
    \State $P \leftarrow \emptyset$ \Comment{Global route}
    \State $E_{rem} \leftarrow E_{req}$, $u \leftarrow u_{start}$
    
    \While{$E_{rem} \neq \emptyset$}
        \State $\tau, E_{cov}, u_{end} \leftarrow \Call{ConstructTrip}{G, E_{rem}, N_d, u, C}$
        
        \If{$\tau = \emptyset$} \textbf{break} \EndIf
        
        \State Append $\tau$ to $P$
        \State $E_{rem} \leftarrow E_{rem} \setminus E_{cov}$
        \State $u \leftarrow u_{end}$ \Comment{Next trip starts where previous ended}
    \EndWhile
    \State \Return $P$
\EndProcedure
\end{algorithmic}
\end{algorithm}

\begin{algorithm}[t!]
\fontsize{11}{14.5}\selectfont
\caption{Construct Trip (Single Vehicle Trip)}
\label{const_trip}
\begin{algorithmic}[1]
\Procedure{ConstructTrip}{$G, E_{rem}, N_d, u_{start}, C$}
    \State $\tau \leftarrow [u_{start}]$, $t_{cur} \leftarrow 0$, $E_{cov} \leftarrow \emptyset$ \Comment{Initialize trip $\tau$}
    \State $i_{ld} \leftarrow 0, i_{lreq} \leftarrow 0$ \Comment{Indices of last visited depot and required edge}
    
    \While{$E_{rem} \neq \emptyset$}
        \State $u \leftarrow \tau[\text{end}]$, $w_{cap} \leftarrow t_{cur} / C$
        \State $S_{max} \leftarrow -\infty, p^* \leftarrow \emptyset, i_{join} \leftarrow |\tau|$
        
        \ForAll{$v \in \Call{Neighbors}{G, u}$}
            \State $t_{req} \leftarrow \Call{TimeToReq}{v, E_{rem}}, t_{depot} \leftarrow \Call{TimeToDepot}{v, N_d}$
            \State $p_{cand} \leftarrow [v], t_{proj} \leftarrow t_{cur} + t(u, v), i_{end} \leftarrow |\tau|$
            
            \If{$(u, v) \in E_{rem}$} 
                \State $i_{pivot} \leftarrow \max(i_{ld}, i_{lreq})$
                \State $p_{sp} \leftarrow \Call{ShortestPath}{\tau[i_{pivot}], v}$,  $t_{sp} \leftarrow \Call{TripTime}{p_{sp}}$
                
                \If{$\Call{TripTime}{\tau[0:i_{pivot}]} + t_{sp} < t_{proj}$} \Comment{Check if shortcut is faster}
                    \State $p_{cand} \leftarrow p_{sp}, t_{proj} \leftarrow \Call{TripTime}{\tau[0:i_{pivot}]} + t_{sp}, i_{end} \leftarrow i_{pivot}$
                \EndIf
            \EndIf
            
            \If{$t_{proj} + t_{depot} \leq C$}   
            
                \State $S \leftarrow (1 - w_{cap}) \cdot e^{-t_{req}} + w_{cap} \cdot e^{-t_{depot}/C}$ \Comment{Convex edge scoring function}
                \If{$S > S_{max}$}
                    \State $S_{max} \leftarrow S, p^* \leftarrow p_{cand}, i_{join} \leftarrow i_{end}$
                \EndIf
            \EndIf
        \EndFor
        
        \If{$p^* = \emptyset$} \textbf{break} \EndIf 
        
        \State $\tau \leftarrow \tau[0 : i_{join}] + p^*, t_{cur} \leftarrow \Call{TripTime}{\tau}$ \Comment{Splice path if optimization found}
        \State $v_{new} \leftarrow p^*[\text{end}], e_{new} \leftarrow (p^*[\text{end}-1], v_{new})$
        
        \If{$v_{new} \in N_d$} $i_{ld} \leftarrow |\tau|$ \EndIf
        \If{$e_{new} \in E_{rem}$} $E_{cov} \leftarrow E_{cov} \cup \{e_{new}\}, i_{lreq} \leftarrow |\tau|$ \EndIf
    \EndWhile
    
    \State $\tau \leftarrow \Call{FinishTripToDepot}{\tau, N_d}, u_{end} \leftarrow \tau[\text{end}]$
    \State \Return $\tau, E_{cov}, u_{end}$
\EndProcedure
\end{algorithmic}
\end{algorithm}

The \textsc{ConstructTrip} procedure (Algorithm \ref{const_trip}) builds a single feasible trip $\tau$ starting from a specific depot $u_{start}$ and extending until the vehicle returns to a depot to recharge (Line 2). Throughout the algorithm, we use array indexing notation where $\tau[0]$ denotes the first node in the trip sequence (the starting depot) and $\tau[\text{end}]$ denotes the last node currently in the sequence (the vehicle's current position). To manage the path efficiently, the procedure maintains two history indices: $i_{ld}$, which marks the index of the last visited depot, and $i_{lreq}$, which marks the index of the last serviced required edge. These indices define a safe history that cannot be modified, protecting previously traversed required edges in the constructed trip while the trip length is optimized.

The algorithm iterates until no required edges remain (Line 4). At the beginning of each iteration (Line 6), the algorithm initializes the insertion pointer $i_{join}$ to $|\tau|$, where $|\tau|$ denotes the total number of nodes currently in the trip sequence. Within each iteration, a critical component is the path refinement mechanism (Lines 10--16). If a neighbor $v$ of the current node $u$ is part of a new required edge (Line 10), the algorithm verifies if $v$ can be reached more efficiently via a direct path from a previous safe point. The pivot index $i_{pivot}$ is determined by the maximum of $i_{ld}$ and $i_{lreq}$ (Line 13). If a shortcut from $\tau[i_{pivot}]$ reduces the total trip time, the algorithm tentatively updates the candidate end index $i_{end}$ to this pivot point (Line 14), identifying it as a potential optimization.

However, identifying a shortcut is only the first step. Before any edge is added to the trip $\tau$, the algorithm must strictly enforce operational constraints. A feasibility check (Line 17) ensures that the vehicle can traverse to the candidate neighbor $v$ and subsequently return to the nearest depot without exceeding its total capacity $C$. Only after passing this validation is the edge $(u, v)$ evaluated against other candidates using a convex scoring function $S(u, v)$ (Line 18). If the candidate edge yields the highest score among all neighbors, the algorithm updates the insertion pointer $i_{join}$ to match the candidate $i_{end}$ (Line 20). This deferred update ensures that the router only commits to truncating the inefficient path segment and splicing in the optimized path (Line 26) after confirming that the resulting trip is both physically feasible and locally optimal.

\begin{figure}[b!]
\centering
\includegraphics[width=\textwidth]{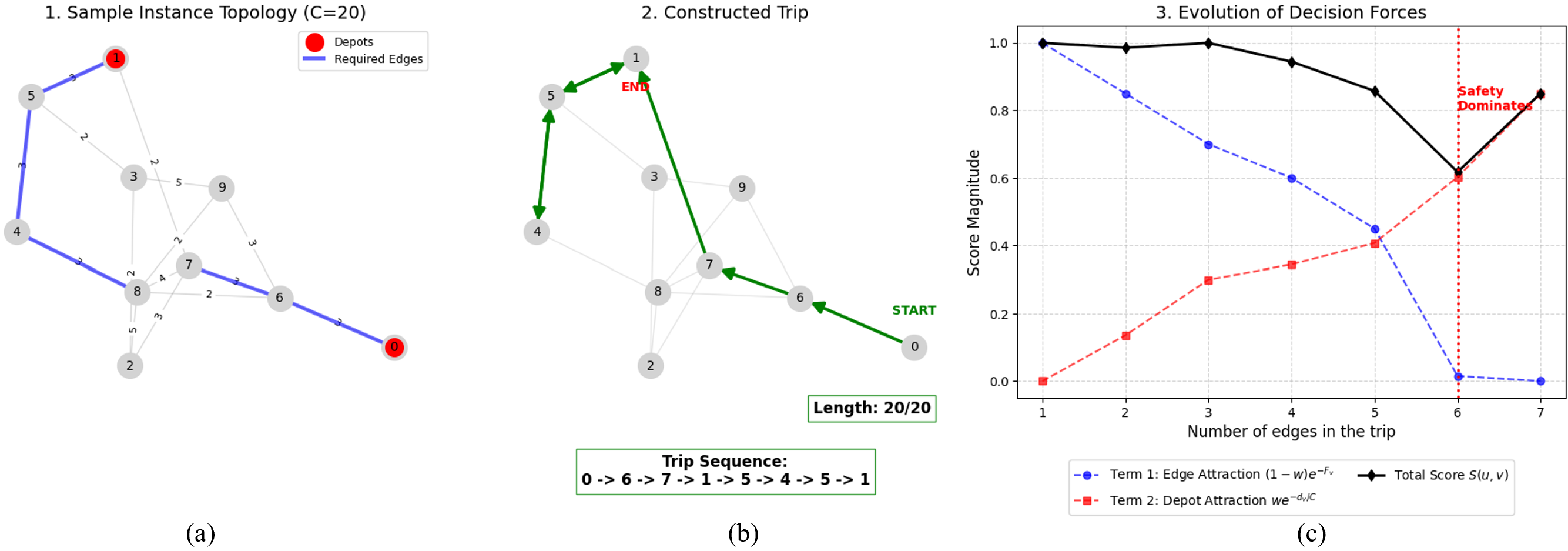}
\caption{Step-by-step construction of a single trip by the Magnetic Field Router. (a) Instance topology showing depots (red) and required edges (blue). (b) The resulting trip path. Note the vehicle visits depot 1 mid-trip but continues to service edges 5-4. (c) Evolution of the decision forces. The vehicle continues servicing edges as long as the Required Edge Attraction (blue) dominates. The trip terminates only when the Depot Attraction (red) overtakes the edge attraction due to depleting capacity.}
\label{fig:router_simulation}
\end{figure}

Before accepting a move, the algorithm performs a feasibility check to ensure the vehicle can reach node $v$ and subsequently return to a depot within capacity $C$. Valid moves are ranked using a convex scoring function $S(u, v)$ (Line 18):
\begin{equation}
    S(u, v) = (1 - w_{cap}) \cdot e^{-t_{req}} + w_{cap} \cdot e^{-t_{depot}/C}
    \label{eq:convex_score}
\end{equation}
In Equation \ref{eq:convex_score}, $t_{req}$ represents the travel time from node $v$ to the nearest unserviced required edge, and $t_{depot}$ represents the travel time to the nearest depot. The weighting factor $w_{cap} = t_{cur} / C$ shifts priority dynamically. Early in the trip, the weight favors the first term to drive exploration. As the battery depletes, the weight shifts to the second term to guide the vehicle toward depots.

Figure \ref{fig:router_simulation} illustrates this behavior on a sample graph instance. The left panel (Figure \ref{fig:router_simulation}a) shows the network topology with depots (red) and required edges (blue). The center panel (Figure \ref{fig:router_simulation}b) depicts the constructed trip. Initially, the vehicle services the required edges (nodes 0-6-7-1). Notice the detour to node 1; although node 1 is a depot, the vehicle does not terminate the trip there. This behavior is explained by the right panel (Figure \ref{fig:router_simulation}c), which tracks the evolution of the decision forces. Around step 5 (visiting node 1), the attraction of required edges (blue line) remains high, indicating nearby unserviced edges, while the attraction from  the depots (red line) is low. Consequently, the router continues the trip to service the remaining edges (nodes 1-5-4-5). As the trip progresses and capacity is consumed, the depot attraction rises sharply, eventually overtaking the edge attraction. This crossover point (marked by the vertical line) forces the vehicle to return to node 1 to terminate the trip safely.

\begin{algorithm}[t!]
\fontsize{11}{14.5}\selectfont
\caption{Finish Trip To Depot}
\label{finish_trip}
\begin{algorithmic}[1]
\Procedure{FinishTripToDepot}{$\tau, N_d, E_{rem}$}
    \State $u \leftarrow \tau[\text{end}], d^* \leftarrow \emptyset, S_{min} \leftarrow \infty, p^* \leftarrow \emptyset$
    
    \ForAll{$d \in N_d$}
        \State $p_{to\_d} \leftarrow \Call{ShortestPath}{u, d}, t_{to\_d} \leftarrow \Call{TripTime}{p_{to\_d}}$
        
        \State $t_{next} \leftarrow \Call{LookAhead}{d, E_{rem}, N_d}$
        \State $S \leftarrow t_{to\_d} + t_{next}$
        
        \If{$S < S_{min}$}
            \State $S_{min} \leftarrow S, d^* \leftarrow d, p^* \leftarrow p_{to\_d}$
        \EndIf
    \EndFor
    
    \If{$d^* = \emptyset$} 
        \State $p^* \leftarrow \Call{ShortestPathToNearestDepot}{u, N_d}$
    \EndIf
    
    \State $\tau \leftarrow \tau + p^*$
    \State \Return $\tau$
\EndProcedure
\end{algorithmic}
\end{algorithm}

The construction loop continues until no feasible neighbors exist ($p^* = \emptyset$, Algorithm \ref{const_trip}, Line 24), indicating the vehicle is operationally boxed in. Finally, the trip is closed using the \textsc{FinishTripToDepot} procedure (Algorithm \ref{finish_trip}). A standard greedy approach would simply route the vehicle to the nearest depot to end the trip. However, such a myopic decision can lead to poor positioning for the subsequent trip to traverse remaining required edges. To address this, the procedure evaluates all reachable depots $d \in N_d$ and selects the destination $d^*$ that minimizes the combined cost $S$:

\begin{equation}
    S = t_{to\_d} + \Call{LookAhead}{d, E_{rem}, N_d}
\end{equation}
Here, $t_{to\_d}$ is the travel time to reach depot $d$, and the look-ahead term estimates the cost to reach the nearest unserviced required edge in the next trip starting from $d$. By minimizing this sum, the algorithm balances immediate efficiency with future strategic positioning.




\subsection{Benchmark Strategy: Reactive Simulated Annealing}
\label{sec:benchmark_sa}

\begin{figure}[t!]
\centering
\includegraphics[width=0.6\textwidth]{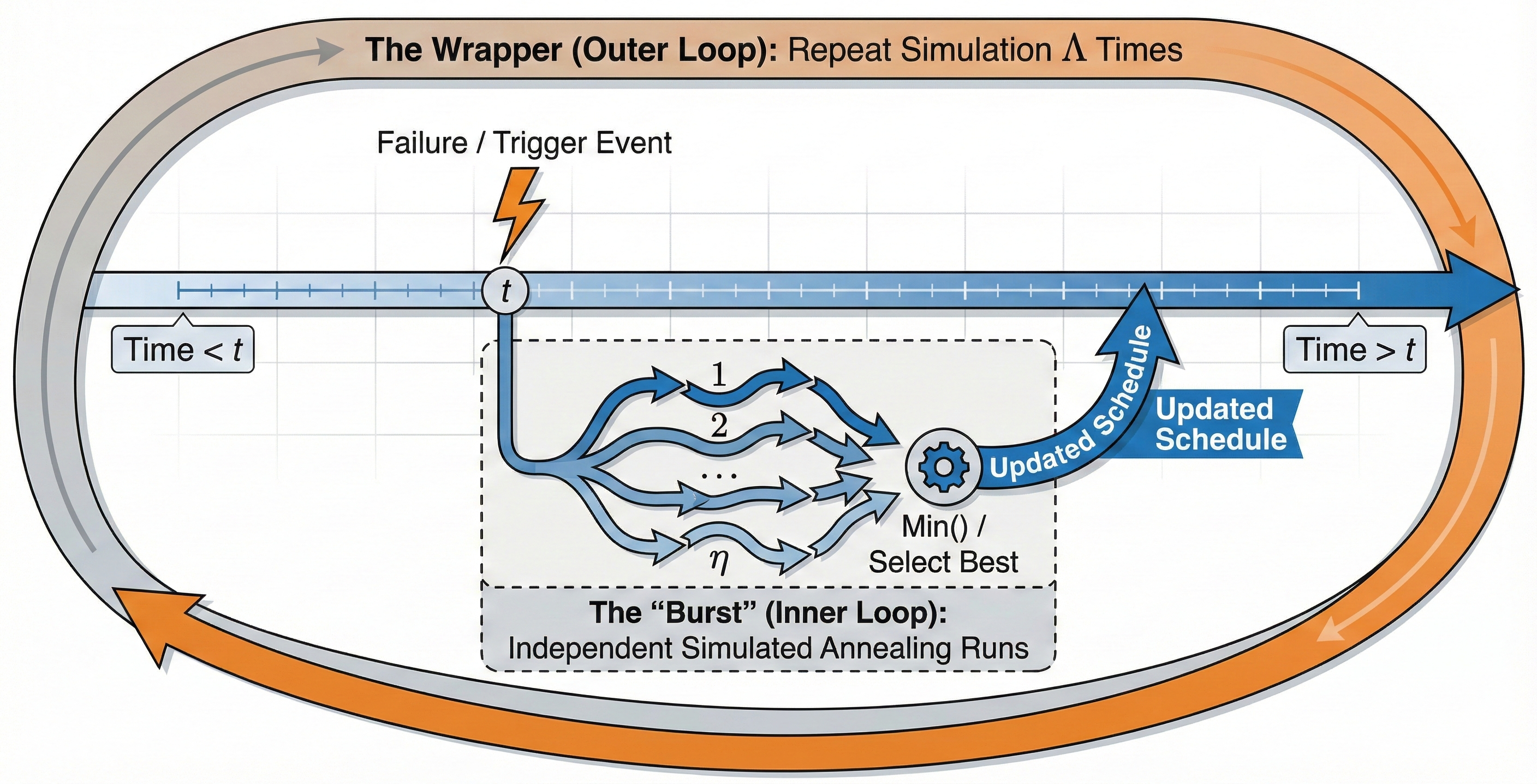}
\caption{Schematic of the Reactive Simulated Annealing benchmark. A failure event triggers a burst of $\eta$ independent re-optimization runs to find the best immediate response. The entire simulation is wrapped in an outer loop of $\Lambda$ replications to account for scenario variance.}
\label{fig:sa_schematic}
\end{figure}

To rigorously evaluate the proposed two-stage reactive framework, we employ a reactive variant of the Simulated Annealing (SA) metaheuristic. This choice is based on comparative studies on the static MD-RPP-RRV \citep{md-rpp-rrv}, where SA demonstrated a superior balance between solution quality and computational efficiency compared to population-based methods like genetic algorithms. While alternative metaheuristics can achieve high solution quality, their computational overhead renders them unsuitable for the time-critical requirements of dynamic rescheduling. Consequently, SA serves as the requisite state-of-the-art baseline for real-time mission recovery, treating every vehicle failure as a trigger for a global re-optimization of the remaining mission. The structure of this reactive protocol is visualized in Figure \ref{fig:sa_schematic}.

The procedure operates within the simulation environment described in Algorithm \ref{simulation}. When a vehicle failure is detected at time $f_k$ (indicated as the trigger event $t$ in Figure \ref{fig:sa_schematic}), the system halts operations. The current state of the fleet, including the locations of active vehicles and unserviced required edges, is extracted to define a new static routing problem. The SA metaheuristic is then invoked to solve this residual problem.

To mitigate the inherent stochasticity of the annealing process, the re-optimization utilizes a burst strategy shown in the inner loop of Figure \ref{fig:sa_schematic}. At each failure, the algorithm executes $\eta$ independent optimization trials. The solution yielding the minimum mission time among these $\eta$ trials is selected as the updated fleet schedule. Furthermore, to account for variance across different failure sequences, the entire simulation from start to finish is wrapped in an outer loop and replicated $\Lambda$ times. This dual-layer redundancy ensures that the reported results reflect the robust capability of the SA approach rather than statistical outliers.

\section{Results}
\label{sec:results}
This section describes the benchmark instances adapted from the literature for failure scenario creation, details the experimental design used to evaluate the proposed reactive framework, and presents a comparative analysis of its performance against the reactive simulated annealing metaheuristic described in Section \ref{sec:benchmark_sa}.

\subsection{Instance Generation Process}
\label{icrin}
We generated 77 MD-RPP-RRV instances derived from standard CARP benchmarks: gdb \citep{gdb}, bccm \citep{bccm}, and eglese \citep{eglese1, eglese2}. For each instance, we randomly designated half of the edges as required edges. We selected a subset of nodes as depots such that every depot lies within a capacity-feasible shortest-path distance $C$ of at least one other depot. The process assigns exactly one vehicle to each depot. To standardize the parameters, we set the vehicle capacity $C$ to twice the maximum edge weight and the recharge time $R_T$ to twice the vehicle capacity.

    
    

\begin{algorithm}[b!]
\fontsize{11}{13.5}\selectfont
\caption{Create Failure Scenarios}
\label{pseudo:failure_scenario_creation}
\begin{algorithmic}[1]
\Procedure{CFS}{$G, N_d, K, n_K, y_K, E_u, R_T, C, F_{max}$}
    \State $S \leftarrow \emptyset, \quad N_{max} \leftarrow \Call{UniformInt}{1, \min(F_{max}, K-1)}$
    \For{$j = 1 \to N_{max}$}
        \State $F \leftarrow \Call{RandomSubset}{\{1, \dots, K\}, j}$ \Comment{Select $j$ distinct vehicles}
        \State $P_K, y_K \leftarrow \Call{SA}{G, N_d, K, n_K, y_K, E_u, R_T, C}$
        \State $f_k \leftarrow \{ k \mapsto \Call{UniformInt}{1, y_k} \mid k \in F \}$ \Comment{Assign failure times}
        \State $S \leftarrow S \cup \{(F, f_k)\}$
    \EndFor
    \State \Return $S$
\EndProcedure
\end{algorithmic}
\end{algorithm}

\subsubsection{Failure Scenario Creation}
\label{failure_scenario_creation}
We utilized the \textsc{CFS} procedure (Algorithm \ref{pseudo:failure_scenario_creation}) to generate 257 distinct failure scenarios. Each scenario is defined by three components: the number of vehicle failures $|F|$, the specific set of failed vehicles $F$, and their respective failure times $f_k$. To ensure mission feasibility while simulating realistic high-impact disruptions, the number of failures is constrained to a maximum of $\min(F_{max}, K-1)$. This upper bound $F_{max} = 6$ prevents the generation of statistically improbable total fleet collapse scenarios in large instances.

The creation process begins by selecting a random integer number of failures (Line 2). The procedure then iteratively samples distinct vehicles to populate the failure set $F$ (Line 4). To determine realistic failure times, we first execute the simulated annealing baseline to obtain the scheduled maximum trip time $y_k$ for each vehicle (Line 5). The specific failure time $f_k$ for a vehicle $k \in F$ is then drawn uniformly from the interval $[1, y_k]$ (Line 6). Table \ref{failure_scenario} summarizes the characteristics of the generated failure scenarios. The following subsection provides the experimentation details.

\begin{table}[t!]
\centering
\renewcommand{\arraystretch}{1.1}
\caption{Failure Scenarios Information}
\label{failure_scenario}
\resizebox{0.95\textwidth}{!}{
\begin{tabular}{ccccccccc}
\hline
\textbf{Instance} & \textbf{Number of} & \textbf{Nodes} & \textbf{Edges} & \textbf{Number of} & \textbf{Number of} & \textbf{\% Average of} & \textbf{\% Average} & \textbf{Failures} \\
\textbf{Name} & \textbf{failure} & \textbf{range} & \textbf{Range} & \textbf{depots} & \textbf{vehicles} & \textbf{required edges} & \textbf{of depots} & \textbf{range} \\
& \textbf{scenarios} & & & \textbf{range} & \textbf{range} & \textbf{in edges} & \textbf{in nodes} & \\
\hline
GDB & 37 & 7--13 & 19--44 & 2--5 & 2--5 & 50.6\% & 25\% & 1--4 \\
BCCM & 108 & 24--50 & 24--97 & 4--10 & 4--10 & 50.5\% & 18\% & 1--6 \\
EGLESE & 112 & 77--140 & 98--190 & 9--12 & 9--12 & 50\% & 11\% & 1--6 \\
\hline
\end{tabular}}
\end{table}

\subsection{Experimentation}
This section describes the experiments conducted to evaluate the proposed two-stage reactive framework. We assess the performance of our approach by comparing its solution quality and computational efficiency against the reactive SA metaheuristic baseline described in Section \ref{sec:benchmark_sa}. The experiments were performed on an AMD EPYC 7763 64-core Processor with 128 physical cores, 128 logical processors, and 8 CPU cores. Up to 32 threads were utilized, and 8 GB of memory was allocated to each CPU core. The instances and failure scenarios can be accessed using the GitHub repository from \href{https://github.com/Eashwar-S/Rescheduling-multi-depot-multi-trip-rural-postman-problem-instances}{{\color{blue}here}}.

Table \ref{tab:parameters} summarizes the algorithmic settings used in this evaluation. The specific parameter values for the peer auction were established via the sensitivity analysis in Section \ref{sensitivity_analysis_sec}, while the simulated annealing parameters follow the optimal configuration reported by \citet{md-rpp-rrv}.

\begin{table}[b!]
    \centering
    \small
    \caption{Configuration of tunable algorithmic parameters}
    \label{tab:parameters}
    \setlength{\tabcolsep}{6pt} 
    \begin{tabular}{llc}
        \toprule
        \textbf{Algorithm} & \textbf{Parameter (Symbol)} & \textbf{Value} \\
        \midrule
        CA & Initial Radius ($r_i$), Step ($\Delta r$) & $C$ \\
        \midrule
        PA & Window Size ($W$) & 2 \\
                     & Comp. Budget ($L$) & 20 \\
                     & Max Rounds ($R$) & 10 \\
        \midrule
        SA & Trials ($\eta$), Simulations ($\Lambda$) & 10 \\
                            & Cooling Rate ($\alpha$) & 0.99 \\
                            & Iterations & 1000 \\
        \bottomrule
    \end{tabular}
\end{table}

\subsubsection{GDB Failure Scenario Results}
\label{sec:gdb_results}

\begin{figure}[b!]
\centering
\includegraphics[width=0.75\textwidth]{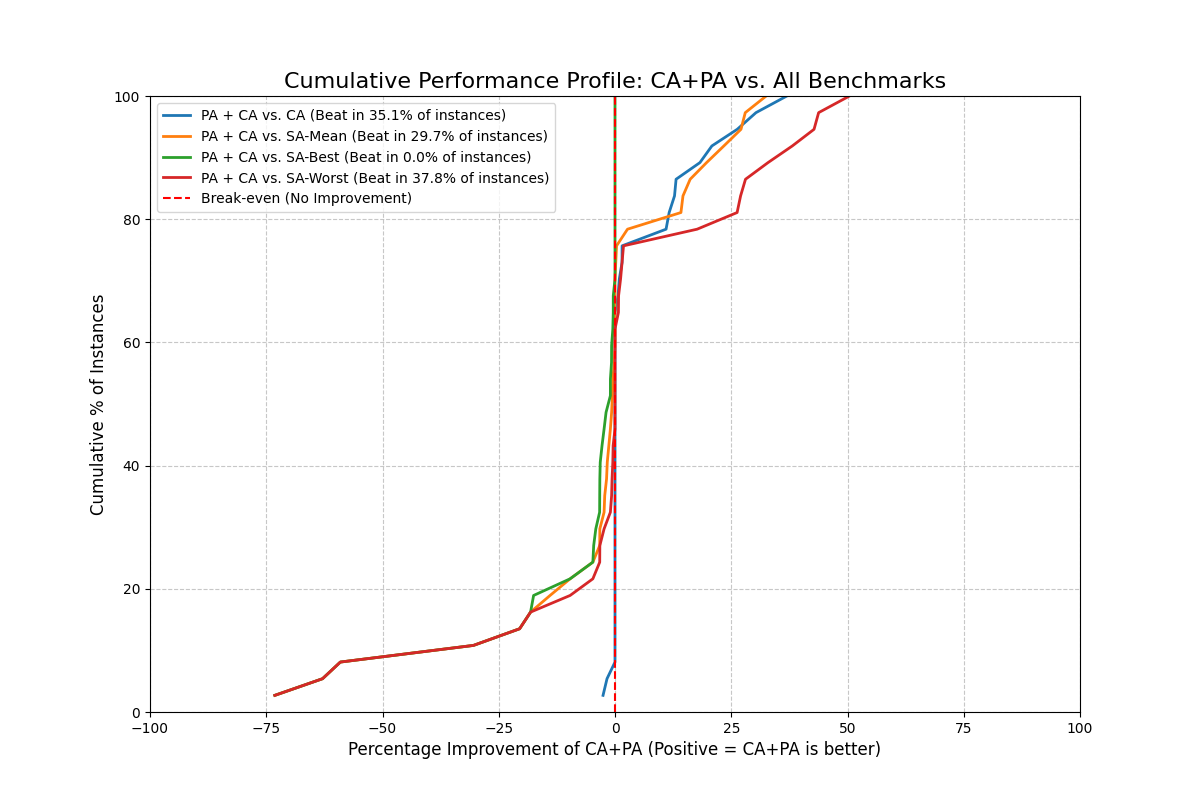}
\caption{Cumulative distribution of the performance gap for GDB instances. The plot illustrates the improvement of the two-stage framework (CA+PA) over the baseline CA and its competitiveness against the SA benchmark distribution.}
\label{fig:gdb_cdf}
\end{figure}

This section analyzes the performance of the proposed framework on the GDB dataset, which comprises 37 failure scenarios generated from 19 small-scale instances. The comprehensive tabular results for all scenarios are provided in Table S2 of the Supplementary Materials.

We first evaluate the solution quality (lower mission time) of the two-stage framework (CA+PA) relative to the centralized auction (CA) and the simulated annealing (SA) benchmark. Figure \ref{fig:gdb_cdf} presents the cumulative distribution of the performance gap. The proposed peer auction refinement successfully improved the solution quality over the centralized baseline in 13 out of 37 instances (35.1\%). While the average improvement across the entire dataset was 4.91\%, the magnitude of improvement in those specific instances where the peer auction was effective was significant, averaging 14.30\%. When compared against the SA benchmark, the CA+PA framework outperformed the SA mean result in 11 instances (29.7\%) and the SA worst-case result in 14 instances (37.8\%). However, given the small size of the GDB instances, the metaheuristic was able to converge to high-quality solutions, and the proposed heuristic framework did not surpass the SA best-found solution in any scenario.

\begin{figure}[t!]
\centering
\includegraphics[width=\textwidth]{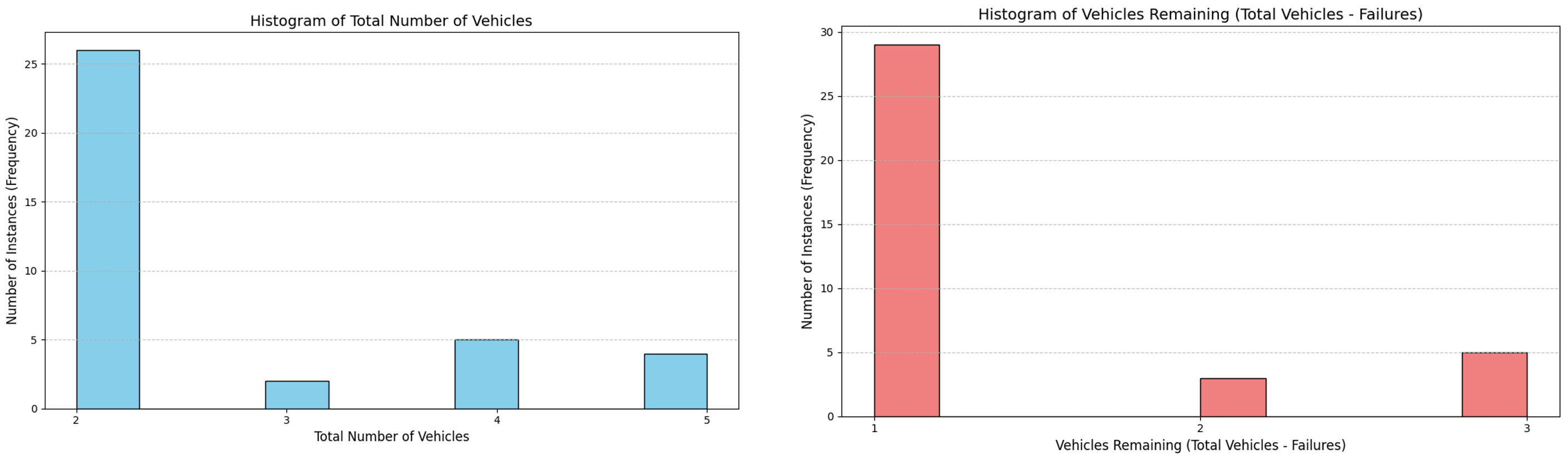}
\caption{Performance discrepancy analysis for GDB instances. The lack of improvement in many scenarios correlates with instances where only a single vehicle remained active, precluding the use of peer negotiation.}
\label{fig:gdb_discrepancy}
\end{figure}

The inability of the peer auction to improve the centralized solution in the remaining 65\% of cases is explained by the fleet characteristics of the GDB dataset. As shown in Figure \ref{fig:gdb_discrepancy}, a large portion of these scenarios resulted in a single active vehicle remaining after failures. The peer auction mechanism relies on negotiation between at least two vehicles to exchange tasks and balance workloads. As outlined in the framework overview (Figure \ref{fig:framework_flowchart}), when the fleet is reduced to a single agent, the peer negotiation phase is bypassed, and the system relies solely on the Magnetic Field Router to locally optimize the single route. In these small, tightly constrained topologies, the greedy insertion performed by the centralized auction often constructs a route that leaves minimal room for further single-vehicle optimization, resulting in no net improvement for those specific instances.

\begin{figure}[b!]
\centering
\includegraphics[width=0.6\textwidth]{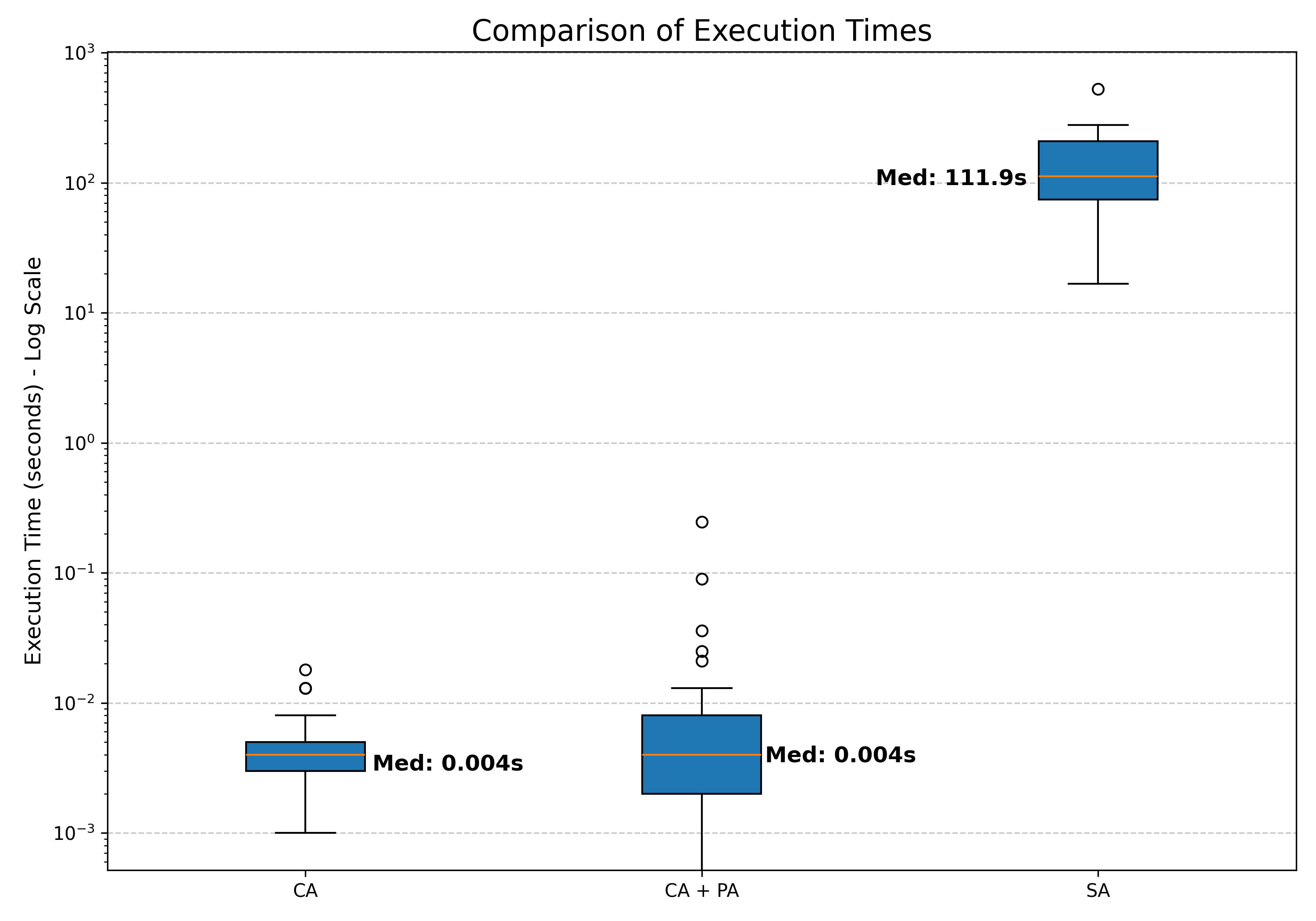}
\caption{Log-scale boxplot comparison of execution times for GDB scenarios. The proposed CA+PA framework delivers solutions orders of magnitude faster than the SA benchmark, with median times in the millisecond range.}
\label{fig:gdb_time}
\end{figure}

While the solution quality on small instances is comparable to the metaheuristic baseline, the proposed framework demonstrates a decisive advantage in computational efficiency. Figure \ref{fig:gdb_time} compares the execution times of the three approaches on a logarithmic scale. The reactive SA benchmark required an average of 142.80 seconds (median 111.9 seconds) to re-optimize a scenario. In contrast, the complete two-stage CA+PA framework required an average of only 0.02 seconds (median 0.004 seconds). The centralized auction alone is even faster, averaging 0.0051 seconds. This represents a speedup factor of approximately 7000 times, validating the framework's suitability for real-time responsiveness.

\begin{figure}[t!]
\centering
\includegraphics[width=0.75\textwidth]{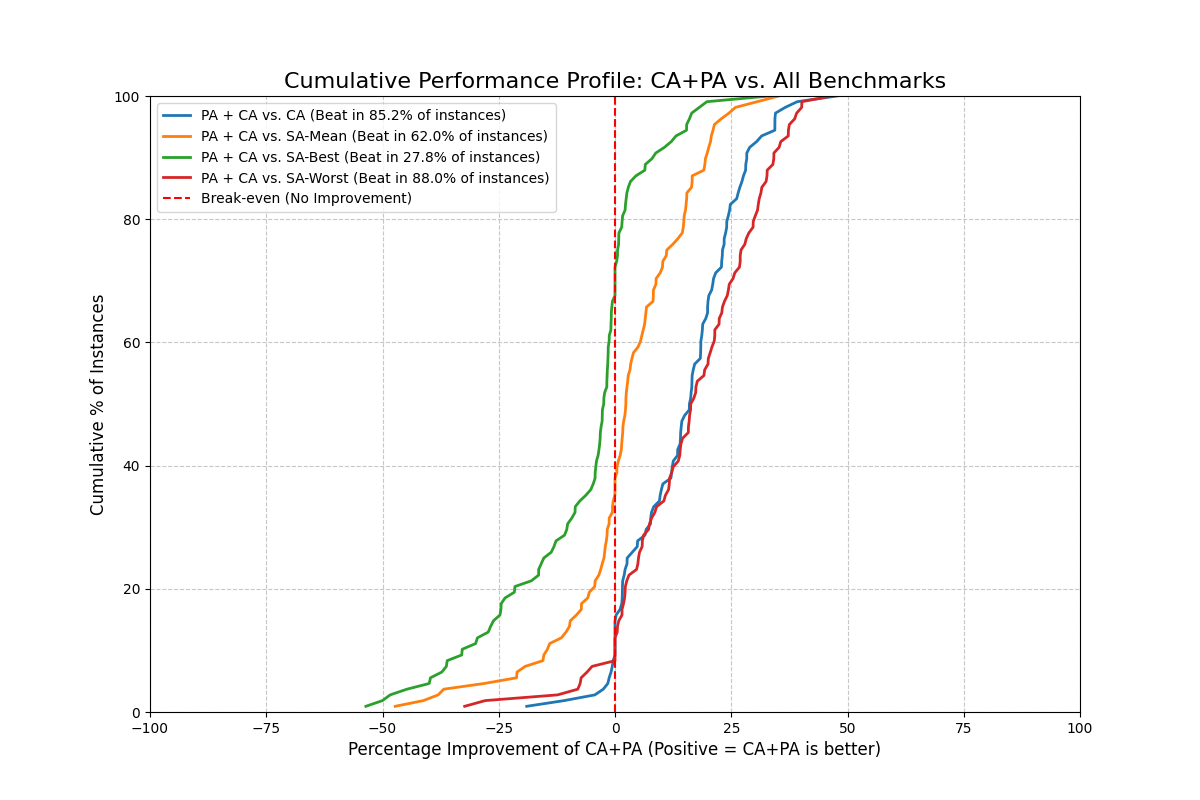}
\caption{Cumulative distribution of the performance gap for BCCM instances. The high percentage of improved instances (85.2\%) demonstrates the effectiveness of the peer auction in medium-scale scenarios compared to the GDB dataset.}
\label{fig:bccm_cdf}
\end{figure}

In summary, the GDB results establish that the proposed framework is exceptionally fast and capable of improving greedy solutions when fleet interactions are possible. However, the small scale of these instances limits the potential for complex route improvements. The computational advantages and solution quality gains become more pronounced in larger, more complex scenarios, as discussed in the following subsections on the BCCM and EGLESE datasets.

\subsubsection{BCCM Failure Scenario Results}
\label{sec:bccm_results}

This section evaluates the proposed framework on the BCCM dataset, which represents medium-scale complexity with 108 failure scenarios generated from 34 instances. Unlike the smaller GDB instances, these scenarios typically involve 4 to 10 vehicles and a larger number of depots (Table \ref{failure_scenario}), creating a richer solution space for the peer auction's negotiation mechanisms. The detailed results for all scenarios are listed in Tables S3 and S4 of the Supplementary Materials.

We first analyze the solution quality improvement of the two-stage framework (CA+PA) compared to the baselines. Figure \ref{fig:bccm_cdf} displays the cumulative distribution of the performance gap. In stark contrast to the GDB results, the peer auction refinement demonstrated high effectiveness, improving upon the centralized auction's baseline in 92 out of 108 instances (85.2\%). The magnitude of this improvement is substantial: the average reduction in mission time across all instances was 14.66\%, rising to 17.66\% for the subset of improved instances. This confirms that as problem scale and fleet size increase, the local repair and negotiation mechanisms of the peer auction become critical for untangling the suboptimal assignments made by the greedy centralized stage.

\begin{figure}[b!]
\centering
\includegraphics[width=0.6\textwidth]{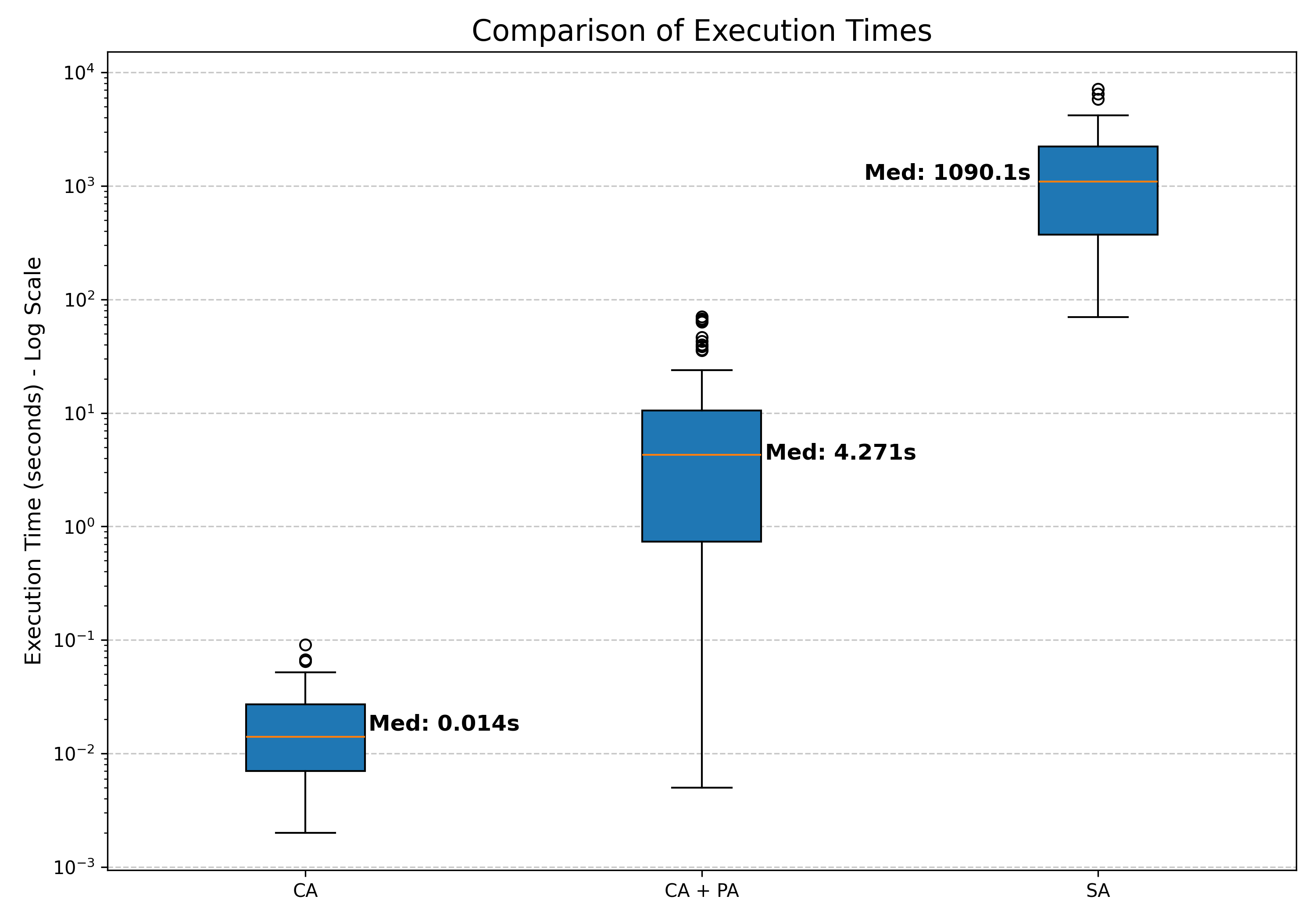}
\caption{Log-scale boxplot comparison of execution times for BCCM scenarios. While the peer auction adds computational cost compared to the centralized baseline (median 4.27s vs 0.014s), it remains orders of magnitude faster than the SA benchmark (median 1090s).}
\label{fig:bccm_time}
\end{figure}

The framework also shows strong competitiveness against the reactive SA benchmark. The CA+PA solution outperformed the SA mean result in 67 instances (62.0\%) and the SA worst-case result in 95 instances (88.0\%), with an average improvement of 16.25\% over the latter. While the metaheuristic's best-found solution (from $\eta$ trials) still held the edge in the majority of cases, the proposed framework beat the SA-Best in 30 instances (27.8\%). This indicates that for medium-scale problems, the heuristic repair is often capable of finding solutions that rival computationally intensive metaheuristics.

The trade-off for this increased solution quality is a moderate increase in computational effort compared to the centralized auction alone. Figure \ref{fig:bccm_time} illustrates the execution times on a logarithmic scale. The addition of the peer auction phase increased the average execution time from 0.0189 seconds (CA only) to 9.85 seconds (CA+PA). However, the median execution time for the full framework remains low at 4.27 seconds. In comparison, the reactive SA benchmark is computationally prohibitive for real-time applications, with an average execution time of 1418.54 seconds (approx. 24 minutes) and a median of 1090.1 seconds.

In summary, the BCCM results highlight the scalability of the proposed framework. While the peer auction adds a computational cost of several seconds, it delivers a 14\% average improvement in solution quality over the greedy baseline and offers a viable real-time alternative to metaheuristics, which require orders of magnitude more time to achieve comparable results.

\subsubsection{EGLESE Failure Scenario Results}
\label{sec:eglese_results}

This section analyzes the performance of the proposed framework on the EGLESE dataset, which comprises 112 failure scenarios generated from 24 large-scale instances. These scenarios involve networks with up to 140 nodes and 190 edges, serviced by fleets of 9 to 12 vehicles (Table \ref{failure_scenario}). This dataset represents the most complex operational environment tested, pushing the limits of coordination and routing efficiency. The complete tabular results for all scenarios are provided in Tables S5 and S6 of the Supplementary Materials.

\begin{figure}[t!]
\centering
\includegraphics[width=0.75\textwidth]{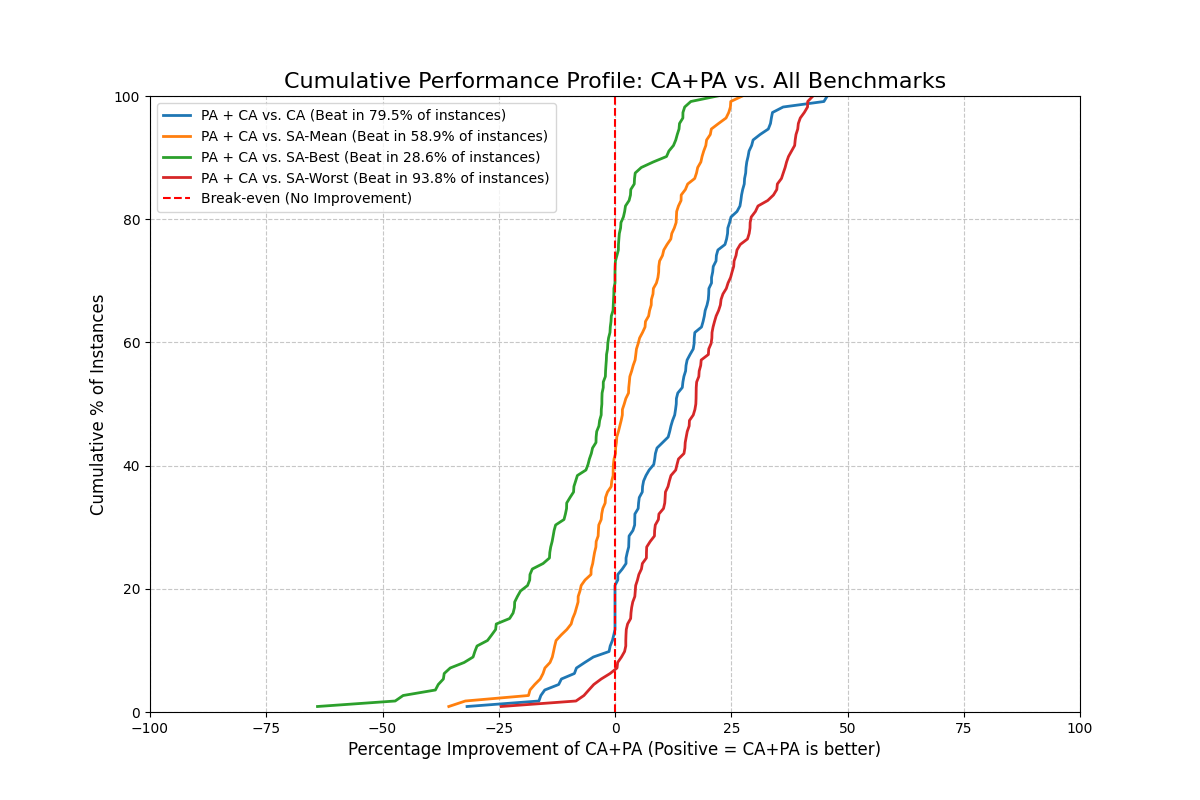}
\caption{Cumulative distribution of the performance gap for EGLESE instances. The framework demonstrates robust performance on large-scale networks, improving upon the centralized baseline in nearly 80\% of scenarios.}
\label{fig:eglese_cdf}
\end{figure}

Evaluating solution quality, the two-stage framework (CA+PA) continues to demonstrate robust performance on these large networks. Figure \ref{fig:eglese_cdf} illustrates the cumulative distribution of the performance gap. Similar to the BCCM results, the peer auction refinement proved highly effective, improving upon the centralized auction's baseline in 89 out of 112 instances (79.5\%). The average improvement across all instances was 12.64\%, increasing to 17.42\% for the subset of improved instances. This consistent performance on large-scale networks confirms the robustness of the magnetic field router and negotiation protocols in handling complex, high-density routing problems.

\begin{figure}[t!]
\centering
\includegraphics[width=0.6\textwidth]{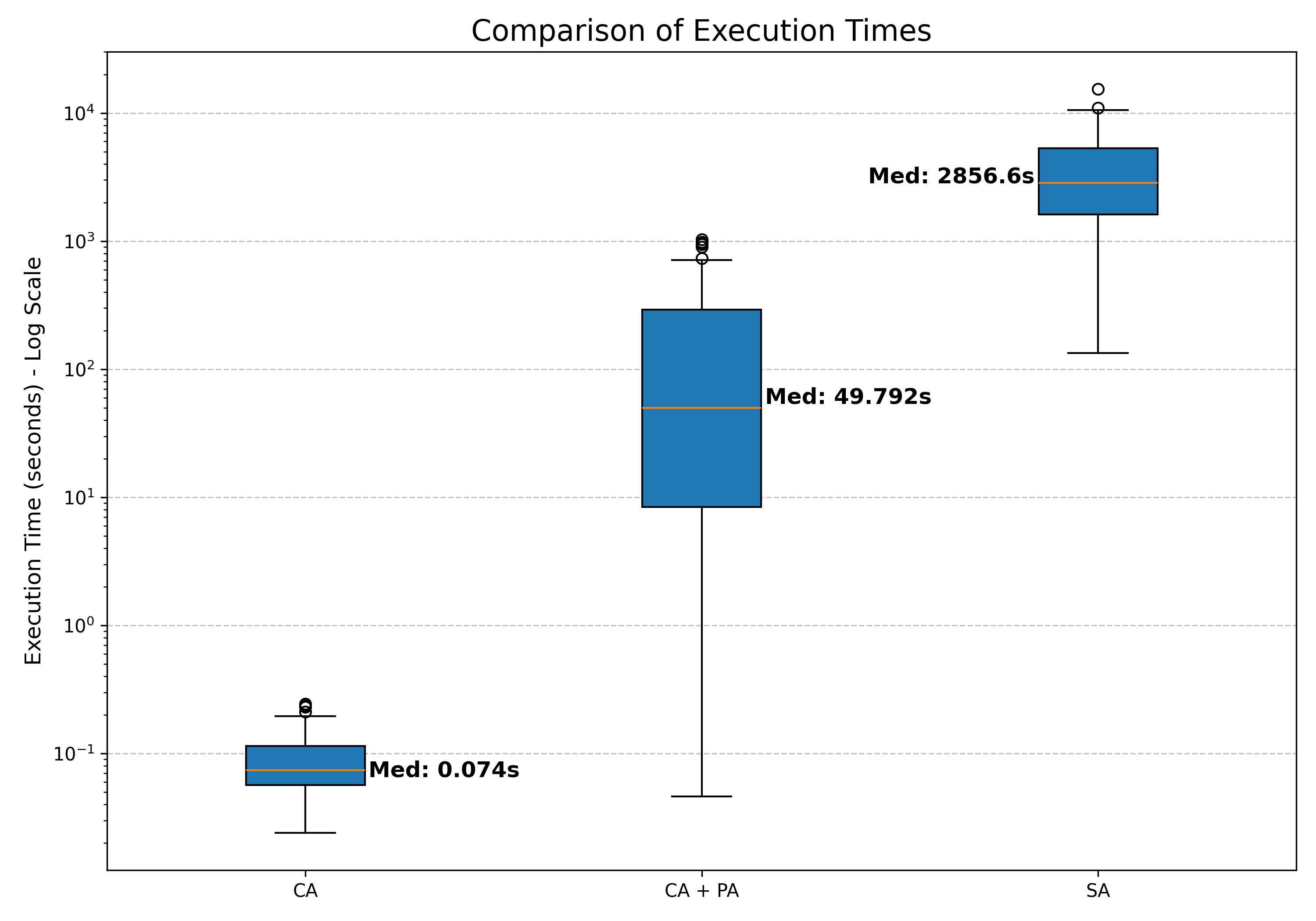}
\caption{Log-scale boxplot comparison of execution times for EGLESE scenarios. The CA+PA framework delivers high-quality solutions in minutes (median 49.79s), whereas the SA benchmark requires nearly an hour (median 2856.6s), highlighting the scalability of the proposed approach.}
\label{fig:eglese_time}
\end{figure}

The comparison against the reactive SA benchmark further underscores the framework's capability. The CA+PA solution outperformed the SA mean result in 66 instances (58.9\%) and the SA worst-case result in 105 instances (93.8\%), with an average improvement of 17.45\% over the latter. Even against the SA best-found solution (the minimum of $\eta$ trials), the heuristic framework achieved a lower mission time in 32 instances (28.6\%). These results indicate that for large-scale logistics, the proposed auction-based repair is not only feasible but often superior to metaheuristics that may struggle to converge within practical time limits.

The computational advantage of the proposed framework is most evident on these large instances. Figure \ref{fig:eglese_time} compares the execution times on a logarithmic scale. While the peer auction increases the computational load compared to the centralized baseline (raising the average time from 0.09 seconds to 188.37 seconds), it remains a viable option for operations where a decision is needed within minutes. In contrast, the reactive SA benchmark becomes effectively unusable for real-time recovery, with an average execution time of 3754.57 seconds (over an hour) and a median of 2856.6 seconds. The two-stage framework provides a critical balance, delivering solution quality comparable to hour-long metaheuristic runs in under a few minutes.

\subsection{Sensitivity analysis}
\label{sensitivity_analysis_sec}
To ensure the proposed peer auction framework generalizes across diverse topologies, we performed a rigorous sensitivity analysis on a calibration set comprising six representative instances from the GDB, BCCM, and EGLESE datasets. We conducted a full grid search over the core algorithm parameters: window size $W \in \{1, 2, 3, 4, 5\}$ and computational budget $L \in \{5, 10, 20, 40, 60\}$, resulting in 25 distinct parameter combinations per instance. For each run, the maximum rounds were bounded at $R=20$; however, the auction terminates early if no improvement in solution quality is observed (Algorithm \ref{peer_auction_pseudo}). We recorded the final round count to determine a sufficient convergence limit for the general case.

To evaluate the trade-offs, we computed relative performance metrics for each parameter pair $(W, L)$. The average best solution gap ($\Delta_{gap}$) measures the deviation of the current solution quality $S_{curr}$ from the best solution $S_{best}$ found across all 25 combinations, calculated as:
\begin{equation}
    \Delta_{gap} = \frac{S_{curr} - S_{best}}{S_{best}} \times 100
\end{equation}
Similarly, the normalized execution time ($T_{norm}$) compares the current runtime $E_{curr}$ against the fastest execution time $E_{min}$ observed in the set:
\begin{equation}
    T_{norm} = \frac{E_{curr}}{E_{min}} \geq 1
\end{equation}

\begin{figure}[b!]
\centering
\includegraphics[width=\textwidth]{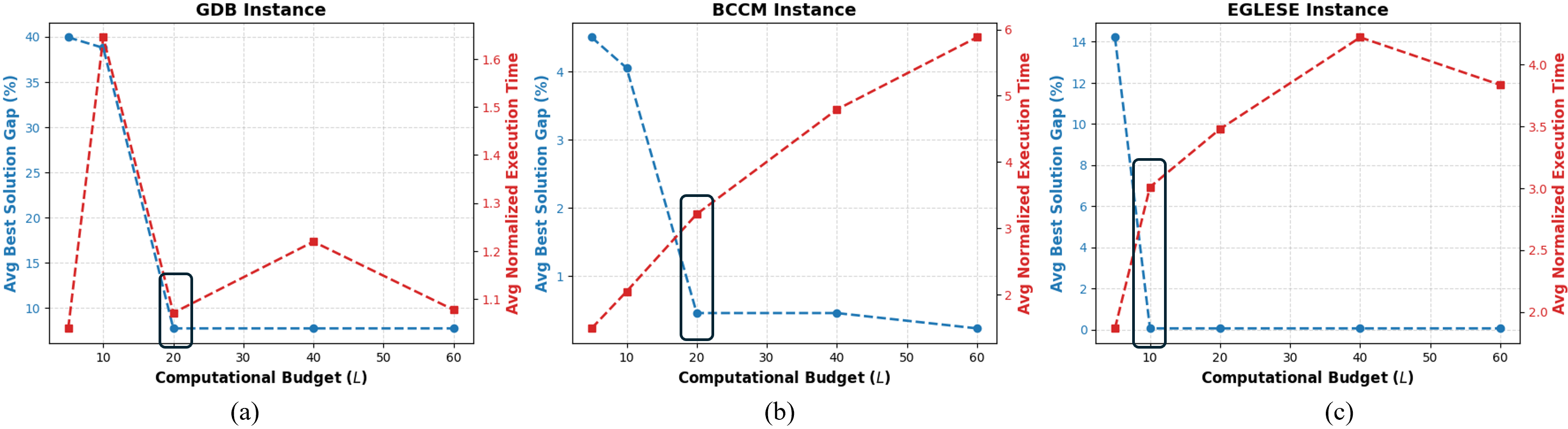}
\caption{Sensitivity analysis of computational budget $L$ on instances (a) \texttt{gdb.13}, (b) \texttt{bccm.105}, and (c) \texttt{eglese.10}. Plots display the trade-off between best solution gap and normalized execution time, averaged across window sizes $W \in [1,5]$. Rectangles in black mark the saturation point ($L=10$ or $20$) where solution quality stabilizes, indicating diminishing returns for higher budgets.}
\label{fig:senstivity_L}
\end{figure}

\begin{figure}[t!]
\centering
\includegraphics[width=\textwidth]{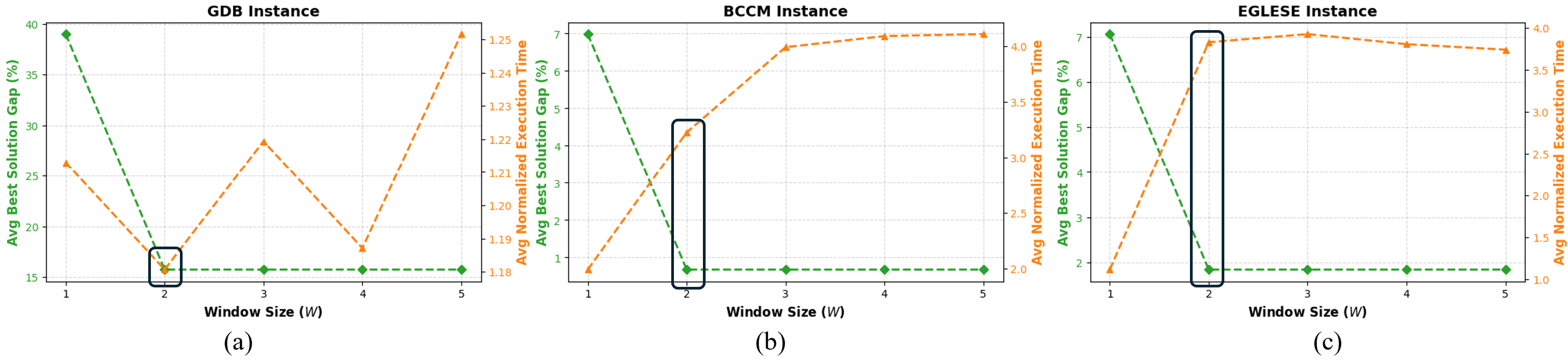}
\caption{Impact of maximum window size $W$ on performance for instances (a) \texttt{gdb.13}, (b) \texttt{bccm.105}, and (c) \texttt{eglese.10}, averaged across budgets $L \in [5,60]$. The boxed selection $W=2$ identifies the critical trade-off point: it secures the primary reduction in best soltuion gap (dotted green) with a manageable increase in execution time compared to $W=1$, whereas $W \ge 3$ incurs increased computational costs for negligible further gains on solution quality.}
\label{fig:senstivity_W}
\end{figure}

In our analysis, values are averaged to isolate the marginal effect of each parameter. For example, when analyzing $W$, the reported metrics are averages over all five values of $L$, and conversely for the analysis of $L$. We performed this sensitivity analysis on six instances, comprising two randomly selected from the upper and lower halves of the GDB, BCCM, and EGLESE datasets in order to examine the consistency of observed trends as instance size increases (Supplementary Material Tables S7 (GDB), S8 (BCCM), and S9 (EGLESE)). For clarity, Figures \ref{fig:senstivity_L} and \ref{fig:senstivity_W} display results for a single instance per dataset, as the trends were consistent in all tested cases.

Figure \ref{fig:senstivity_L} illustrates the sensitivity of the average best solution gap to the computational budget $L$. The results show that the  gap stabilizes at $L=20$ for the GDB and BCCM datasets and $L=10$ for EGLESE. Increasing the budget beyond these saturation points yields diminishing returns, increasing execution time without improving the solution quality.

Figure \ref{fig:senstivity_W} demonstrates the impact of the window size $W$. A critical trade-off occurs when shifting from $W=1$ to $W=2$, which provides the primary reduction in the average best solution gap. While $W \ge 3$ offers theoretical improvements, the normalized execution time grows disproportionately. Consequently, we selected $W=2$ as our choice. Finally, regarding convergence, empirical data indicated that the auction consistently stabilizes within 3 to 6 rounds. To provide a conservative safety margin, we set the final parameter configuration to $\{W=2, L=20, R=10\}$ for the full benchmark suite.

\subsection{Effectiveness of Magnetic Field Router}
\label{sec:magnetic_effectiveness}

This section evaluates the standalone contribution of the magnetic field router in improving solution quality. Although the router functions as the underlying local search mechanism within the peer auction for multi-vehicle fleets, quantifying its individual impact is difficult during cooperative exchanges. Therefore, we focus on specific instances from the GDB and BCCM datasets (Multiple vehicles remain in EGLESE instances in all failure scenarios, so not considered) where only a single vehicle remains active. In these scenarios, the peer auction phase is operationally bypassed as it requires at least two vehicles (Figure \ref{fig:framework_flowchart}). This isolation allows us to strictly measure the router's ability to optimize  baseline routes generated by the centralized auction without the confounding effects of peer-to-peer trading, as initially discussed in Section \ref{sec:gdb_results}.

To measure effectiveness, we compare the maximum trip time of the route optimized by the magnetic field router ($T_{MFR}$) against the baseline trip time returned by the centralized auction ($T_{base}$). The percentage improvement is defined as:
\begin{equation}
    \text{Improvement (\%)} = \frac{T_{base} - T_{MFR}}{T_{base}} \times 100
\end{equation}

\begin{figure}[b!]
    \centering
    \includegraphics[width=\textwidth]{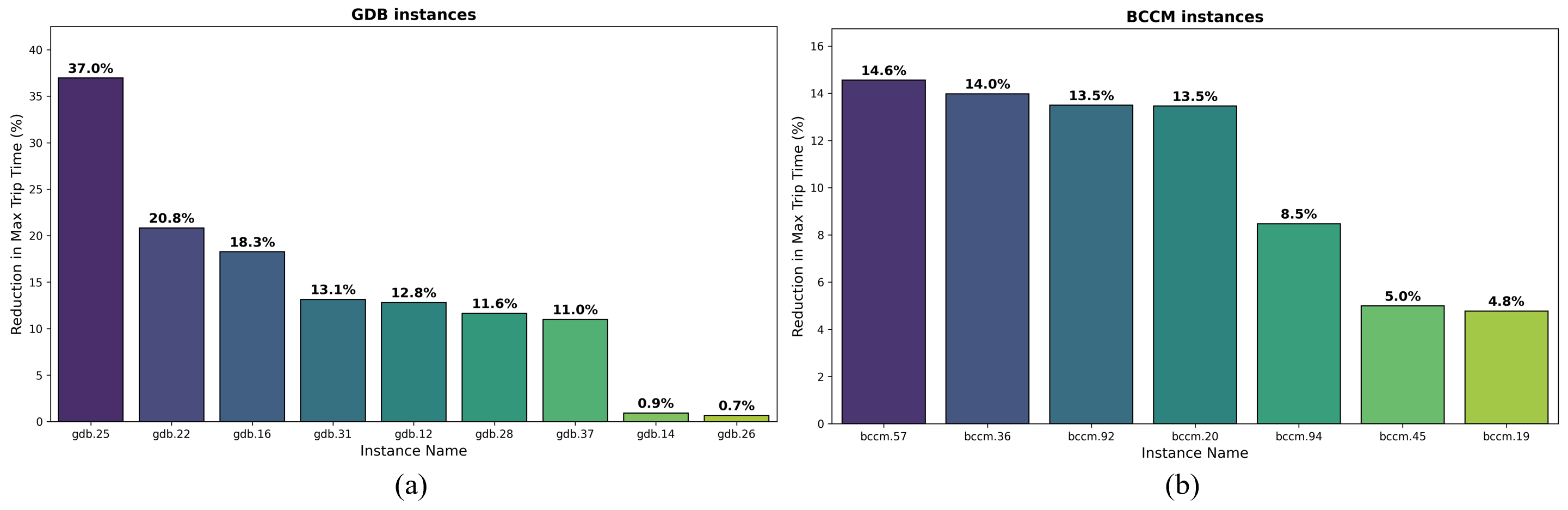}
    \caption{Percentage reduction in maximum trip time achieved by the magnetic field router in single-vehicle scenarios for (a) GDB instances and (b) BCCM instances. The plots highlight specific cases where the router successfully optimized the baseline centralized route, recovering significant performance gains without peer auction intervention.}
    \label{fig:magnetic_effectiveness}
\end{figure}

Figure \ref{fig:magnetic_effectiveness} presents the improvement metrics for instances where the router successfully optimized the route. For the GDB dataset (29 single-vehicle instances), the router improved the solution in 9 cases. In these active instances, it achieved an average improvement of 14.02\%, with a maximum reduction in maximum trip time of 36.96\%. Similarly, for the BCCM dataset (21 single-vehicle instances), the router improved 7 cases, yielding an average improvement of 10.53\% and a maximum of 14.55\%.

These results demonstrate that, although the centralized auction provides a sufficient solution in simple topologies, the magnetic field router is essential for local repair in more complex scenarios. By constructing the route through its convex scoring function, the router proves effective in recovering significant performance gains that would otherwise be lost in the absence of a multi-vehicle peer auction.

\subsection{Theoretical Performance Bound}
\label{sec:theoretical_ratio}

We rigorously assessed the efficiency of the proposed reactive framework by analyzing the performance deviation of the centralized auction relative to an offline optimal solution. 
We focused on the centralized auction because it establishes the foundational feasible solution with a provable upper bound on the rescheduling cost.

Let $\beta_{CA}$ denote the mission time produced by the centralized auction, and $\beta_{OPT_f}$ denote the mission time achieved by an offline optimal solver with perfect foreknowledge of failure times and locations. Since the offline solver utilizes failing vehicles until the exact moment of failure, whereas the online algorithm must reactively reassign work, any online strategy satisfies $\beta_{CA} \ge \beta_{OPT_f}$.

The performance gap is driven by the rescheduling penalty, defined as the unavoidable deadhead and recharge time required to reach the site of a failure. In the Supplementary Materials, we provide a detailed derivation demonstrating that the worst-case mission time is bounded additively by the vehicle's physical constraints rather than the total mission duration:

\begin{equation}
\beta_{CA} \le \beta_{OPT_f} + 2| \mathcal{J}_{fail}|(C + R_T)
\end{equation}

Here, $|\mathcal{J}_{fail}|$ is the number of reassigned trips, $C$ is the battery capacity, and $R_T$ is the recharge time. The factor $2(C+R_T)$ represents the maximum rescheduling cost to traverse the graph to a failure location and return to a depot. We present this as an additive deviation because the cost of reaching a failure is fixed. In scenarios where the optimal mission time $\beta_{OPT_f}$ is short, a standard multiplicative competitive ratio would become arbitrarily large due to a small denominator, rendering it a poor metric for stability. A full proof, along with a visual analysis of the best-case and worst-case relocation scenarios (Figure S1), is provided in the Supplementary Materials (Section 1).

\subsection{Computational Complexity of Reactive Framework}

The computational complexity of the proposed reactive framework is derived by aggregating the operational costs of the centralized auction and the peer auction refinement. The centralized auction achieves a complexity of $O(K (|N_d| + D/\Delta r))$, scaling linearly with the fleet size $K$ to ensure immediate feasibility. The subsequent peer auction utilizes a computational budget $L$ to bound the search space of route exchanges, resulting in a complexity dominated by the routing operations within the iterative improvement loop. Combining these stages, the total complexity is $O(K (|N_d| + D/\Delta r) + R (M^2 W^2 + L \cdot |E_{rem}| \cdot \text{deg}(G)))$. By fixing the window size $W$ and the computational budget $L$, the framework effectively transforms the re-optimization problem into a polynomial-time heuristic operation, guaranteeing predictable scalability for large instances. A detailed derivation of this complexity analysis, including the breakdown of the trip generation and magnetic field routing procedures, is provided in the Supplementary Materials (Section 2).

\section{Conclusion}
\label{sec:conclusion}

This paper introduced a two-stage reactive framework for ensuring mission continuity in the MD-RPP-RRV involving stochastic vehicle failures. To address vehicle failures dynamically, the approach integrates a centralized auction for rapid baseline generation with a peer auction utilizing a new magnetic field router for local schedule repair. Theoretical analysis established a worst-case additive performance bound for the centralized stage, proving that the rescheduling cost is physically constrained by the vehicle's battery capacity and recharge time and scales linearly with the number of failed trips.

Empirically, the framework bridges the gap between fast greedy heuristics and intensive global optimization. By tuning the peer auction window and budget parameters to remain within a tractable regime, the framework prevents combinatorial growth, achieving solution quality within 8\% of the best-known metaheuristic baselines while reducing computational runtime by orders of magnitude. This performance confirms that local schedule repair strategies can effectively surrogate global re-optimization in time-critical contexts, satisfying the strict latency requirements of real-time operations.

Future research will extend this framework to predictive maintenance, utilizing failure precursors to initiate mitigation strategies before breakdowns occur. Furthermore, adapting the auction mechanisms to handle stochastic demands and dynamic graph updates would broaden the system's applicability to more complex, unstructured environments.

\section*{Acknowledgments}
This work was partly supported by a gift from Dr.\ Alex Mehr (Ph.D. ’03) through the Design Decision Support Laboratory Research and Education Fund, the Dean's fellowship program and partly by an Army Cooperative Agreement W911NF2120076. Finally, the authors acknowledge the University of Maryland High-Performance Computing resources (\href{http://hpcc.umd.edu}{{\color{blue}Zaratan}}) made available for conducting the research reported in this article.

\bibliographystyle{apalike-catalan}
\bibliography{cas-refs}

\clearpage

\section*{SUMMARY OF SUPPLEMENTARY MATERIALS}

This section presents the supplementary materials to the paper: ``A Two-Stage Reactive Auction Framework for the Multi-Depot Rural Postman Problem with Dynamic Vehicle Failures''. The detailed experimental data provided here supports the analysis presented in Section 5.2 (Experimentation) of the main manuscript. This document is organized into subsequent tables that categorize the results by instance type. Table \ref{tab:symbol_definitions} provides the nomenclature and symbol definitions used for interpreting the data columns used in Tables \ref{gdb_failure_scenario} through \ref{eglese_failure_scenario_part_b}. Table \ref{gdb_failure_scenario} presents the complete experimental results for the GDB instances.Tables \ref{bccm_failure_scenario_part_a} and \ref{bccm_failure_scenario_part_b} contains the results for the BCCM instances. Finally, Tables \ref{eglese_failure_scenario_part_a} and \ref{eglese_failure_scenario_part_b} details the results for the EGLESE instances.

The experimental tables detail the specific parameters for each failure scenario, including graph topology, fleet configuration, and failure characteristics. These tables provide a comparative analysis between the benchmark Reactive Simulated Annealing (SA) metaheuristic and the proposed Two-Stage Framework. For the SA benchmark, we report the initialization time, average execution time, and statistical distribution of the mission completion times (mean, standard deviation, best, and worst) over multiple trials. For the proposed framework, the results are broken down into the first stage (Centralized Auction, CA) and the complete process (Centralized Auction plus Peer Auction, CA+PA), reporting both the computational execution time and the final mission completion time for each instance.

Tables \ref{sensitivity_analysis_gdb}, \ref{sensitivity_analysis_bccm}, and \ref{sensitivity_analysis_eglese} contain the sensitivity analysis on dataset GDB, BCCM, and EGLESE respectively. These results correspond to Section 5.3 in the main manuscript, which discusses the sensitivity analysis of the algorithmic parameters. Additionally, to ensure reproducibility, the detailed pseudocode for the SEARCH, TRIPINDEX, CALCBID, and INSERTTRIP procedures is provided.

The final sections of this document address the theoretical and computational properties of the proposed approach. Section \ref{sec:supp_theoretical_bound} provides the Derivation of Theoretical Performance Bound of the centralized auction in the proposed two stage reactive framework. Section \ref{sec:comp_complex} describes the Detailed Computational Complexity Analysis of the two stage reactive framework's centralized and peer auction. Finally, Section \ref{sec:prob-for} present a MILP formulation of the studied MD-RPP-RRV with vehicle failures problem.

\begin{table}[h!]
\centering
\caption{Description of column headers and symbols used in the experimental results tables.}
\label{tab:symbol_definitions}
\renewcommand{\arraystretch}{1.3}
\begin{tabular}{lp{11cm}}
\hline
\textbf{Symbol} & \textbf{Description} \\ \hline

\multicolumn{2}{l}{\textit{\textbf{Problem Instance Parameters}}} \\ \hline
$C$ & Vehicle battery capacity (maximum travel distance/time). \\
$|E|$ & Total number of edges in the graph. \\
$|E_u|$ & Number of required edges (tasks) that must be serviced. \\
$|F|$ & Number of simulated vehicle failures in the scenario. \\
Instance Name & The identifier for the specific benchmark graph instance. \\
$K$ & Total size of the vehicle fleet. \\
$|N|$ & Total number of nodes in the graph. \\
$|N_d|$ & Number of depot nodes in the graph. \\
$R_T$ & Time required to fully recharge a vehicle's battery. \\ \hline

\multicolumn{2}{l}{\textit{\textbf{Simulated Annealing (SA) Benchmarks}}} \\ \hline
Best $\beta_{SA}$ & The best (minimum) mission time achieved by the reactive simulated annealing metaheuristic. \\
$\beta_{SA}^{\text{init}}$ & The mission time of the initial schedule (before any failure occurs), generated by the offline simulated annealing solver. \\
$ET_{SA}$ & Average computational execution time (sec) of the reactive simulated annealing metaheuristic. \\
Mean $\beta_{SA}$ & Average mission time achieved by the reactive simulated annealing metaheuristic over multiple runs. \\
STDEV & Standard deviation of the mission times produced by the reactive simulated annealing metaheuristic. \\
Worst $\beta_{SA}$ & The worst (maximum) mission time achieved by the reactive simulated annealing metaheuristic. \\ \hline

\multicolumn{2}{l}{\textit{\textbf{Proposed Framework Performance}}} \\ \hline
$\beta_{CA}$ & Mission time achieved by the centralized auction (Stage 1). \\
$\beta_{CA+PA}$ & Mission time achieved by the proposed two-stage framework (centralized + peer auction). \\
$ET_{CA}$ & Computational execution time (sec) of the centralized auction (Stage 1). \\
$ET_{CA+PA}$ & Total computational execution time (sec) of the proposed two-stage framework (centralized + peer auction). \\ \hline

\multicolumn{2}{l}{\textit{\textbf{Sensitivity Analysis \& Peer Auction Parameters}}} \\ \hline
$L$ & Computational budget: The maximum number of transactions evaluated per round. \\
$R$ & The configured maximum number of rounds (iterations) allowed for the peer auction. \\
$R_{avg}$ & The average number of rounds actually performed before convergence or termination across scenarios. \\
$R_{max}$ & The maximum number of rounds actually performed before convergence or termination across scenarios. \\
$W$ & Window size: The length of contiguous trip segments generated for potential swaps. \\ \hline

\end{tabular}
\end{table}

\begin{table}[t!]
\renewcommand{\arraystretch}{1.15}
\caption{\textrm{GDB Failure Scenario Results}}
\label{gdb_failure_scenario}
\resizebox{\textwidth}{!}{
\begin{tabular}{ccccccccccccccccccc}
\hline
\textbf{\begin{tabular}[c]{@{}c@{}}Instance\\ Name\end{tabular}} & 
\textbf{$|N|$} & 
\textbf{$|E|$} & 
\textbf{$|E_u|$} & 
\textbf{$C$} & 
\textbf{$R_T$} & 
\textbf{$K$} & 
\textbf{$|N_d|$} & 
\textbf{$|F|$} & 
\textbf{$\beta_{SA}^{\text{init}}$} & 
\textbf{\begin{tabular}[c]{@{}c@{}}$ET_{SA}$\\ (sec)\end{tabular}} & 
\textbf{\begin{tabular}[c]{@{}c@{}}Mean\\ $\beta_{SA}$\end{tabular}} & 
\textbf{STDEV} & 
\textbf{\begin{tabular}[c]{@{}c@{}}Best\\ $\beta_{SA}$\end{tabular}} & 
\textbf{\begin{tabular}[c]{@{}c@{}}Worst\\ $\beta_{SA}$\end{tabular}} & 
\textbf{\begin{tabular}[c]{@{}c@{}}$ET_{CA}$\\ (sec)\end{tabular}} & 
\textbf{$\beta_{CA}$} & 
\textbf{\begin{tabular}[c]{@{}c@{}}$ET_{CA+PA}$\\ (sec)\end{tabular}} & 
\textbf{$\beta_{CA+PA}$} \\ \hline
gdb.1                                                            & 11           & 19           & 10              & 40         & 80            & 5          & 5               & 3            & 37                                                            & 181.9       & 256.6                                                                                & 2.2            & 252                                                                                  & 264                                                                                   & 0.006               & 264                                   & 0.021                  & 260                                      \\
gdb.2                                                            & 7            & 21           & 11              & 16         & 32            & 2          & 2               & 1            & 97                                                            & 111.9       & 193                                                                                  & 0              & 193                                                                                  & 193                                                                                   & 0.003               & 193                                   & 0.002                  & 193                                      \\
gdb.3                                                            & 7            & 21           & 11              & 12         & 24            & 2          & 2               & 1            & 12                                                            & 50.7        & 48                                                                                   & 0              & 48                                                                                   & 48                                                                                    & 0.001               & 48                                    & 0.001                  & 48                                       \\
gdb.4                                                            & 7            & 21           & 11              & 12         & 24            & 2          & 2               & 1            & 12                                                            & 53          & 48                                                                                   & 0              & 48                                                                                   & 48                                                                                    & 0.001               & 48                                    & 0                      & 48                                       \\
gdb.5                                                            & 7            & 21           & 11              & 12         & 24            & 2          & 2               & 1            & 12                                                            & 53.9        & 48                                                                                   & 0              & 48                                                                                   & 48                                                                                    & 0.001               & 48                                    & 0                      & 48                                       \\
gdb.6                                                            & 12           & 22           & 11              & 40         & 80            & 4          & 4               & 3            & 136                                                           & 254.9       & 594.9                                                                                & 27.5           & 577                                                                                  & 678                                                                                   & 0.013               & 661                                   & 0.09                   & 678                                      \\
gdb.7                                                            & 12           & 22           & 11              & 40         & 80            & 5          & 5               & 2            & 122                                                           & 75.1        & 149                                                                                  & 0              & 149                                                                                  & 149                                                                                   & 0.005               & 233                                   & 0.012                  & 237                                      \\
gdb.8                                                            & 12           & 22           & 11              & 40         & 80            & 4          & 4               & 1            & 127                                                           & 16.7        & 212.6                                                                                & 19.2           & 155                                                                                  & 271                                                                                   & 0.003               & 155                                   & 0.001                  & 155                                      \\
gdb.9                                                            & 12           & 22           & 11              & 44         & 88            & 4          & 4               & 1            & 43                                                            & 33.1        & 159                                                                                  & 0              & 159                                                                                  & 159                                                                                   & 0.004               & 259                                   & 0.006                  & 259                                      \\
gdb.10                                                           & 12           & 22           & 11              & 40         & 80            & 5          & 5               & 2            & 39                                                            & 61          & 149                                                                                  & 0              & 149                                                                                  & 149                                                                                   & 0.005               & 258                                   & 0.013                  & 258                                      \\
gdb.11                                                           & 12           & 22           & 11              & 44         & 88            & 4          & 4               & 1            & 39                                                            & 44.4        & 145                                                                                  & 0              & 145                                                                                  & 145                                                                                   & 0.003               & 159                                   & 0.004                  & 159                                      \\
gdb.12                                                           & 11           & 22           & 11              & 18         & 36            & 2          & 2               & 1            & 107                                                           & 108.3       & 213                                                                                  & 0              & 213                                                                                  & 213                                                                                   & 0.004               & 250                                   & 0.004                  & 218                                      \\
gdb.13                                                           & 12           & 22           & 11              & 40         & 80            & 4          & 4               & 2            & 136                                                           & 106.6       & 253.4                                                                                & 2              & 251                                                                                  & 260                                                                                   & 0.007               & 350                                   & 0.036                  & 258                                      \\
gdb.14                                                           & 11           & 22           & 11              & 18         & 36            & 2          & 2               & 1            & 108                                                           & 110.3       & 213                                                                                  & 0              & 213                                                                                  & 213                                                                                   & 0.003               & 216                                   & 0.01                   & 214                                      \\
gdb.15                                                           & 13           & 23           & 12              & 60         & 120           & 3          & 3               & 2            & 217                                                           & 224.1       & 749                                                                                  & 0              & 749                                                                                  & 749                                                                                   & 0.013               & 885                                   & 0.025                  & 885                                      \\
gdb.16                                                           & 12           & 25           & 13              & 38         & 76            & 2          & 2               & 1            & 239                                                           & 83.7        & 457                                                                                  & 30             & 367                                                                                  & 547                                                                                   & 0.006               & 449                                   & 0.004                  & 367                                      \\
gdb.17                                                           & 12           & 25           & 13              & 38         & 76            & 2          & 2               & 1            & 239                                                           & 62.6        & 520.1                                                                                & 62.6           & 351                                                                                  & 708                                                                                   & 0.005               & 351                                   & 0.003                  & 351                                      \\
gdb.18                                                           & 12           & 25           & 13              & 38         & 76            & 2          & 2               & 1            & 239                                                           & 124.7       & 549                                                                                  & 0              & 549                                                                                  & 549                                                                                   & 0.008               & 553                                   & 0.008                  & 553                                      \\
gdb.19                                                           & 12           & 26           & 13              & 40         & 80            & 3          & 3               & 1            & 239                                                           & 63.7        & 358.4                                                                                & 33.5           & 258                                                                                  & 459                                                                                   & 0.004               & 262                                   & 0.002                  & 258                                      \\
gdb.20                                                           & 13           & 26           & 13              & 44         & 88            & 5          & 5               & 4            & 134                                                           & 526.6       & 680.3                                                                                & 41.2           & 633                                                                                  & 804                                                                                   & 0.018               & 950                                   & 0.246                  & 662                                      \\
gdb.21                                                           & 8            & 28           & 14              & 16         & 32            & 2          & 2               & 1            & 64                                                            & 128.8       & 156.4                                                                                & 0.5            & 156                                                                                  & 158                                                                                   & 0.003               & 159                                   & 0.003                  & 159                                      \\
gdb.22                                                           & 8            & 28           & 14              & 14         & 28            & 2          & 2               & 1            & 46                                                            & 97.4        & 92                                                                                   & 0              & 92                                                                                   & 92                                                                                    & 0.003               & 120                                   & 0.002                  & 95                                       \\
gdb.23                                                           & 8            & 28           & 14              & 14         & 28            & 2          & 2               & 1            & 46                                                            & 103.3       & 92                                                                                   & 0              & 92                                                                                   & 92                                                                                    & 0.002               & 120                                   & 0.003                  & 120                                      \\
gdb.24                                                           & 8            & 28           & 14              & 16         & 32            & 2          & 2               & 1            & 95                                                            & 129         & 156                                                                                  & 0              & 156                                                                                  & 156                                                                                   & 0.003               & 188                                   & 0.003                  & 188                                      \\
gdb.25                                                           & 10           & 28           & 14              & 198        & 396           & 2          & 2               & 1            & 110                                                           & 74.3        & 609                                                                                  & 0              & 609                                                                                  & 609                                                                                   & 0.013               & 1012                                  & 0.004                  & 638                                      \\
gdb.26                                                           & 10           & 28           & 14              & 198        & 396           & 2          & 2               & 1            & 109                                                           & 89          & 604                                                                                  & 0              & 604                                                                                  & 604                                                                                   & 0.008               & 614                                   & 0.005                  & 610                                      \\
gdb.27                                                           & 11           & 33           & 17              & 18         & 36            & 2          & 2               & 1            & 106                                                           & 145.1       & 213                                                                                  & 0              & 213                                                                                  & 213                                                                                   & 0.004               & 213                                   & 0.004                  & 213                                      \\
gdb.28                                                           & 11           & 33           & 17              & 18         & 36            & 2          & 2               & 1            & 107                                                           & 165.3       & 213                                                                                  & 0              & 213                                                                                  & 213                                                                                   & 0.004               & 249                                   & 0.001                  & 220                                      \\
gdb.29                                                           & 11           & 33           & 17              & 18         & 36            & 2          & 2               & 1            & 107                                                           & 114.8       & 207.6                                                                                & 16.2           & 159                                                                                  & 257                                                                                   & 0.002               & 159                                   & 0.002                  & 159                                      \\
gdb.30                                                           & 9            & 36           & 18              & 16         & 32            & 2          & 2               & 1            & 109                                                           & 224.5       & 244.6                                                                                & 1.3            & 242                                                                                  & 249                                                                                   & 0.004               & 250                                   & 0.006                  & 250                                      \\
gdb.31                                                           & 9            & 36           & 18              & 16         & 32            & 2          & 2               & 1            & 109                                                           & 182.7       & 240.1                                                                                & 12.4           & 203                                                                                  & 278                                                                                   & 0.004               & 236                                   & 0.004                  & 205                                      \\
gdb.32                                                           & 9            & 36           & 18              & 16         & 32            & 2          & 2               & 1            & 108                                                           & 209         & 248.4                                                                                & 0.7            & 248                                                                                  & 251                                                                                   & 0.004               & 248                                   & 0.006                  & 248                                      \\
gdb.33                                                           & 11           & 44           & 22              & 18         & 36            & 2          & 2               & 1            & 117                                                           & 262.2       & 268.5                                                                                & 0.5            & 268                                                                                  & 270                                                                                   & 0.005               & 270                                   & 0.008                  & 270                                      \\
gdb.34                                                           & 11           & 44           & 22              & 18         & 36            & 2          & 2               & 1            & 118                                                           & 226.6       & 266                                                                                  & 14.4           & 223                                                                                  & 310                                                                                   & 0.003               & 223                                   & 0                      & 223                                      \\
gdb.35                                                           & 11           & 44           & 22              & 18         & 36            & 2          & 2               & 1            & 116                                                           & 270.8       & 268.9                                                                                & 1.1            & 267                                                                                  & 273                                                                                   & 0.005               & 268                                   & 0.009                  & 268                                      \\
gdb.36                                                           & 11           & 44           & 22              & 18         & 36            & 2          & 2               & 1            & 117                                                           & 278         & 269.6                                                                                & 0.7            & 269                                                                                  & 272                                                                                   & 0.004               & 270                                   & 0.006                  & 270                                      \\
gdb.37                                                           & 11           & 44           & 22              & 18         & 36            & 2          & 2               & 1            & 117                                                           & 235.7       & 264.5                                                                                & 15.5           & 218                                                                                  & 311                                                                                   & 0.004               & 255                                   & 0.005                  & 227               \\ \hline                      
\end{tabular}}
\end{table}

\clearpage

\begin{table}[t!]
\renewcommand{\arraystretch}{1.15}
\caption{\textrm{BCCM Failure Scenario Results Part-A}}
\label{bccm_failure_scenario_part_a}
\resizebox{\textwidth}{!}{
\begin{tabular}{ccccccccccccccccccc}
\hline
\textbf{\begin{tabular}[c]{@{}c@{}}Instance\\ Name\end{tabular}} & 
\textbf{$|N|$} & 
\textbf{$|E|$} & 
\textbf{$|E_u|$} & 
\textbf{$C$} & 
\textbf{$R_T$} & 
\textbf{$K$} & 
\textbf{$|N_d|$} & 
\textbf{$|F|$} & 
\textbf{$\beta_{SA}^{\text{init}}$} & 
\textbf{\begin{tabular}[c]{@{}c@{}}$ET_{SA}$\\ (sec)\end{tabular}} & 
\textbf{\begin{tabular}[c]{@{}c@{}}Mean\\ $\beta_{SA}$\end{tabular}} & 
\textbf{STDEV} & 
\textbf{\begin{tabular}[c]{@{}c@{}}Best\\ $\beta_{SA}$\end{tabular}} & 
\textbf{\begin{tabular}[c]{@{}c@{}}Worst\\ $\beta_{SA}$\end{tabular}} & 
\textbf{\begin{tabular}[c]{@{}c@{}}$ET_{CA}$\\ (sec)\end{tabular}} & 
\textbf{$\beta_{CA}$} & 
\textbf{\begin{tabular}[c]{@{}c@{}}$ET_{CA+PA}$\\ (sec)\end{tabular}} & 
\textbf{$\beta_{CA+PA}$} \\ \hline
bccm.1                                                           & 24           & 34           & 17              & 24         & 48            & 7          & 7               & 3            & 72                                                            & 261.60      & 135.70                                                                               & 2.10           & 135                                                                                  & 142                                                                                   & 0.01                & 152                                   & 0.07                   & 149                                      \\
bccm.2                                                           & 24           & 34           & 17              & 24         & 48            & 7          & 7               & 2            & 76                                                            & 176         & 80                                                                                   & 0              & 80                                                                                   & 80                                                                                    & 0.01                & 153                                   & 0.05                   & 80                                       \\
bccm.3                                                           & 24           & 34           & 17              & 24         & 48            & 7          & 7               & 4            & 27                                                            & 621         & 172.20                                                                               & 18.70          & 159                                                                                  & 229                                                                                   & 0.01                & 219                                   & 0.18                   & 161                                      \\
bccm.4                                                           & 24           & 35           & 18              & 8          & 16            & 6          & 6               & 2            & 28                                                            & 150.70      & 56.20                                                                                & 9.70           & 48                                                                                   & 86                                                                                    & 0.00                & 72                                    & 0.02                   & 72                                       \\
bccm.5                                                           & 24           & 35           & 18              & 8          & 16            & 6          & 6               & 4            & 28                                                            & 755         & 95.80                                                                                & 1.20           & 94                                                                                   & 100                                                                                   & 0.01                & 153                                   & 2.47                   & 97                                       \\
bccm.6                                                           & 24           & 35           & 18              & 8          & 16            & 6          & 6               & 1            & 28                                                            & 69.70       & 49                                                                                   & 0              & 49                                                                                   & 49                                                                                    & 0.00                & 49                                    & 0.01                   & 49                                       \\
bccm.7                                                           & 24           & 35           & 18              & 8          & 16            & 6          & 6               & 1            & 28                                                            & 97.20       & 29.20                                                                                & 3.60           & 28                                                                                   & 40                                                                                    & 0.00                & 50                                    & 0.06                   & 43                                       \\
bccm.8                                                           & 24           & 35           & 18              & 8          & 16            & 6          & 6               & 1            & 25                                                            & 88.70       & 29                                                                                   & 0              & 29                                                                                   & 29                                                                                    & 0.00                & 31                                    & 0.02                   & 29                                       \\
bccm.9                                                           & 24           & 39           & 20              & 20         & 40            & 4          & 4               & 2            & 124                                                           & 360.70      & 228.70                                                                               & 22             & 182                                                                                  & 295                                                                                   & 0.01                & 228                                   & 0.28                   & 185                                      \\
bccm.10                                                          & 24           & 39           & 20              & 20         & 40            & 4          & 4               & 3            & 120                                                           & 810.20      & 442                                                                                  & 20.70          & 419                                                                                  & 505                                                                                   & 0.02                & 622                                   & 0.66                   & 474                                      \\
bccm.11                                                          & 24           & 39           & 20              & 20         & 40            & 4          & 4               & 2            & 125                                                           & 427         & 234.10                                                                               & 14.60          & 192                                                                                  & 278                                                                                   & 0.01                & 237                                   & 0.75                   & 239                                      \\
bccm.12                                                          & 24           & 39           & 20              & 20         & 40            & 4          & 4               & 1            & 77                                                            & 159.40      & 129.50                                                                               & 0.80           & 128                                                                                  & 132                                                                                   & 0.00                & 183                                   & 0.19                   & 130                                      \\
bccm.13                                                          & 24           & 39           & 20              & 20         & 40            & 4          & 4               & 1            & 115                                                           & 82.70       & 133.30                                                                               & 2.10           & 127                                                                                  & 140                                                                                   & 0.00                & 178                                   & 0.01                   & 128                                      \\
bccm.14                                                          & 24           & 39           & 20              & 20         & 40            & 4          & 4               & 1            & 119                                                           & 172.70      & 127.50                                                                               & 1.60           & 126                                                                                  & 133                                                                                   & 0.00                & 185                                   & 0.45                   & 176                                      \\
bccm.15                                                          & 24           & 39           & 20              & 20         & 40            & 4          & 4               & 3            & 121                                                           & 720.60      & 429                                                                                  & 18.50          & 413                                                                                  & 485                                                                                   & 0.02                & 612                                   & 0.51                   & 419                                      \\
bccm.16                                                          & 31           & 50           & 25              & 20         & 40            & 4          & 4               & 1            & 129                                                           & 152.10      & 174.40                                                                               & 12.60          & 137                                                                                  & 213                                                                                   & 0.00                & 189                                   & 0.09                   & 182                                      \\
bccm.17                                                          & 31           & 50           & 25              & 20         & 40            & 4          & 4               & 3            & 123                                                           & 709.60      & 354.10                                                                               & 3.30           & 353                                                                                  & 364                                                                                   & 0.01                & 453                                   & 0.69                   & 409                                      \\
bccm.18                                                          & 31           & 50           & 25              & 20         & 40            & 4          & 4               & 1            & 125                                                           & 230.90      & 175.90                                                                               & 1.80           & 173                                                                                  & 182                                                                                   & 0.01                & 228                                   & 2.74                   & 182                                      \\
bccm.19                                                          & 31           & 50           & 25              & 20         & 40            & 4          & 4               & 3            & 132                                                           & 858.80      & 487.60                                                                               & 21.70          & 463                                                                                  & 553                                                                                   & 0.02                & 673                                   & 1.16                   & 539                                      \\
bccm.20                                                          & 31           & 50           & 25              & 20         & 40            & 4          & 4               & 3            & 133                                                           & 1,097       & 611.80                                                                               & 22.90          & 584                                                                                  & 681                                                                                   & 0.02                & 691                                   & 0.89                   & 598                                      \\
bccm.21                                                          & 31           & 50           & 25              & 20         & 40            & 4          & 4               & 1            & 135                                                           & 168.90      & 189                                                                                  & 1.40           & 185                                                                                  & 194                                                                                   & 0.00                & 193                                   & 0.09                   & 190                                      \\
bccm.22                                                          & 31           & 50           & 25              & 20         & 40            & 4          & 4               & 3            & 131                                                           & 1,094.10    & 587.60                                                                               & 13.40          & 576                                                                                  & 628                                                                                   & 0.02                & 673                                   & 2.45                   & 539                                      \\
bccm.23                                                          & 31           & 50           & 25              & 20         & 40            & 4          & 4               & 3            & 132                                                           & 760.60      & 515.40                                                                               & 16.80          & 465                                                                                  & 566                                                                                   & 0.01                & 522                                   & 0.20                   & 524                                      \\
bccm.24                                                          & 30           & 63           & 32              & 20         & 40            & 8          & 8               & 2            & 128                                                           & 436.20      & 184.50                                                                               & 3.10           & 177                                                                                  & 194                                                                                   & 0.01                & 225                                   & 2.11                   & 180                                      \\
bccm.25                                                          & 30           & 63           & 32              & 20         & 40            & 8          & 8               & 5            & 128                                                           & 1,187.20    & 326.70                                                                               & 17.30          & 305                                                                                  & 379                                                                                   & 0.02                & 392                                   & 5.24                   & 300                                      \\
bccm.26                                                          & 30           & 63           & 32              & 20         & 40            & 8          & 8               & 6            & 132                                                           & 2,344.10    & 522.20                                                                               & 9.30           & 504                                                                                  & 551                                                                                   & 0.04                & 707                                   & 19.68                  & 464                                      \\
bccm.27                                                          & 30           & 63           & 32              & 20         & 40            & 8          & 8               & 1            & 128                                                           & 232.70      & 160.40                                                                               & 21             & 132                                                                                  & 224                                                                                   & 0.01                & 187                                   & 2.19                   & 134                                      \\
bccm.28                                                          & 30           & 63           & 32              & 20         & 40            & 8          & 8               & 3            & 132                                                           & 603.30      & 209.90                                                                               & 21.10          & 182                                                                                  & 274                                                                                   & 0.01                & 239                                   & 3.32                   & 242                                      \\
bccm.29                                                          & 30           & 63           & 32              & 20         & 40            & 8          & 8               & 4            & 124                                                           & 1,313.40    & 280                                                                                  & 12             & 247                                                                                  & 316                                                                                   & 0.02                & 337                                   & 23.88                  & 339                                      \\
bccm.30                                                          & 30           & 63           & 32              & 20         & 40            & 8          & 8               & 3            & 78                                                            & 834.20      & 225.90                                                                               & 12             & 196                                                                                  & 262                                                                                   & 0.01                & 229                                   & 8.11                   & 191                                      \\
bccm.31                                                          & 30           & 63           & 32              & 20         & 40            & 8          & 8               & 6            & 79                                                            & 1,629.10    & 414.80                                                                               & 37.70          & 357                                                                                  & 528                                                                                   & 0.02                & 366                                   & 6.25                   & 406                                      \\
bccm.32                                                          & 30           & 63           & 32              & 20         & 40            & 8          & 8               & 6            & 117                                                           & 2,291.10    & 496.20                                                                               & 20.60          & 465                                                                                  & 558                                                                                   & 0.03                & 508                                   & 10.71                  & 469                                      \\
bccm.33                                                          & 30           & 63           & 32              & 20         & 40            & 8          & 8               & 5            & 117                                                           & 1,673.80    & 360.90                                                                               & 15.10          & 335                                                                                  & 407                                                                                   & 0.02                & 407                                   & 15.54                  & 356                                      \\
bccm.34                                                          & 30           & 63           & 32              & 20         & 40            & 8          & 8               & 1            & 117                                                           & 236.40      & 176.80                                                                               & 7.10           & 160                                                                                  & 199                                                                                   & 0.01                & 181                                   & 0.43                   & 131                                      \\
bccm.35                                                          & 34           & 65           & 33              & 30         & 60            & 5          & 5               & 3            & 275                                                           & 1,291.20    & 631.90                                                                               & 45.30          & 540                                                                                  & 768                                                                                   & 0.02                & 599                                   & 2.53                   & 527                                      \\
bccm.36                                                          & 34           & 65           & 33              & 30         & 60            & 5          & 5               & 4            & 279                                                           & 1,282.90    & 761.40                                                                               & 66             & 639                                                                                  & 960                                                                                   & 0.03                & 948                                   & 4.26                   & 806                                      \\
bccm.37                                                          & 34           & 65           & 33              & 30         & 60            & 5          & 5               & 4            & 201                                                           & 2,252.30    & 1,167.10                                                                             & 67.20          & 1,047                                                                                & 1,369                                                                                 & 0.03                & 1,263                                 & 5.57                   & 1,137                                    \\
bccm.38                                                          & 34           & 65           & 33              & 30         & 60            & 5          & 5               & 4            & 201                                                           & 1,689.40    & 982                                                                                  & 38.10          & 901                                                                                  & 1,097                                                                                 & 0.03                & 1,337                                 & 1.69                   & 978                                      \\
bccm.39                                                          & 34           & 65           & 33              & 30         & 60            & 5          & 5               & 4            & 201                                                           & 2,044.80    & 1,076.90                                                                             & 82.60          & 969                                                                                  & 1,325                                                                                 & 0.03                & 1,269                                 & 4.28                   & 1,059                                    \\
bccm.40                                                          & 34           & 65           & 33              & 30         & 60            & 5          & 5               & 1            & 201                                                           & 247.60      & 326.30                                                                               & 35.10          & 279                                                                                  & 432                                                                                   & 0.01                & 352                                   & 0.45                   & 278                                      \\
bccm.41                                                          & 34           & 65           & 33              & 30         & 60            & 5          & 5               & 4            & 199                                                           & 2,321.40    & 1,270.70                                                                             & 77             & 1,115                                                                                & 1,502                                                                                 & 0.03                & 1,259                                 & 7.76                   & 1,078                                    \\
bccm.42                                                          & 34           & 65           & 33              & 30         & 60            & 5          & 5               & 2            & 201                                                           & 557.50      & 326.60                                                                               & 35.90          & 287                                                                                  & 435                                                                                   & 0.01                & 364                                   & 5.84                   & 358                                      \\
bccm.43                                                          & 34           & 65           & 33              & 30         & 60            & 5          & 5               & 4            & 274                                                           & 2,379.50    & 1,084                                                                                & 87.80          & 973                                                                                  & 1,348                                                                                 & 0.04                & 1,276                                 & 9.07                   & 1,072                                    \\
bccm.44                                                          & 34           & 65           & 33              & 30         & 60            & 5          & 5               & 2            & 269                                                           & 541.20      & 359                                                                                  & 3.60           & 352                                                                                  & 370                                                                                   & 0.01                & 450                                   & 2.43                   & 367                                      \\
bccm.45                                                          & 34           & 65           & 33              & 30         & 60            & 5          & 5               & 4            & 272                                                           & 2,129.40    & 1,019                                                                                & 95.30          & 795                                                                                  & 1,305                                                                                 & 0.04                & 1,509                                 & 9.33                   & 990                                      \\
bccm.46                                                          & 34           & 65           & 33              & 30         & 60            & 5          & 5               & 2            & 270                                                           & 723.90      & 384.70                                                                               & 27.60          & 356                                                                                  & 468                                                                                   & 0.01                & 434                                   & 6.38                   & 353                                      \\
bccm.47                                                          & 34           & 65           & 33              & 30         & 60            & 5          & 5               & 2            & 272                                                           & 703.80      & 397.40                                                                               & 45.60          & 358                                                                                  & 535                                                                                   & 0.01                & 455                                   & 5.87                   & 371                                      \\
bccm.48                                                          & 40           & 66           & 33              & 20         & 40            & 6          & 6               & 2            & 123                                                           & 333.60      & 175.60                                                                               & 14.20          & 133                                                                                  & 219                                                                                   & 0.01                & 183                                   & 0.23                   & 181                                      \\
bccm.49                                                          & 40           & 66           & 33              & 20         & 40            & 6          & 6               & 5            & 130                                                           & 2,450.30    & 750.20                                                                               & 44.90          & 688                                                                                  & 885                                                                                   & 0.04                & 787                                   & 8.87                   & 657                                      \\
bccm.50                                                          & 40           & 66           & 33              & 20         & 40            & 6          & 6               & 3            & 132                                                           & 1,178.30    & 275.30                                                                               & 20.80          & 244                                                                                  & 338                                                                                   & 0.01                & 295                                   & 8.07                   & 247                                      \\
bccm.51                                                          & 40           & 66           & 33              & 20         & 40            & 6          & 6               & 1            & 168                                                           & 288.80      & 226.50                                                                               & 14.30          & 189                                                                                  & 270                                                                                   & 0.01                & 242                                   & 1.95                   & 223                                      \\
bccm.52                                                          & 40           & 66           & 33              & 20         & 40            & 6          & 6               & 1            & 131                                                           & 350.20      & 182.60                                                                               & 1.10           & 181                                                                                  & 186                                                                                   & 0.01                & 191                                   & 3.57                   & 186                                      \\
bccm.53                                                          & 40           & 66           & 33              & 20         & 40            & 6          & 6               & 4            & 135                                                           & 1,532.90    & 422.40                                                                               & 21.10          & 400                                                                                  & 486                                                                                   & 0.02                & 408                                   & 5.32                   & 407                                      \\
bccm.54                                                          & 40           & 66           & 33              & 20         & 40            & 6          & 6               & 3            & 126                                                           & 825.30      & 219.70                                                                               & 19.10          & 188                                                                                  & 277                                                                                   & 0.01                & 285                                   & 3.54                   & 187                                   \\ \hline                      
\end{tabular}}
\end{table}

\clearpage

\begin{table}[t!]
\renewcommand{\arraystretch}{1.15}
\caption{\textrm{BCCM Failure Scenario Results Part-B}}
\label{bccm_failure_scenario_part_b}
\resizebox{\textwidth}{!}{
\begin{tabular}{ccccccccccccccccccc}
\hline
\textbf{\begin{tabular}[c]{@{}c@{}}Instance\\ Name\end{tabular}} & 
\textbf{$|N|$} & 
\textbf{$|E|$} & 
\textbf{$|E_u|$} & 
\textbf{$C$} & 
\textbf{$R_T$} & 
\textbf{$K$} & 
\textbf{$|N_d|$} & 
\textbf{$|F|$} & 
\textbf{$\beta_{SA}^{\text{init}}$} & 
\textbf{\begin{tabular}[c]{@{}c@{}}$ET_{SA}$\\ (sec)\end{tabular}} & 
\textbf{\begin{tabular}[c]{@{}c@{}}Mean\\ $\beta_{SA}$\end{tabular}} & 
\textbf{STDEV} & 
\textbf{\begin{tabular}[c]{@{}c@{}}Best\\ $\beta_{SA}$\end{tabular}} & 
\textbf{\begin{tabular}[c]{@{}c@{}}Worst\\ $\beta_{SA}$\end{tabular}} & 
\textbf{\begin{tabular}[c]{@{}c@{}}$ET_{CA}$\\ (sec)\end{tabular}} & 
\textbf{$\beta_{CA}$} & 
\textbf{\begin{tabular}[c]{@{}c@{}}$ET_{CA+PA}$\\ (sec)\end{tabular}} & 
\textbf{$\beta_{CA+PA}$} \\ \hline
bccm.55                                                          & 40           & 66           & 33              & 20         & 40            & 6          & 6               & 2            & 126                                                           & 631.50      & 227.80                                                                               & 22.30          & 182                                                                                  & 295                                                                                   & 0.01                & 194                                   & 0.52                   & 181                                      \\
bccm.56                                                          & 40           & 66           & 33              & 20         & 40            & 6          & 6               & 1            & 126                                                           & 157.10      & 167.90                                                                               & 12             & 132                                                                                  & 204                                                                                   & 0.01                & 132                                   & 0.03                   & 132                                      \\
bccm.57                                                          & 40           & 66           & 33              & 20         & 40            & 6          & 6               & 5            & 126                                                           & 2,378.90    & 711                                                                                  & 56.40          & 593                                                                                  & 881                                                                                   & 0.03                & 787                                   & 8.55                   & 593                                      \\
bccm.58                                                          & 40           & 66           & 33              & 20         & 40            & 6          & 6               & 2            & 126                                                           & 580.80      & 181.70                                                                               & 0.80           & 180                                                                                  & 185                                                                                   & 0.01                & 187                                   & 2.97                   & 184                                      \\
bccm.59                                                          & 41           & 69           & 35              & 28         & 56            & 7          & 7               & 4            & 173                                                           & 2,108.90    & 404.30                                                                               & 20.70          & 347                                                                                  & 467                                                                                   & 0.02                & 425                                   & 7.67                   & 342                                      \\
bccm.60                                                          & 41           & 69           & 35              & 28         & 56            & 7          & 7               & 4            & 172                                                           & 1,993.40    & 364                                                                                  & 23.80          & 340                                                                                  & 436                                                                                   & 0.03                & 564                                   & 21.07                  & 343                                      \\
bccm.61                                                          & 41           & 69           & 35              & 28         & 56            & 7          & 7               & 3            & 173                                                           & 1,375.90    & 322.30                                                                               & 19.10          & 266                                                                                  & 380                                                                                   & 0.01                & 338                                   & 4.11                   & 260                                      \\
bccm.62                                                          & 41           & 69           & 35              & 28         & 56            & 7          & 7               & 1            & 173                                                           & 362.50      & 182.10                                                                               & 5.70           & 171                                                                                  & 200                                                                                   & 0.01                & 188                                   & 1.08                   & 179                                      \\
bccm.63                                                          & 41           & 69           & 35              & 28         & 56            & 7          & 7               & 5            & 173                                                           & 2,833.50    & 551.80                                                                               & 33.50          & 499                                                                                  & 653                                                                                   & 0.03                & 662                                   & 18.37                  & 503                                      \\
bccm.64                                                          & 41           & 69           & 35              & 28         & 56            & 7          & 7               & 1            & 169                                                           & 344.70      & 180.50                                                                               & 3.20           & 174                                                                                  & 191                                                                                   & 0.01                & 183                                   & 0.49                   & 169                                      \\
bccm.65                                                          & 41           & 69           & 35              & 28         & 56            & 7          & 7               & 6            & 169                                                           & 3,298.40    & 844.70                                                                               & 27.60          & 825                                                                                  & 928                                                                                   & 0.05                & 1,215                                 & 5.54                   & 845                                      \\
bccm.66                                                          & 41           & 69           & 35              & 28         & 56            & 7          & 7               & 5            & 169                                                           & 2,415.40    & 438.30                                                                               & 29.70          & 406                                                                                  & 528                                                                                   & 0.03                & 637                                   & 4.46                   & 502                                      \\
bccm.67                                                          & 41           & 69           & 35              & 28         & 56            & 7          & 7               & 4            & 110                                                           & 1,789.60    & 327.90                                                                               & 21.20          & 267                                                                                  & 392                                                                                   & 0.02                & 416                                   & 3.19                   & 346                                      \\
bccm.68                                                          & 41           & 69           & 35              & 28         & 56            & 7          & 7               & 5            & 171                                                           & 2,333.60    & 450.70                                                                               & 29.20          & 423                                                                                  & 539                                                                                   & 0.03                & 573                                   & 10.28                  & 514                                      \\
bccm.69                                                          & 41           & 69           & 35              & 28         & 56            & 7          & 7               & 4            & 112                                                           & 1,011.10    & 261.80                                                                               & 3.50           & 257                                                                                  & 273                                                                                   & 0.01                & 330                                   & 1.77                   & 268                                      \\
bccm.70                                                          & 41           & 69           & 35              & 28         & 56            & 7          & 7               & 6            & 176                                                           & 3,565.40    & 881.40                                                                               & 31.40          & 829                                                                                  & 976                                                                                   & 0.05                & 1,201                                 & 15.21                  & 919                                      \\
bccm.71                                                          & 41           & 69           & 35              & 28         & 56            & 7          & 7               & 3            & 110                                                           & 1,168.70    & 264.90                                                                               & 1.30           & 263                                                                                  & 269                                                                                   & 0.01                & 259                                   & 4.29                   & 263                                      \\
bccm.72                                                          & 41           & 69           & 35              & 28         & 56            & 7          & 7               & 1            & 110                                                           & 118         & 177.60                                                                               & 1.90           & 172                                                                                  & 184                                                                                   & 0.00                & 115                                   & 0.01                   & 115                                      \\
bccm.73                                                          & 50           & 92           & 46              & 14         & 28            & 10         & 10              & 5            & 93                                                            & 2,687.20    & 178                                                                                  & 13.60          & 167                                                                                  & 219                                                                                   & 0.04                & 240                                   & 40.06                  & 172                                      \\
bccm.74                                                          & 50           & 92           & 46              & 14         & 28            & 10         & 10              & 6            & 122                                                           & 3,686.90    & 224.80                                                                               & 24.20          & 207                                                                                  & 298                                                                                   & 0.05                & 368                                   & 67.70                  & 241                                      \\
bccm.75                                                          & 50           & 92           & 46              & 14         & 28            & 10         & 10              & 2            & 95                                                            & 640.20      & 131.60                                                                               & 1.40           & 129                                                                                  & 136                                                                                   & 0.01                & 166                                   & 2.39                   & 128                                      \\
bccm.76                                                          & 50           & 92           & 46              & 14         & 28            & 10         & 10              & 3            & 94                                                            & 1,087.70    & 128.80                                                                               & 1.20           & 127                                                                                  & 133                                                                                   & 0.02                & 170                                   & 15.60                  & 131                                      \\
bccm.77                                                          & 50           & 92           & 46              & 14         & 28            & 10         & 10              & 2            & 99                                                            & 586.80      & 95.60                                                                                & 9.50           & 91                                                                                   & 125                                                                                   & 0.01                & 137                                   & 3.37                   & 135                                      \\
bccm.78                                                          & 50           & 92           & 46              & 14         & 28            & 10         & 10              & 1            & 88                                                            & 348.70      & 120.60                                                                               & 2.50           & 114                                                                                  & 129                                                                                   & 0.01                & 180                                   & 4.99                   & 165                                      \\
bccm.79                                                          & 50           & 92           & 46              & 14         & 28            & 10         & 10              & 3            & 83                                                            & 1,339.60    & 133                                                                                  & 10.10          & 125                                                                                  & 164                                                                                   & 0.02                & 124                                   & 6.39                   & 124                                      \\
bccm.80                                                          & 50           & 92           & 46              & 14         & 28            & 10         & 10              & 1            & 86                                                            & 377.30      & 87.60                                                                                & 1.20           & 87                                                                                   & 92                                                                                    & 0.01                & 119                                   & 2.20                   & 90                                       \\
bccm.81                                                          & 50           & 92           & 46              & 14         & 28            & 10         & 10              & 2            & 83                                                            & 714         & 114.50                                                                               & 13.40          & 91                                                                                   & 155                                                                                   & 0.01                & 126                                   & 1.51                   & 124                                      \\
bccm.82                                                          & 50           & 92           & 46              & 14         & 28            & 10         & 10              & 5            & 86                                                            & 1,477.20    & 166.10                                                                               & 14.50          & 134                                                                                  & 210                                                                                   & 0.02                & 137                                   & 4.44                   & 163                                      \\
bccm.83                                                          & 50           & 92           & 46              & 14         & 28            & 10         & 10              & 2            & 79                                                            & 890.40      & 93                                                                                   & 0.40           & 92                                                                                   & 95                                                                                    & 0.01                & 92                                    & 2.14                   & 90                                       \\
bccm.84                                                          & 50           & 92           & 46              & 14         & 28            & 10         & 10              & 1            & 93                                                            & 427.40      & 94.40                                                                                & 1.30           & 91                                                                                   & 99                                                                                    & 0.01                & 125                                   & 1.84                   & 94                                       \\
bccm.85                                                          & 50           & 92           & 46              & 14         & 28            & 10         & 10              & 6            & 90                                                            & 2,332.90    & 189.70                                                                               & 16.60          & 170                                                                                  & 240                                                                                   & 0.03                & 197                                   & 16.99                  & 173                                      \\
bccm.86                                                          & 50           & 92           & 46              & 14         & 28            & 10         & 10              & 5            & 91                                                            & 2,859.60    & 197.90                                                                               & 21.80          & 169                                                                                  & 264                                                                                   & 0.03                & 214                                   & 18.99                  & 158                                      \\
bccm.87                                                          & 50           & 92           & 46              & 14         & 28            & 10         & 10              & 5            & 91                                                            & 2,585.80    & 176                                                                                  & 11.60          & 165                                                                                  & 211                                                                                   & 0.03                & 250                                   & 64.95                  & 210                                      \\
bccm.88                                                          & 50           & 92           & 46              & 14         & 28            & 10         & 10              & 3            & 93                                                            & 1,406       & 142.90                                                                               & 14.50          & 131                                                                                  & 187                                                                                   & 0.02                & 158                                   & 7.03                   & 129                                      \\
bccm.89                                                          & 50           & 97           & 49              & 20         & 40            & 7          & 7               & 3            & 192                                                           & 2,105.50    & 377.60                                                                               & 32.30          & 343                                                                                  & 475                                                                                   & 0.02                & 347                                   & 35.53                  & 298                                      \\
bccm.90                                                          & 50           & 97           & 49              & 20         & 40            & 7          & 7               & 5            & 192                                                           & 3,425.60    & 591.90                                                                               & 66.30          & 413                                                                                  & 791                                                                                   & 0.03                & 554                                   & 18.64                  & 578                                      \\
bccm.91                                                          & 50           & 97           & 49              & 20         & 40            & 7          & 7               & 1            & 192                                                           & 240.90      & 264.10                                                                               & 21.70          & 226                                                                                  & 330                                                                                   & 0.01                & 235                                   & 0.01                   & 235                                      \\
bccm.92                                                          & 50           & 97           & 49              & 20         & 40            & 7          & 7               & 6            & 192                                                           & 7,157.60    & 1,342.50                                                                             & 57.30          & 1,242                                                                                & 1,515                                                                                 & 0.09                & 1,611                                 & 63.58                  & 1,276                                    \\
bccm.93                                                          & 50           & 97           & 49              & 20         & 40            & 7          & 7               & 4            & 190                                                           & 2,025.20    & 359.70                                                                               & 11.80          & 348                                                                                  & 396                                                                                   & 0.02                & 397                                   & 14.35                  & 311                                      \\
bccm.94                                                          & 50           & 97           & 49              & 20         & 40            & 7          & 7               & 6            & 196                                                           & 6,487.10    & 1,117.70                                                                             & 62.60          & 1,045                                                                                & 1,306                                                                                 & 0.07                & 1,344                                 & 70.82                  & 1,124                                    \\
bccm.95                                                          & 50           & 97           & 49              & 20         & 40            & 7          & 7               & 1            & 196                                                           & 312.80      & 229.30                                                                               & 16.70          & 196                                                                                  & 280                                                                                   & 0.01                & 232                                   & 0.41                   & 226                                      \\
bccm.96                                                          & 50           & 97           & 49              & 20         & 40            & 7          & 7               & 3            & 196                                                           & 2,236.20    & 356.80                                                                               & 25             & 314                                                                                  & 432                                                                                   & 0.03                & 460                                   & 43.03                  & 398                                      \\
bccm.97                                                          & 50           & 97           & 49              & 20         & 40            & 7          & 7               & 4            & 188                                                           & 2,662.60    & 436.60                                                                               & 20.60          & 407                                                                                  & 499                                                                                   & 0.03                & 516                                   & 36.34                  & 529                                      \\
bccm.98                                                          & 50           & 97           & 49              & 20         & 40            & 7          & 7               & 4            & 196                                                           & 2,235.10    & 376.90                                                                               & 28.40          & 340                                                                                  & 463                                                                                   & 0.03                & 466                                   & 13.11                  & 354                                      \\
bccm.99                                                          & 50           & 97           & 49              & 20         & 40            & 7          & 7               & 6            & 181                                                           & 5,802.50    & 1,119.10                                                                             & 34.20          & 1,039                                                                                & 1,222                                                                                 & 0.07                & 1,233                                 & 38.60                  & 948                                      \\
bccm.100                                                         & 50           & 97           & 49              & 20         & 40            & 7          & 7               & 5            & 181                                                           & 3,797.50    & 564.70                                                                               & 51             & 476                                                                                  & 718                                                                                   & 0.04                & 672                                   & 46.52                  & 483                                      \\
bccm.101                                                         & 50           & 97           & 49              & 20         & 40            & 7          & 7               & 4            & 181                                                           & 2,257.60    & 364.40                                                                               & 36.30          & 299                                                                                  & 474                                                                                   & 0.02                & 348                                   & 6                      & 253                                      \\
bccm.102                                                         & 50           & 97           & 49              & 20         & 40            & 7          & 7               & 2            & 181                                                           & 1,049.60    & 258.30                                                                               & 20.30          & 243                                                                                  & 320                                                                                   & 0.01                & 239                                   & 5.04                   & 195                                      \\
bccm.103                                                         & 50           & 97           & 49              & 20         & 40            & 7          & 7               & 1            & 181                                                           & 357.80      & 202.70                                                                               & 18.50          & 181                                                                                  & 259                                                                                   & 0.01                & 182                                   & 0.25                   & 182                                      \\
bccm.104                                                         & 50           & 97           & 49              & 20         & 40            & 7          & 7               & 2            & 184                                                           & 1,206.30    & 306.60                                                                               & 16.50          & 292                                                                                  & 357                                                                                   & 0.01                & 288                                   & 23.53                  & 247                                      \\
bccm.105                                                         & 50           & 97           & 49              & 20         & 40            & 7          & 7               & 4            & 184                                                           & 2,908.10    & 449                                                                                  & 29.70          & 405                                                                                  & 539                                                                                   & 0.02                & 406                                   & 21.04                  & 356                                      \\
bccm.106                                                         & 50           & 97           & 49              & 20         & 40            & 7          & 7               & 3            & 184                                                           & 1,816.90    & 345.80                                                                               & 33.90          & 293                                                                                  & 448                                                                                   & 0.02                & 356                                   & 16.43                  & 295                                      \\
bccm.107                                                         & 50           & 97           & 49              & 20         & 40            & 7          & 7               & 5            & 184                                                           & 4,178.50    & 686.90                                                                               & 24             & 631                                                                                  & 759                                                                                   & 0.03                & 586                                   & 21.65                  & 530                                      \\
bccm.108                                                         & 50           & 97           & 49              & 20         & 40            & 7          & 7               & 2            & 184                                                           & 1,092.50    & 296.90                                                                               & 7.20           & 285                                                                                  & 319                                                                                   & 0.01                & 277                                   & 10.47                  & 238               \\ \hline                      
\end{tabular}}
\end{table}

\clearpage

\begin{table}[t!]
\renewcommand{\arraystretch}{1.15}
\caption{\textrm{EGLESE Failure Scenario Results Part-A}}
\label{eglese_failure_scenario_part_a}
\resizebox{\textwidth}{!}{
\begin{tabular}{ccccccccccccccccccc}
\hline
\textbf{\begin{tabular}[c]{@{}c@{}}Instance\\ Name\end{tabular}} & 
\textbf{$|N|$} & 
\textbf{$|E|$} & 
\textbf{$|E_u|$} & 
\textbf{$C$} & 
\textbf{$R_T$} & 
\textbf{$K$} & 
\textbf{$|N_d|$} & 
\textbf{$|F|$} & 
\textbf{$\beta_{SA}^{\text{init}}$} & 
\textbf{\begin{tabular}[c]{@{}c@{}}$ET_{SA}$\\ (sec)\end{tabular}} & 
\textbf{\begin{tabular}[c]{@{}c@{}}Mean\\ $\beta_{SA}$\end{tabular}} & 
\textbf{STDEV} & 
\textbf{\begin{tabular}[c]{@{}c@{}}Best\\ $\beta_{SA}$\end{tabular}} & 
\textbf{\begin{tabular}[c]{@{}c@{}}Worst\\ $\beta_{SA}$\end{tabular}} & 
\textbf{\begin{tabular}[c]{@{}c@{}}$ET_{CA}$\\ (sec)\end{tabular}} & 
\textbf{$\beta_{CA}$} & 
\textbf{\begin{tabular}[c]{@{}c@{}}$ET_{CA+PA}$\\ (sec)\end{tabular}} & 
\textbf{$\beta_{CA+PA}$} \\ \hline
eglese.1                                                         & 77           & 98           & 49              & 184        & 368           & 12         & 12              & 5            & 641                                                           & 2735.3      & 1497.1                                                                               & 144.3          & 1184                                                                                 & 1930                                                                                  & 0.071               & 2021                                  & 9.941                  & 1621                                     \\
eglese.2                                                         & 77           & 98           & 49              & 184        & 368           & 12         & 12              & 5            & 620                                                           & 2636.7      & 1665.4                                                                               & 38.4           & 1619                                                                                 & 1781                                                                                  & 0.074               & 2068                                  & 11.138                 & 1713                                     \\
eglese.3                                                         & 77           & 98           & 49              & 184        & 368           & 12         & 12              & 4            & 641                                                           & 1988        & 1349.2                                                                               & 199            & 1167                                                                                 & 1947                                                                                  & 0.063               & 2030                                  & 8.023                  & 1118                                     \\
eglese.4                                                         & 77           & 98           & 49              & 184        & 368           & 12         & 12              & 3            & 619                                                           & 1600.5      & 1101.3                                                                               & 30.4           & 1041                                                                                 & 1193                                                                                  & 0.046               & 1597                                  & 4.66                   & 1152                                     \\
eglese.5                                                         & 77           & 98           & 49              & 184        & 368           & 12         & 12              & 5            & 655                                                           & 2820.9      & 1467.6                                                                               & 183.9          & 1155                                                                                 & 2020                                                                                  & 0.063               & 1643                                  & 11.78                  & 1594                                     \\
eglese.6                                                         & 77           & 98           & 49              & 184        & 368           & 12         & 12              & 6            & 638                                                           & 4403.2      & 2119.1                                                                               & 190.8          & 1745                                                                                 & 2692                                                                                  & 0.095               & 2615                                  & 159.04                 & 2193                                     \\
eglese.7                                                         & 77           & 98           & 49              & 184        & 368           & 12         & 12              & 1            & 645                                                           & 133.8       & 1111                                                                                 & 0              & 1111                                                                                 & 1111                                                                                  & 0.024               & 1144                                  & 0.046                  & 1144                                     \\
eglese.8                                                         & 77           & 98           & 49              & 184        & 368           & 12         & 12              & 6            & 642                                                           & 3974.3      & 2138                                                                                 & 190.9          & 1730                                                                                 & 2711                                                                                  & 0.094               & 2599                                  & 90.866                 & 1730                                     \\
eglese.9                                                         & 77           & 98           & 49              & 184        & 368           & 12         & 12              & 3            & 655                                                           & 1290.9      & 1172.1                                                                               & 35.1           & 1093                                                                                 & 1278                                                                                  & 0.04                & 1207                                  & 2.416                  & 1591                                     \\
eglese.10                                                        & 77           & 98           & 49              & 184        & 368           & 12         & 12              & 3            & 655                                                           & 1614.1      & 1206.2                                                                               & 18.7           & 1186                                                                                 & 1263                                                                                  & 0.048               & 1686                                  & 9.682                  & 1119                                     \\
eglese.11                                                        & 77           & 98           & 49              & 184        & 368           & 12         & 12              & 5            & 642                                                           & 1815.7      & 1418.5                                                                               & 219.3          & 1158                                                                                 & 2077                                                                                  & 0.056               & 1594                                  & 2.836                  & 1217                                     \\
eglese.12                                                        & 77           & 98           & 49              & 184        & 368           & 12         & 12              & 6            & 656                                                           & 3640.2      & 1836.3                                                                               & 144.8          & 1669                                                                                 & 2271                                                                                  & 0.088               & 2469                                  & 24.145                 & 2172                                     \\
eglese.13                                                        & 77           & 98           & 49              & 184        & 368           & 12         & 12              & 1            & 642                                                           & 480.7       & 1088.8                                                                               & 25.1           & 1052                                                                                 & 1165                                                                                  & 0.025               & 1141                                  & 0.678                  & 1141                                     \\
eglese.14                                                        & 77           & 98           & 49              & 184        & 368           & 12         & 12              & 5            & 646                                                           & 2941.3      & 1613.9                                                                               & 24.9           & 1562                                                                                 & 1689                                                                                  & 0.073               & 2065                                  & 21.538                 & 1648                                     \\
eglese.15                                                        & 77           & 98           & 49              & 184        & 368           & 12         & 12              & 2            & 636                                                           & 1047.3      & 1131.4                                                                               & 15.1           & 1089                                                                                 & 1177                                                                                  & 0.038               & 1529                                  & 2.582                  & 1075                                     \\
eglese.16                                                        & 77           & 98           & 49              & 184        & 368           & 12         & 12              & 2            & 718                                                           & 935.9       & 1165.5                                                                               & 53.5           & 1100                                                                                 & 1326                                                                                  & 0.038               & 1569                                  & 1.289                  & 1120                                     \\
eglese.17                                                        & 77           & 98           & 49              & 184        & 368           & 12         & 12              & 6            & 718                                                           & 4683.9      & 2344.9                                                                               & 168.2          & 2152                                                                                 & 2850                                                                                  & 0.105               & 3256                                  & 70.065                 & 2240                                     \\
eglese.18                                                        & 77           & 98           & 49              & 184        & 368           & 12         & 12              & 5            & 718                                                           & 3175.7      & 1704.6                                                                               & 33.3           & 1664                                                                                 & 1805                                                                                  & 0.074               & 2241                                  & 14.039                 & 1682                                     \\
eglese.19                                                        & 77           & 98           & 49              & 184        & 368           & 12         & 12              & 1            & 718                                                           & 570.7       & 1145.7                                                                               & 17.9           & 1120                                                                                 & 1200                                                                                  & 0.031               & 1558                                  & 1.922                  & 1119                                     \\
eglese.20                                                        & 77           & 98           & 49              & 184        & 368           & 12         & 12              & 5            & 718                                                           & 2932.6      & 1822.4                                                                               & 177.8          & 1665                                                                                 & 2356                                                                                  & 0.073               & 2110                                  & 7.471                  & 1672                                     \\
eglese.21                                                        & 77           & 98           & 49              & 184        & 368           & 12         & 12              & 4            & 728                                                           & 2893.8      & 1697.3                                                                               & 10.2           & 1677                                                                                 & 1728                                                                                  & 0.06                & 1795                                  & 26.64                  & 1704                                     \\
eglese.22                                                        & 77           & 98           & 49              & 184        & 368           & 12         & 12              & 5            & 809                                                           & 3421.9      & 1900.6                                                                               & 190.5          & 1704                                                                                 & 2473                                                                                  & 0.07                & 1876                                  & 15.448                 & 1881                                     \\
eglese.23                                                        & 77           & 98           & 49              & 184        & 368           & 12         & 12              & 4            & 699                                                           & 2460.3      & 1584.4                                                                               & 127.9          & 1211                                                                                 & 1969                                                                                  & 0.057               & 1657                                  & 8.524                  & 1206                                     \\
eglese.24                                                        & 77           & 98           & 49              & 184        & 368           & 12         & 12              & 2            & 758                                                           & 1096.6      & 1411.6                                                                               & 203.9          & 1185                                                                                 & 2024                                                                                  & 0.04                & 1662                                  & 2.758                  & 1226                                     \\
eglese.25                                                        & 77           & 98           & 49              & 184        & 368           & 12         & 12              & 6            & 699                                                           & 3878        & 2118.8                                                                               & 272.9          & 1742                                                                                 & 2938                                                                                  & 0.083               & 2237                                  & 26.054                 & 1742                                     \\
eglese.26                                                        & 77           & 98           & 49              & 184        & 368           & 12         & 12              & 2            & 671                                                           & 1044.5      & 1152.6                                                                               & 22.8           & 1118                                                                                 & 1221                                                                                  & 0.04                & 1589                                  & 4.012                  & 1162                                     \\
eglese.27                                                        & 77           & 98           & 49              & 184        & 368           & 12         & 12              & 3            & 651                                                           & 1744.4      & 1254.5                                                                               & 92.8           & 1188                                                                                 & 1533                                                                                  & 0.04                & 1188                                  & 8.951                  & 1152                                     \\
eglese.28                                                        & 77           & 98           & 49              & 184        & 368           & 12         & 12              & 3            & 664                                                           & 1514.9      & 1146.6                                                                               & 2.4            & 1140                                                                                 & 1154                                                                                  & 0.042               & 1200                                  & 3.265                  & 1149                                     \\
eglese.29                                                        & 77           & 98           & 49              & 184        & 368           & 12         & 12              & 3            & 705                                                           & 1247.8      & 1172.3                                                                               & 15.1           & 1148                                                                                 & 1218                                                                                  & 0.041               & 1261                                  & 5.919                  & 1177                                     \\
eglese.30                                                        & 77           & 98           & 49              & 184        & 368           & 12         & 12              & 3            & 705                                                           & 1279.3      & 1537.9                                                                               & 190.2          & 1244                                                                                 & 2109                                                                                  & 0.049               & 1749                                  & 1.872                  & 1235                                     \\
eglese.31                                                        & 77           & 98           & 49              & 184        & 368           & 12         & 12              & 5            & 697                                                           & 2258.4      & 1872.9                                                                               & 190.3          & 1680                                                                                 & 2444                                                                                  & 0.074               & 2163                                  & 19.692                 & 1706                                     \\
eglese.32                                                        & 77           & 98           & 49              & 184        & 368           & 12         & 12              & 2            & 775                                                           & 705.1       & 1410.6                                                                               & 213.4          & 1136                                                                                 & 2051                                                                                  & 0.042               & 1740                                  & 2.3                    & 1673                                     \\
eglese.33                                                        & 77           & 98           & 49              & 184        & 368           & 12         & 12              & 5            & 625                                                           & 1512.9      & 1302.3                                                                               & 206.6          & 1151                                                                                 & 1923                                                                                  & 0.062               & 1658                                  & 2.199                  & 1180                                     \\
eglese.34                                                        & 77           & 98           & 49              & 184        & 368           & 12         & 12              & 3            & 649                                                           & 1215.6      & 1329.8                                                                               & 198.2          & 1139                                                                                 & 1925                                                                                  & 0.048               & 1646                                  & 7.194                  & 1194                                     \\
eglese.35                                                        & 77           & 98           & 49              & 184        & 368           & 12         & 12              & 3            & 631                                                           & 890.3       & 1143.2                                                                               & 50.2           & 1089                                                                                 & 1294                                                                                  & 0.045               & 1493                                  & 1.378                  & 1125                                     \\
eglese.36                                                        & 77           & 98           & 49              & 184        & 368           & 12         & 12              & 5            & 624                                                           & 2217.9      & 1556.8                                                                               & 132.6          & 1166                                                                                 & 1955                                                                                  & 0.075               & 2164                                  & 6.355                  & 1176                                     \\
eglese.37                                                        & 77           & 98           & 49              & 184        & 368           & 12         & 12              & 6            & 649                                                           & 3470.9      & 1823.5                                                                               & 190.4          & 1658                                                                                 & 2395                                                                                  & 0.082               & 2177                                  & 24.076                 & 1704                                     \\
eglese.38                                                        & 77           & 98           & 49              & 184        & 368           & 12         & 12              & 6            & 1088                                                          & 3959.5      & 2380.3                                                                               & 148.8          & 2229                                                                                 & 2827                                                                                  & 0.108               & 3077                                  & 124.102                & 2682                                     \\
eglese.39                                                        & 77           & 98           & 49              & 184        & 368           & 12         & 12              & 2            & 720                                                           & 853.2       & 1209                                                                                 & 20.4           & 1182                                                                                 & 1271                                                                                  & 0.044               & 1768                                  & 13.188                 & 1185                                     \\
eglese.40                                                        & 77           & 98           & 49              & 184        & 368           & 12         & 12              & 2            & 849                                                           & 1031.7      & 1296.3                                                                               & 61.3           & 1199                                                                                 & 1481                                                                                  & 0.039               & 1542                                  & 10.05                  & 1172                                     \\
eglese.41                                                        & 77           & 98           & 49              & 184        & 368           & 12         & 12              & 6            & 1036                                                          & 3459.7      & 2085.6                                                                               & 139.4          & 1703                                                                                 & 2504                                                                                  & 0.097               & 2775                                  & 61.293                 & 2170                                     \\
eglese.42                                                        & 77           & 98           & 49              & 184        & 368           & 12         & 12              & 6            & 1071                                                          & 3846.4      & 2264.9                                                                               & 78.3           & 2169                                                                                 & 2500                                                                                  & 0.095               & 2754                                  & 31.013                 & 2199                                     \\
eglese.43                                                        & 77           & 98           & 49              & 184        & 368           & 12         & 12              & 5            & 665                                                           & 2569.6      & 1665.4                                                                               & 31.1           & 1626                                                                                 & 1759                                                                                  & 0.063               & 1669                                  & 10.035                 & 1659                                     \\
eglese.44                                                        & 77           & 98           & 49              & 184        & 368           & 12         & 12              & 2            & 654                                                           & 820         & 1031.2                                                                               & 165.6          & 694                                                                                  & 1528                                                                                  & 0.04                & 1588                                  & 2.311                  & 1138                                     \\
eglese.45                                                        & 77           & 98           & 49              & 184        & 368           & 12         & 12              & 3            & 648                                                           & 1234        & 1409                                                                                 & 144.1          & 1203                                                                                 & 1842                                                                                  & 0.046               & 1566                                  & 3.762                  & 1237                                     \\
eglese.46                                                        & 77           & 98           & 49              & 184        & 368           & 12         & 12              & 6            & 665                                                           & 2990.9      & 1691.6                                                                               & 134.7          & 1613                                                                                 & 2096                                                                                  & 0.091               & 2534                                  & 46.942                 & 2236                                     \\
eglese.47                                                        & 77           & 98           & 49              & 184        & 368           & 12         & 12              & 4            & 641                                                           & 1981.9      & 1728                                                                                 & 202.8          & 1546                                                                                 & 2337                                                                                  & 0.065               & 2032                                  & 21.743                 & 1482                                     \\
eglese.48                                                        & 77           & 98           & 49              & 184        & 368           & 12         & 12              & 6            & 692                                                           & 2941.2      & 2177.7                                                                               & 227.1          & 1770                                                                                 & 2859                                                                                  & 0.093               & 2730                                  & 116.92                 & 2222                                     \\
eglese.49                                                        & 77           & 98           & 49              & 184        & 368           & 12         & 12              & 5            & 692                                                           & 2200.8      & 1625.5                                                                               & 94.6           & 1348                                                                                 & 1910                                                                                  & 0.083               & 2738                                  & 25.723                 & 1748                                     \\
eglese.50                                                        & 77           & 98           & 49              & 184        & 368           & 12         & 12              & 5            & 692                                                           & 1731.3      & 1526.1                                                                               & 148.4          & 1233                                                                                 & 1972                                                                                  & 0.059               & 1568                                  & 3.301                  & 1670                                     \\
eglese.51                                                        & 77           & 98           & 49              & 184        & 368           & 12         & 12              & 5            & 692                                                           & 1754        & 1473                                                                                 & 202.9          & 1217                                                                                 & 2082                                                                                  & 0.064               & 1709                                  & 3.995                  & 1664                                     \\
eglese.52                                                        & 77           & 98           & 49              & 184        & 368           & 12         & 12              & 2            & 692                                                           & 489         & 1131.7                                                                               & 2.1            & 1128                                                                                 & 1138                                                                                  & 0.033               & 1178                                  & 1.213                  & 1150                                     \\
eglese.53                                                        & 140          & 190          & 95              & 206        & 412           & 9          & 9               & 4            & 2418                                                          & 5389        & 4857                                                                                 & 456.7          & 4254                                                                                 & 6228                                                                                  & 0.108               & 4261                                  & 257.151                & 4633                                     \\
eglese.54                                                        & 140          & 190          & 95              & 206        & 412           & 9          & 9               & 6            & 2418                                                          & 10412       & 7725.8                                                                               & 652.3          & 7111                                                                                 & 9683                                                                                  & 0.212               & 7777                                  & 664.77                 & 6217                                     \\
eglese.55                                                        & 140          & 190          & 95              & 206        & 412           & 9          & 9               & 5            & 2418                                                          & 7456.9      & 5886.6                                                                               & 318.4          & 5411                                                                                 & 6842                                                                                  & 0.161               & 5935                                  & 315.538                & 5428                                     \\
eglese.56                                                        & 140          & 190          & 95              & 206        & 412           & 9          & 9               & 1            & 2418                                                          & 395.7       & 2446.1                                                                               & 3.6            & 2443                                                                                 & 2457                                                                                  & 0.041               & 2443                                  & 0.205                  & 2443                                     \\ \hline                      
\end{tabular}}
\end{table}

\clearpage

\begin{table}[t!]
\renewcommand{\arraystretch}{1.15}
\caption{\textrm{EGLESE Failure Scenario Results Part-B}}
\label{eglese_failure_scenario_part_b}
\resizebox{\textwidth}{!}{
\begin{tabular}{ccccccccccccccccccc}
\hline
\textbf{\begin{tabular}[c]{@{}c@{}}Instance\\ Name\end{tabular}} & 
\textbf{$|N|$} & 
\textbf{$|E|$} & 
\textbf{$|E_u|$} & 
\textbf{$C$} & 
\textbf{$R_T$} & 
\textbf{$K$} & 
\textbf{$|N_d|$} & 
\textbf{$|F|$} & 
\textbf{$\beta_{SA}^{\text{init}}$} & 
\textbf{\begin{tabular}[c]{@{}c@{}}$ET_{SA}$\\ (sec)\end{tabular}} & 
\textbf{\begin{tabular}[c]{@{}c@{}}Mean\\ $\beta_{SA}$\end{tabular}} & 
\textbf{STDEV} & 
\textbf{\begin{tabular}[c]{@{}c@{}}Best\\ $\beta_{SA}$\end{tabular}} & 
\textbf{\begin{tabular}[c]{@{}c@{}}Worst\\ $\beta_{SA}$\end{tabular}} & 
\textbf{\begin{tabular}[c]{@{}c@{}}$ET_{CA}$\\ (sec)\end{tabular}} & 
\textbf{$\beta_{CA}$} & 
\textbf{\begin{tabular}[c]{@{}c@{}}$ET_{CA+PA}$\\ (sec)\end{tabular}} & 
\textbf{$\beta_{CA+PA}$} \\ \hline
eglese.57                                                        & 140          & 190          & 95              & 206        & 412           & 9          & 9               & 3            & 2418                                                          & 3076.1      & 3806.7                                                                               & 222.9          & 3619                                                                                 & 4476                                                                                  & 0.073               & 3186                                  & 274.31                 & 3694                                     \\
eglese.58                                                        & 140          & 190          & 95              & 206        & 412           & 9          & 9               & 5            & 2015                                                          & 7335.2      & 4908                                                                                 & 311.4          & 4286                                                                                 & 5843                                                                                  & 0.149               & 5563                                  & 531.934                & 5594                                     \\
eglese.59                                                        & 140          & 190          & 95              & 206        & 412           & 9          & 9               & 5            & 2405                                                          & 7472.4      & 4824.2                                                                               & 47.2           & 4721                                                                                 & 4966                                                                                  & 0.177               & 6641                                  & 710.157                & 5383                                     \\
eglese.60                                                        & 140          & 190          & 95              & 206        & 412           & 9          & 9               & 6            & 1997                                                          & 9151.8      & 6263.6                                                                               & 381.4          & 5461                                                                                 & 7408                                                                                  & 0.196               & 7159                                  & 632.441                & 7233                                     \\
eglese.61                                                        & 140          & 190          & 95              & 206        & 412           & 9          & 9               & 3            & 2015                                                          & 3568.5      & 4011                                                                                 & 291.5          & 3597                                                                                 & 4886                                                                                  & 0.071               & 3140                                  & 148.333                & 3180                                     \\
eglese.62                                                        & 140          & 190          & 95              & 206        & 412           & 9          & 9               & 3            & 2029                                                          & 3532.1      & 3798.9                                                                               & 149.1          & 3669                                                                                 & 4247                                                                                  & 0.071               & 3228                                  & 105.029                & 3133                                     \\
eglese.63                                                        & 140          & 190          & 95              & 206        & 412           & 9          & 9               & 2            & 2422                                                          & 1725.3      & 3426.6                                                                               & 208.2          & 3090                                                                                 & 4052                                                                                  & 0.058               & 3067                                  & 8.026                  & 3067                                     \\
eglese.64                                                        & 140          & 190          & 95              & 206        & 412           & 9          & 9               & 3            & 2438                                                          & 3933        & 3434.3                                                                               & 350.9          & 3048                                                                                 & 4487                                                                                  & 0.109               & 4678                                  & 540.269                & 3707                                     \\
eglese.65                                                        & 140          & 190          & 95              & 206        & 412           & 9          & 9               & 6            & 2404                                                          & 9386.5      & 6889.2                                                                               & 582.3          & 6015                                                                                 & 8637                                                                                  & 0.183               & 6575                                  & 1026.921               & 7333                                     \\
eglese.66                                                        & 140          & 190          & 95              & 206        & 412           & 9          & 9               & 4            & 1986                                                          & 4607.2      & 4497.5                                                                               & 298.3          & 3825                                                                                 & 5393                                                                                  & 0.103               & 4349                                  & 115.648                & 4163                                     \\
eglese.67                                                        & 140          & 190          & 95              & 206        & 412           & 9          & 9               & 1            & 2403                                                          & 1247.9      & 2692.3                                                                               & 179.7          & 2495                                                                                 & 3232                                                                                  & 0.055               & 3157                                  & 36.825                 & 3157                                     \\
eglese.68                                                        & 140          & 190          & 95              & 206        & 412           & 9          & 9               & 1            & 2479                                                          & 1068.5      & 3034.7                                                                               & 24.6           & 2988                                                                                 & 3109                                                                                  & 0.057               & 3158                                  & 28.031                 & 2926                                     \\
eglese.69                                                        & 140          & 190          & 95              & 206        & 412           & 9          & 9               & 3            & 2475                                                          & 3136.9      & 3605.9                                                                               & 276.8          & 3115                                                                                 & 4437                                                                                  & 0.085               & 3741                                  & 77.272                 & 3173                                     \\
eglese.70                                                        & 140          & 190          & 95              & 206        & 412           & 9          & 9               & 4            & 2467                                                          & 4977.7      & 4784.7                                                                               & 363.1          & 4296                                                                                 & 5874                                                                                  & 0.113               & 4783                                  & 289.468                & 4153                                     \\
eglese.71                                                        & 140          & 190          & 95              & 206        & 412           & 9          & 9               & 5            & 2477                                                          & 5263        & 4740.7                                                                               & 382.6          & 4180                                                                                 & 5889                                                                                  & 0.144               & 5985                                  & 649.141                & 4303                                     \\
eglese.72                                                        & 140          & 190          & 95              & 206        & 412           & 9          & 9               & 5            & 2474                                                          & 6496.4      & 5329.8                                                                               & 416.8          & 4956                                                                                 & 6581                                                                                  & 0.132               & 5356                                  & 731.996                & 5605                                     \\
eglese.73                                                        & 140          & 190          & 95              & 206        & 412           & 9          & 9               & 5            & 2447                                                          & 7361.3      & 5907.4                                                                               & 208.6          & 5539                                                                                 & 6534                                                                                  & 0.176               & 6448                                  & 374.297                & 5351                                     \\
eglese.74                                                        & 140          & 190          & 95              & 206        & 412           & 9          & 9               & 5            & 2412                                                          & 6993.7      & 5321.1                                                                               & 353.4          & 4864                                                                                 & 6382                                                                                  & 0.163               & 6012                                  & 528.676                & 5511                                     \\
eglese.75                                                        & 140          & 190          & 95              & 206        & 412           & 9          & 9               & 5            & 2440                                                          & 6880.6      & 5490.9                                                                               & 156.4          & 5359                                                                                 & 5961                                                                                  & 0.145               & 5871                                  & 368.664                & 5510                                     \\
eglese.76                                                        & 140          & 190          & 95              & 206        & 412           & 9          & 9               & 2            & 2440                                                          & 1381.6      & 3077.9                                                                               & 70.6           & 2972                                                                                 & 3290                                                                                  & 0.06                & 3171                                  & 5.508                  & 2983                                     \\
eglese.77                                                        & 140          & 190          & 95              & 206        & 412           & 9          & 9               & 4            & 2421                                                          & 5470.1      & 4671.7                                                                               & 206.5          & 4253                                                                                 & 5292                                                                                  & 0.138               & 5302                                  & 450.88                 & 4841                                     \\
eglese.78                                                        & 140          & 190          & 95              & 206        & 412           & 9          & 9               & 4            & 2499                                                          & 5209.2      & 5454.8                                                                               & 421.3          & 4917                                                                                 & 6719                                                                                  & 0.118               & 5248                                  & 170.738                & 4240                                     \\
eglese.79                                                        & 140          & 190          & 95              & 206        & 412           & 9          & 9               & 2            & 2499                                                          & 1834.3      & 3207.2                                                                               & 174.7          & 3001                                                                                 & 3732                                                                                  & 0.073               & 3631                                  & 138.098                & 3631                                     \\
eglese.80                                                        & 140          & 190          & 95              & 206        & 412           & 9          & 9               & 2            & 2499                                                          & 1833.2      & 3851.2                                                                               & 264.4          & 3602                                                                                 & 4645                                                                                  & 0.06                & 3062                                  & 14.407                 & 3062                                     \\
eglese.81                                                        & 140          & 190          & 95              & 206        & 412           & 9          & 9               & 2            & 2499                                                          & 2536.6      & 3705.4                                                                               & 158.9          & 3550                                                                                 & 4183                                                                                  & 0.083               & 4171                                  & 304.087                & 3694                                     \\
eglese.82                                                        & 140          & 190          & 95              & 206        & 412           & 9          & 9               & 6            & 2499                                                          & 10924.3     & 7542.7                                                                               & 450.4          & 6732                                                                                 & 8894                                                                                  & 0.237               & 8345                                  & 531.936                & 7929                                     \\
eglese.83                                                        & 140          & 190          & 95              & 206        & 412           & 9          & 9               & 2            & 2495                                                          & 1748.6      & 3577.6                                                                               & 32.9           & 3525                                                                                 & 3677                                                                                  & 0.068               & 3564                                  & 100.815                & 3036                                     \\
eglese.84                                                        & 140          & 190          & 95              & 206        & 412           & 9          & 9               & 6            & 2495                                                          & 6664.2      & 5851.4                                                                               & 653.5          & 4829                                                                                 & 7812                                                                                  & 0.158               & 5919                                  & 209.779                & 4771                                     \\
eglese.85                                                        & 140          & 190          & 95              & 206        & 412           & 9          & 9               & 2            & 2495                                                          & 2308.8      & 3604.7                                                                               & 46.1           & 3501                                                                                 & 3743                                                                                  & 0.071               & 3599                                  & 299.45                 & 3113                                     \\
eglese.86                                                        & 140          & 190          & 95              & 206        & 412           & 9          & 9               & 3            & 2495                                                          & 2522.1      & 3500.1                                                                               & 294.4          & 3174                                                                                 & 4384                                                                                  & 0.079               & 3650                                  & 80.58                  & 3036                                     \\
eglese.87                                                        & 140          & 190          & 95              & 206        & 412           & 9          & 9               & 5            & 2495                                                          & 7749.9      & 5633.3                                                                               & 266            & 5414                                                                                 & 6432                                                                                  & 0.176               & 6529                                  & 944.693                & 6146                                     \\
eglese.88                                                        & 140          & 190          & 95              & 206        & 412           & 9          & 9               & 2            & 2429                                                          & 2205.8      & 3357.8                                                                               & 265.7          & 2953                                                                                 & 4155                                                                                  & 0.061               & 2985                                  & 55.508                 & 2522                                     \\
eglese.89                                                        & 140          & 190          & 95              & 206        & 412           & 9          & 9               & 6            & 2450                                                          & 9944.8      & 6606.3                                                                               & 217.8          & 6308                                                                                 & 7260                                                                                  & 0.23                & 8544                                  & 955.299                & 7669                                     \\
eglese.90                                                        & 140          & 190          & 95              & 206        & 412           & 9          & 9               & 1            & 2434                                                          & 1283.9      & 2972                                                                                 & 15.8           & 2951                                                                                 & 3020                                                                                  & 0.049               & 2997                                  & 11.46                  & 2950                                     \\
eglese.91                                                        & 140          & 190          & 95              & 206        & 412           & 9          & 9               & 3            & 2470                                                          & 2410.4      & 3750.2                                                                               & 208.7          & 3552                                                                                 & 4377                                                                                  & 0.084               & 4125                                  & 11.189                 & 3588                                     \\
eglese.92                                                        & 140          & 190          & 95              & 206        & 412           & 9          & 9               & 6            & 2456                                                          & 15407.4     & 6831.2                                                                               & 507.2          & 6118                                                                                 & 8353                                                                                  & 0.244               & 8725                                  & 903.579                & 6613                                     \\
eglese.93                                                        & 140          & 190          & 95              & 206        & 412           & 9          & 9               & 1            & 2520                                                          & 907.9       & 3489                                                                                 & 192.7          & 3100                                                                                 & 4068                                                                                  & 0.053               & 3118                                  & 19.653                 & 3044                                     \\
eglese.94                                                        & 140          & 190          & 95              & 206        & 412           & 9          & 9               & 5            & 2520                                                          & 7638.7      & 5736.4                                                                               & 429.5          & 4896                                                                                 & 7025                                                                                  & 0.122               & 4846                                  & 233.023                & 5433                                     \\
eglese.95                                                        & 140          & 190          & 95              & 206        & 412           & 9          & 9               & 3            & 2520                                                          & 3094.3      & 4336.8                                                                               & 142.6          & 4209                                                                                 & 4765                                                                                  & 0.078               & 3681                                  & 24.417                 & 3659                                     \\
eglese.96                                                        & 140          & 190          & 95              & 206        & 412           & 9          & 9               & 4            & 2520                                                          & 6504        & 5242                                                                                 & 198.3          & 4881                                                                                 & 5837                                                                                  & 0.153               & 5755                                  & 979.997                & 3806                                     \\
eglese.97                                                        & 140          & 190          & 95              & 206        & 412           & 9          & 9               & 5            & 2520                                                          & 8322.3      & 5556.6                                                                               & 359            & 4953                                                                                 & 6634                                                                                  & 0.164               & 6012                                  & 486.017                & 5469                                     \\
eglese.98                                                        & 140          & 190          & 95              & 206        & 412           & 9          & 9               & 2            & 2508                                                          & 2467        & 3232.3                                                                               & 91.7           & 3101                                                                                 & 3508                                                                                  & 0.087               & 4226                                  & 88.567                 & 3668                                     \\
eglese.99                                                        & 140          & 190          & 95              & 206        & 412           & 9          & 9               & 5            & 2508                                                          & 10522.5     & 5707.6                                                                               & 379.1          & 5127                                                                                 & 6845                                                                                  & 0.167               & 5943                                  & 295.306                & 4291                                     \\
eglese.100                                                       & 140          & 190          & 95              & 206        & 412           & 9          & 9               & 1            & 2508                                                          & 1344.1      & 2713.8                                                                               & 174.9          & 2546                                                                                 & 3239                                                                                  & 0.055               & 3123                                  & 52.642                 & 3123                                     \\
eglese.101                                                       & 140          & 190          & 95              & 206        & 412           & 9          & 9               & 2            & 2508                                                          & 2892.3      & 3352.4                                                                               & 274.8          & 3119                                                                                 & 4177                                                                                  & 0.07                & 3613                                  & 135.113                & 3090                                     \\
eglese.102                                                       & 140          & 190          & 95              & 206        & 412           & 9          & 9               & 6            & 2508                                                          & 8228.3      & 6439                                                                                 & 446            & 5562                                                                                 & 7777                                                                                  & 0.193               & 6972                                  & 522.504                & 6608                                     \\
eglese.103                                                       & 140          & 190          & 95              & 206        & 412           & 9          & 9               & 2            & 2502                                                          & 2419.2      & 3421.4                                                                               & 297.6          & 3085                                                                                 & 4315                                                                                  & 0.068               & 3562                                  & 72.041                 & 3562                                     \\
eglese.104                                                       & 140          & 190          & 95              & 206        & 412           & 9          & 9               & 2            & 2515                                                          & 1989.6      & 3274.8                                                                               & 219.6          & 3079                                                                                 & 3934                                                                                  & 0.077               & 3668                                  & 215.853                & 3513                                     \\
eglese.105                                                       & 140          & 190          & 95              & 206        & 412           & 9          & 9               & 5            & 2521                                                          & 7237.6      & 5538.9                                                                               & 207.2          & 5359                                                                                 & 6161                                                                                  & 0.154               & 5930                                  & 662.341                & 4919                                     \\
eglese.106                                                       & 140          & 190          & 95              & 206        & 412           & 9          & 9               & 2            & 2521                                                          & 2023.9      & 3371.9                                                                               & 340.7          & 3075                                                                                 & 4394                                                                                  & 0.083               & 4153                                  & 180.287                & 3639                                     \\
eglese.107                                                       & 140          & 190          & 95              & 206        & 412           & 9          & 9               & 5            & 2521                                                          & 7527.6      & 6291.3                                                                               & 250.1          & 5905                                                                                 & 7042                                                                                  & 0.186               & 6986                                  & 981.446                & 5919                                     \\
eglese.108                                                       & 140          & 190          & 95              & 206        & 412           & 9          & 9               & 4            & 1933                                                          & 5487.1      & 4199.5                                                                               & 156.8          & 3750                                                                                 & 4670                                                                                  & 0.097               & 3674                                  & 238.16                 & 4229                                     \\
eglese.109                                                       & 140          & 190          & 95              & 206        & 412           & 9          & 9               & 6            & 1933                                                          & 9573.6      & 6819.2                                                                               & 376            & 6491                                                                                 & 7948                                                                                  & 0.181               & 6574                                  & 399.227                & 7119                                     \\
eglese.110                                                       & 140          & 190          & 95              & 206        & 412           & 9          & 9               & 3            & 1944                                                          & 5068.4      & 3823                                                                                 & 248.9          & 3606                                                                                 & 4570                                                                                  & 0.071               & 3073                                  & 58.637                 & 3576                                     \\
eglese.111                                                       & 140          & 190          & 95              & 206        & 412           & 9          & 9               & 6            & 1948                                                          & 8533.9      & 6030.4                                                                               & 240            & 5669                                                                                 & 6751                                                                                  & 0.191               & 7053                                  & 430.491                & 6023                                     \\
eglese.112                                                       & 140          & 190          & 95              & 206        & 412           & 9          & 9               & 5            & 1948                                                          & 6289.7      & 5069.3                                                                               & 246            & 4757                                                                                 & 5808                                                                                  & 0.13                & 5440                                  & 321.707                & 4120                   \\ \hline                      
\end{tabular}}
\end{table}

\clearpage

\begin{table}[t]
\renewcommand{\arraystretch}{1.15}
\caption{\textrm{Sensitivity Analysis GDB instances}}
\label{sensitivity_analysis_gdb}
\resizebox{\textwidth}{!}{
\begin{tabular}{ccccccccccccccccc}
\hline
\textbf{\begin{tabular}[c]{@{}c@{}}Instance\\ Name\end{tabular}} & 
\textbf{$|N|$} & 
\textbf{$|E|$} & 
\textbf{$|E_u|$} & 
\textbf{$C$} & 
\textbf{$R_T$} & 
\textbf{$K$} & 
\textbf{$|N_d|$} & 
\textbf{$|F|$} & 
\textbf{$\beta_{SA}^{\text{init}}$}  & 
\textbf{$W$} & 
\textbf{$L$} & 
\textbf{$R$} & 
\textbf{$R_{avg}$} &
\textbf{$R_{max}$} & 
\textbf{\begin{tabular}[c]{@{}c@{}}$ET_{CA+PA}$\\ (sec)\end{tabular}} & 
\textbf{$\beta_{CA+PA}$} \\ \hline
gdb.1                                                                                & 11                               & 19                               & 10                                  & 40                             & 80                                & 5                              & 5                                   & 3                                & 37                                                                                & 1          & 5          & 20         & 1.3                 & 2                   & 0.016                                      & 263                                                          \\
gdb.1                                                                                & 11                               & 19                               & 10                                  & 40                             & 80                                & 5                              & 5                                   & 3                                & 37                                                                                & 1          & 10         & 20         & 1.3                 & 2                   & 0.032                                      & 263                                                          \\
gdb.1                                                                                & 11                               & 19                               & 10                                  & 40                             & 80                                & 5                              & 5                                   & 3                                & 37                                                                                & 1          & 20         & 20         & 1.3                 & 2                   & 0.016                                      & 263                                                          \\
gdb.1                                                                                & 11                               & 19                               & 10                                  & 40                             & 80                                & 5                              & 5                                   & 3                                & 37                                                                                & 1          & 40         & 20         & 1.3                 & 2                   & 0.016                                      & 263                                                          \\
gdb.1                                                                                & 11                               & 19                               & 10                                  & 40                             & 80                                & 5                              & 5                                   & 3                                & 37                                                                                & 1          & 60         & 20         & 1.3                 & 2                   & 0.004                                      & 263                                                          \\
gdb.1                                                                                & 11                               & 19                               & 10                                  & 40                             & 80                                & 5                              & 5                                   & 3                                & 37                                                                                & 2          & 5          & 20         & 1.3                 & 2                   & 0.021                                      & 263                                                          \\
gdb.1                                                                                & 11                               & 19                               & 10                                  & 40                             & 80                                & 5                              & 5                                   & 3                                & 37                                                                                & 2          & 10         & 20         & 1.3                 & 2                   & 0.022                                      & 260                                                          \\
gdb.1                                                                                & 11                               & 19                               & 10                                  & 40                             & 80                                & 5                              & 5                                   & 3                                & 37                                                                                & 2          & 20         & 20         & 1.3                 & 2                   & 0.033                                      & 260                                                          \\
gdb.1                                                                                & 11                               & 19                               & 10                                  & 40                             & 80                                & 5                              & 5                                   & 3                                & 37                                                                                & 2          & 40         & 20         & 1.3                 & 2                   & 0.031                                      & 260                                                          \\
gdb.1                                                                                & 11                               & 19                               & 10                                  & 40                             & 80                                & 5                              & 5                                   & 3                                & 37                                                                                & 2          & 60         & 20         & 1.3                 & 2                   & 0.032                                      & 260                                                          \\
gdb.1                                                                                & 11                               & 19                               & 10                                  & 40                             & 80                                & 5                              & 5                                   & 3                                & 37                                                                                & 3          & 5          & 20         & 1.3                 & 2                   & 0.016                                      & 263                                                          \\
gdb.1                                                                                & 11                               & 19                               & 10                                  & 40                             & 80                                & 5                              & 5                                   & 3                                & 37                                                                                & 3          & 10         & 20         & 1.3                 & 2                   & 0.011                                      & 260                                                          \\
gdb.1                                                                                & 11                               & 19                               & 10                                  & 40                             & 80                                & 5                              & 5                                   & 3                                & 37                                                                                & 3          & 20         & 20         & 1.3                 & 2                   & 0.016                                      & 260                                                          \\
gdb.1                                                                                & 11                               & 19                               & 10                                  & 40                             & 80                                & 5                              & 5                                   & 3                                & 37                                                                                & 3          & 40         & 20         & 1.3                 & 2                   & 0.027                                      & 260                                                          \\
gdb.1                                                                                & 11                               & 19                               & 10                                  & 40                             & 80                                & 5                              & 5                                   & 3                                & 37                                                                                & 3          & 60         & 20         & 1.3                 & 2                   & 0.018                                      & 260                                                          \\
gdb.1                                                                                & 11                               & 19                               & 10                                  & 40                             & 80                                & 5                              & 5                                   & 3                                & 37                                                                                & 4          & 5          & 20         & 1.3                 & 2                   & 0.016                                      & 263                                                          \\
gdb.1                                                                                & 11                               & 19                               & 10                                  & 40                             & 80                                & 5                              & 5                                   & 3                                & 37                                                                                & 4          & 10         & 20         & 1.3                 & 2                   & 0.032                                      & 260                                                          \\
gdb.1                                                                                & 11                               & 19                               & 10                                  & 40                             & 80                                & 5                              & 5                                   & 3                                & 37                                                                                & 4          & 20         & 20         & 1.3                 & 2                   & 0.025                                      & 260                                                          \\
gdb.1                                                                                & 11                               & 19                               & 10                                  & 40                             & 80                                & 5                              & 5                                   & 3                                & 37                                                                                & 4          & 40         & 20         & 1.3                 & 2                   & 0.025                                      & 260                                                          \\
gdb.1                                                                                & 11                               & 19                               & 10                                  & 40                             & 80                                & 5                              & 5                                   & 3                                & 37                                                                                & 4          & 60         & 20         & 1.3                 & 2                   & 0.032                                      & 260                                                          \\
gdb.1                                                                                & 11                               & 19                               & 10                                  & 40                             & 80                                & 5                              & 5                                   & 3                                & 37                                                                                & 5          & 5          & 20         & 1.3                 & 2                   & 0.016                                      & 263                                                          \\
gdb.1                                                                                & 11                               & 19                               & 10                                  & 40                             & 80                                & 5                              & 5                                   & 3                                & 37                                                                                & 5          & 10         & 20         & 1.3                 & 2                   & 0.026                                      & 260                                                          \\
gdb.1                                                                                & 11                               & 19                               & 10                                  & 40                             & 80                                & 5                              & 5                                   & 3                                & 37                                                                                & 5          & 20         & 20         & 1.3                 & 2                   & 0.016                                      & 260                                                          \\
gdb.1                                                                                & 11                               & 19                               & 10                                  & 40                             & 80                                & 5                              & 5                                   & 3                                & 37                                                                                & 5          & 40         & 20         & 1.3                 & 2                   & 0.031                                      & 260                                                          \\
gdb.1                                                                                & 11                               & 19                               & 10                                  & 40                             & 80                                & 5                              & 5                                   & 3                                & 37                                                                                & 5          & 60         & 20         & 1.3                 & 2                   & 0.032                                      & 260                                                          \\
gdb.13                                                                               & 12                               & 22                               & 11                                  & 40                             & 80                                & 4                              & 4                                   & 2                                & 136                                                                               & 1          & 5          & 20         & 2                   & 2                   & 0.032                                      & 361                                                          \\
gdb.13                                                                               & 12                               & 22                               & 11                                  & 40                             & 80                                & 4                              & 4                                   & 2                                & 136                                                                               & 1          & 10         & 20         & 2.5                 & 3                   & 0.036                                      & 358                                                          \\
gdb.13                                                                               & 12                               & 22                               & 11                                  & 40                             & 80                                & 4                              & 4                                   & 2                                & 136                                                                               & 1          & 20         & 20         & 2.5                 & 3                   & 0.032                                      & 358                                                          \\
gdb.13                                                                               & 12                               & 22                               & 11                                  & 40                             & 80                                & 4                              & 4                                   & 2                                & 136                                                                               & 1          & 40         & 20         & 2.5                 & 3                   & 0.048                                      & 358                                                          \\
gdb.13                                                                               & 12                               & 22                               & 11                                  & 40                             & 80                                & 4                              & 4                                   & 2                                & 136                                                                               & 1          & 60         & 20         & 2.5                 & 3                   & 0.04                                       & 358                                                          \\
gdb.13                                                                               & 12                               & 22                               & 11                                  & 40                             & 80                                & 4                              & 4                                   & 2                                & 136                                                                               & 2          & 5          & 20         & 2                   & 2                   & 0.032                                      & 361                                                          \\
gdb.13                                                                               & 12                               & 22                               & 11                                  & 40                             & 80                                & 4                              & 4                                   & 2                                & 136                                                                               & 2          & 10         & 20         & 2.5                 & 3                   & 0.055                                      & 358                                                          \\
gdb.13                                                                               & 12                               & 22                               & 11                                  & 40                             & 80                                & 4                              & 4                                   & 2                                & 136                                                                               & 2          & 20         & 20         & 1.5                 & 2                   & 0.032                                      & 258                                                          \\
gdb.13                                                                               & 12                               & 22                               & 11                                  & 40                             & 80                                & 4                              & 4                                   & 2                                & 136                                                                               & 2          & 40         & 20         & 1.5                 & 2                   & 0.032                                      & 258                                                          \\
gdb.13                                                                               & 12                               & 22                               & 11                                  & 40                             & 80                                & 4                              & 4                                   & 2                                & 136                                                                               & 2          & 60         & 20         & 1.5                 & 2                   & 0.032                                      & 258                                                          \\
gdb.13                                                                               & 12                               & 22                               & 11                                  & 40                             & 80                                & 4                              & 4                                   & 2                                & 136                                                                               & 3          & 5          & 20         & 2                   & 2                   & 0.032                                      & 361                                                          \\
gdb.13                                                                               & 12                               & 22                               & 11                                  & 40                             & 80                                & 4                              & 4                                   & 2                                & 136                                                                               & 3          & 10         & 20         & 2.5                 & 3                   & 0.048                                      & 358                                                          \\
gdb.13                                                                               & 12                               & 22                               & 11                                  & 40                             & 80                                & 4                              & 4                                   & 2                                & 136                                                                               & 3          & 20         & 20         & 1.5                 & 2                   & 0.036                                      & 258                                                          \\
gdb.13                                                                               & 12                               & 22                               & 11                                  & 40                             & 80                                & 4                              & 4                                   & 2                                & 136                                                                               & 3          & 40         & 20         & 1.5                 & 2                   & 0.042                                      & 258                                                          \\
gdb.13                                                                               & 12                               & 22                               & 11                                  & 40                             & 80                                & 4                              & 4                                   & 2                                & 136                                                                               & 3          & 60         & 20         & 1.5                 & 2                   & 0.031                                      & 258                                                          \\
gdb.13                                                                               & 12                               & 22                               & 11                                  & 40                             & 80                                & 4                              & 4                                   & 2                                & 136                                                                               & 4          & 5          & 20         & 2                   & 2                   & 0.033                                      & 361                                                          \\
gdb.13                                                                               & 12                               & 22                               & 11                                  & 40                             & 80                                & 4                              & 4                                   & 2                                & 136                                                                               & 4          & 10         & 20         & 2.5                 & 3                   & 0.051                                      & 358                                                          \\
gdb.13                                                                               & 12                               & 22                               & 11                                  & 40                             & 80                                & 4                              & 4                                   & 2                                & 136                                                                               & 4          & 20         & 20         & 1.5                 & 2                   & 0.034                                      & 258                                                          \\
gdb.13                                                                               & 12                               & 22                               & 11                                  & 40                             & 80                                & 4                              & 4                                   & 2                                & 136                                                                               & 4          & 40         & 20         & 1.5                 & 2                   & 0.033                                      & 258                                                          \\
gdb.13                                                                               & 12                               & 22                               & 11                                  & 40                             & 80                                & 4                              & 4                                   & 2                                & 136                                                                               & 4          & 60         & 20         & 1.5                 & 2                   & 0.033                                      & 258                                                          \\
gdb.13                                                                               & 12                               & 22                               & 11                                  & 40                             & 80                                & 4                              & 4                                   & 2                                & 136                                                                               & 5          & 5          & 20         & 2                   & 2                   & 0.032                                      & 361                                                          \\
gdb.13                                                                               & 12                               & 22                               & 11                                  & 40                             & 80                                & 4                              & 4                                   & 2                                & 136                                                                               & 5          & 10         & 20         & 2.5                 & 3                   & 0.065                                      & 358                                                          \\
gdb.13                                                                               & 12                               & 22                               & 11                                  & 40                             & 80                                & 4                              & 4                                   & 2                                & 136                                                                               & 5          & 20         & 20         & 1.5                 & 2                   & 0.032                                      & 258                                                          \\
gdb.13                                                                               & 12                               & 22                               & 11                                  & 40                             & 80                                & 4                              & 4                                   & 2                                & 136                                                                               & 5          & 40         & 20         & 1.5                 & 2                   & 0.034                                      & 258                                                          \\
gdb.13                                                                               & 12                               & 22                               & 11                                  & 40                             & 80                                & 4                              & 4                                   & 2                                & 136                                                                               & 5          & 60         & 20         & 1.5                 & 2                   & 0.031                                      & 258                                                                    \\ \hline
\end{tabular}}
\end{table}

\begin{table}[t]
\renewcommand{\arraystretch}{1.15}
\caption{\textrm{Sensitivity Analysis BCCM instances}}
\label{sensitivity_analysis_bccm}
\resizebox{\textwidth}{!}{
\begin{tabular}{ccccccccccccccccc}
\hline
\textbf{\begin{tabular}[c]{@{}c@{}}Instance\\ Name\end{tabular}} & 
\textbf{$|N|$} & 
\textbf{$|E|$} & 
\textbf{$|E_u|$} & 
\textbf{$C$} & 
\textbf{$R_T$} & 
\textbf{$K$} & 
\textbf{$|N_d|$} & 
\textbf{$|F|$} & 
\textbf{$\beta_{SA}^{\text{init}}$}  & 
\textbf{$W$} & 
\textbf{$L$} & 
\textbf{$R$} & 
\textbf{$R_{avg}$} &
\textbf{$R_{max}$} & 
\textbf{\begin{tabular}[c]{@{}c@{}}$ET_{CA+PA}$\\ (sec)\end{tabular}} & 
\textbf{$\beta_{CA+PA}$} \\ \hline
bccm.10                                                                              & 24                               & 39                               & 20                                  & 20                             & 40                                & 4                              & 4                                   & 3                                & 120                                                                               & 1          & 5          & 20         & 1.3                 & 2                   & 0.169                                      & 533                                                          \\
bccm.10                                                                              & 24                               & 39                               & 20                                  & 20                             & 40                                & 4                              & 4                                   & 3                                & 120                                                                               & 1          & 10         & 20         & 1.3                 & 2                   & 0.21                                       & 520                                                          \\
bccm.10                                                                              & 24                               & 39                               & 20                                  & 20                             & 40                                & 4                              & 4                                   & 3                                & 120                                                                               & 1          & 20         & 20         & 1.3                 & 2                   & 0.276                                      & 522                                                          \\
bccm.10                                                                              & 24                               & 39                               & 20                                  & 20                             & 40                                & 4                              & 4                                   & 3                                & 120                                                                               & 1          & 40         & 20         & 1.3                 & 2                   & 0.27                                       & 522                                                          \\
bccm.10                                                                              & 24                               & 39                               & 20                                  & 20                             & 40                                & 4                              & 4                                   & 3                                & 120                                                                               & 1          & 60         & 20         & 1.3                 & 2                   & 0.26                                       & 522                                                          \\
bccm.10                                                                              & 24                               & 39                               & 20                                  & 20                             & 40                                & 4                              & 4                                   & 3                                & 120                                                                               & 2          & 5          & 20         & 1.3                 & 2                   & 0.214                                      & 533                                                          \\
bccm.10                                                                              & 24                               & 39                               & 20                                  & 20                             & 40                                & 4                              & 4                                   & 3                                & 120                                                                               & 2          & 10         & 20         & 1.3                 & 2                   & 0.29                                       & 533                                                          \\
bccm.10                                                                              & 24                               & 39                               & 20                                  & 20                             & 40                                & 4                              & 4                                   & 3                                & 120                                                                               & 2          & 20         & 20         & 2                   & 4                   & 0.565                                      & 474                                                          \\
bccm.10                                                                              & 24                               & 39                               & 20                                  & 20                             & 40                                & 4                              & 4                                   & 3                                & 120                                                                               & 2          & 40         & 20         & 2                   & 4                   & 0.662                                      & 474                                                          \\
bccm.10                                                                              & 24                               & 39                               & 20                                  & 20                             & 40                                & 4                              & 4                                   & 3                                & 120                                                                               & 2          & 60         & 20         & 2                   & 4                   & 0.696                                      & 418                                                          \\
bccm.10                                                                              & 24                               & 39                               & 20                                  & 20                             & 40                                & 4                              & 4                                   & 3                                & 120                                                                               & 3          & 5          & 20         & 1.3                 & 2                   & 0.235                                      & 533                                                          \\
bccm.10                                                                              & 24                               & 39                               & 20                                  & 20                             & 40                                & 4                              & 4                                   & 3                                & 120                                                                               & 3          & 10         & 20         & 1.3                 & 2                   & 0.323                                      & 533                                                          \\
bccm.10                                                                              & 24                               & 39                               & 20                                  & 20                             & 40                                & 4                              & 4                                   & 3                                & 120                                                                               & 3          & 20         & 20         & 2                   & 4                   & 0.603                                      & 474                                                          \\
bccm.10                                                                              & 24                               & 39                               & 20                                  & 20                             & 40                                & 4                              & 4                                   & 3                                & 120                                                                               & 3          & 40         & 20         & 2                   & 4                   & 0.736                                      & 474                                                          \\
bccm.10                                                                              & 24                               & 39                               & 20                                  & 20                             & 40                                & 4                              & 4                                   & 3                                & 120                                                                               & 3          & 60         & 20         & 2                   & 4                   & 0.844                                      & 474                                                          \\
bccm.10                                                                              & 24                               & 39                               & 20                                  & 20                             & 40                                & 4                              & 4                                   & 3                                & 120                                                                               & 4          & 5          & 20         & 1.3                 & 2                   & 0.25                                       & 533                                                          \\
bccm.10                                                                              & 24                               & 39                               & 20                                  & 20                             & 40                                & 4                              & 4                                   & 3                                & 120                                                                               & 4          & 10         & 20         & 1.3                 & 2                   & 0.339                                      & 533                                                          \\
bccm.10                                                                              & 24                               & 39                               & 20                                  & 20                             & 40                                & 4                              & 4                                   & 3                                & 120                                                                               & 4          & 20         & 20         & 2                   & 4                   & 0.623                                      & 474                                                          \\
bccm.10                                                                              & 24                               & 39                               & 20                                  & 20                             & 40                                & 4                              & 4                                   & 3                                & 120                                                                               & 4          & 40         & 20         & 2                   & 4                   & 0.789                                      & 474                                                          \\
bccm.10                                                                              & 24                               & 39                               & 20                                  & 20                             & 40                                & 4                              & 4                                   & 3                                & 120                                                                               & 4          & 60         & 20         & 2                   & 4                   & 0.861                                      & 474                                                          \\
bccm.10                                                                              & 24                               & 39                               & 20                                  & 20                             & 40                                & 4                              & 4                                   & 3                                & 120                                                                               & 5          & 5          & 20         & 1.3                 & 2                   & 0.239                                      & 533                                                          \\
bccm.10                                                                              & 24                               & 39                               & 20                                  & 20                             & 40                                & 4                              & 4                                   & 3                                & 120                                                                               & 5          & 10         & 20         & 1.3                 & 2                   & 0.335                                      & 533                                                          \\
bccm.10                                                                              & 24                               & 39                               & 20                                  & 20                             & 40                                & 4                              & 4                                   & 3                                & 120                                                                               & 5          & 20         & 20         & 2                   & 4                   & 0.608                                      & 474                                                          \\
bccm.10                                                                              & 24                               & 39                               & 20                                  & 20                             & 40                                & 4                              & 4                                   & 3                                & 120                                                                               & 5          & 40         & 20         & 2                   & 4                   & 0.749                                      & 474                                                          \\
bccm.10                                                                              & 24                               & 39                               & 20                                  & 20                             & 40                                & 4                              & 4                                   & 3                                & 120                                                                               & 5          & 60         & 20         & 2                   & 4                   & 0.847                                      & 474                                                          \\
bccm.105                                                                             & 50                               & 97                               & 49                                  & 20                             & 40                                & 7                              & 7                                   & 4                                & 184                                                                               & 1          & 5          & 20         & 2                   & 3                   & 3.827                                      & 411                                                          \\
bccm.105                                                                             & 50                               & 97                               & 49                                  & 20                             & 40                                & 7                              & 7                                   & 4                                & 184                                                                               & 1          & 10         & 20         & 2                   & 3                   & 6.011                                      & 411                                                          \\
bccm.105                                                                             & 50                               & 97                               & 49                                  & 20                             & 40                                & 7                              & 7                                   & 4                                & 184                                                                               & 1          & 20         & 20         & 2.2                 & 4                   & 9.356                                      & 359                                                          \\
bccm.105                                                                             & 50                               & 97                               & 49                                  & 20                             & 40                                & 7                              & 7                                   & 4                                & 184                                                                               & 1          & 40         & 20         & 2.2                 & 4                   & 9.33                                       & 359                                                          \\
bccm.105                                                                             & 50                               & 97                               & 49                                  & 20                             & 40                                & 7                              & 7                                   & 4                                & 184                                                                               & 1          & 60         & 20         & 2.2                 & 4                   & 9.545                                      & 359                                                          \\
bccm.105                                                                             & 50                               & 97                               & 49                                  & 20                             & 40                                & 7                              & 7                                   & 4                                & 184                                                                               & 2          & 5          & 20         & 2                   & 3                   & 5.339                                      & 364                                                          \\
bccm.105                                                                             & 50                               & 97                               & 49                                  & 20                             & 40                                & 7                              & 7                                   & 4                                & 184                                                                               & 2          & 10         & 20         & 2.2                 & 3                   & 7.727                                      & 356                                                          \\
bccm.105                                                                             & 50                               & 97                               & 49                                  & 20                             & 40                                & 7                              & 7                                   & 4                                & 184                                                                               & 2          & 20         & 20         & 2.2                 & 3                   & 11.486                                     & 356                                                          \\
bccm.105                                                                             & 50                               & 97                               & 49                                  & 20                             & 40                                & 7                              & 7                                   & 4                                & 184                                                                               & 2          & 40         & 20         & 2                   & 3                   & 17.536                                     & 356                                                          \\
bccm.105                                                                             & 50                               & 97                               & 49                                  & 20                             & 40                                & 7                              & 7                                   & 4                                & 184                                                                               & 2          & 60         & 20         & 2.2                 & 3                   & 19.665                                     & 355                                                          \\
bccm.105                                                                             & 50                               & 97                               & 49                                  & 20                             & 40                                & 7                              & 7                                   & 4                                & 184                                                                               & 3          & 5          & 20         & 2                   & 3                   & 6.186                                      & 360                                                          \\
bccm.105                                                                             & 50                               & 97                               & 49                                  & 20                             & 40                                & 7                              & 7                                   & 4                                & 184                                                                               & 3          & 10         & 20         & 2                   & 3                   & 8.233                                      & 360                                                          \\
bccm.105                                                                             & 50                               & 97                               & 49                                  & 20                             & 40                                & 7                              & 7                                   & 4                                & 184                                                                               & 3          & 20         & 20         & 2.2                 & 3                   & 13.343                                     & 356                                                          \\
bccm.105                                                                             & 50                               & 97                               & 49                                  & 20                             & 40                                & 7                              & 7                                   & 4                                & 184                                                                               & 3          & 40         & 20         & 2                   & 3                   & 21.396                                     & 356                                                          \\
bccm.105                                                                             & 50                               & 97                               & 49                                  & 20                             & 40                                & 7                              & 7                                   & 4                                & 184                                                                               & 3          & 60         & 20         & 2.2                 & 3                   & 27.257                                     & 355                                                          \\
bccm.105                                                                             & 50                               & 97                               & 49                                  & 20                             & 40                                & 7                              & 7                                   & 4                                & 184                                                                               & 4          & 5          & 20         & 2                   & 3                   & 6.477                                      & 360                                                          \\
bccm.105                                                                             & 50                               & 97                               & 49                                  & 20                             & 40                                & 7                              & 7                                   & 4                                & 184                                                                               & 4          & 10         & 20         & 2                   & 3                   & 8.503                                      & 360                                                          \\
bccm.105                                                                             & 50                               & 97                               & 49                                  & 20                             & 40                                & 7                              & 7                                   & 4                                & 184                                                                               & 4          & 20         & 20         & 2.2                 & 3                   & 13.568                                     & 356                                                          \\
bccm.105                                                                             & 50                               & 97                               & 49                                  & 20                             & 40                                & 7                              & 7                                   & 4                                & 184                                                                               & 4          & 40         & 20         & 2                   & 3                   & 21.652                                     & 356                                                          \\
bccm.105                                                                             & 50                               & 97                               & 49                                  & 20                             & 40                                & 7                              & 7                                   & 4                                & 184                                                                               & 4          & 60         & 20         & 2.2                 & 3                   & 28.128                                     & 355                                                          \\
bccm.105                                                                             & 50                               & 97                               & 49                                  & 20                             & 40                                & 7                              & 7                                   & 4                                & 184                                                                               & 5          & 5          & 20         & 2                   & 3                   & 6.6                                        & 360                                                          \\
bccm.105                                                                             & 50                               & 97                               & 49                                  & 20                             & 40                                & 7                              & 7                                   & 4                                & 184                                                                               & 5          & 10         & 20         & 2                   & 3                   & 8.63                                       & 360                                                          \\
bccm.105                                                                             & 50                               & 97                               & 49                                  & 20                             & 40                                & 7                              & 7                                   & 4                                & 184                                                                               & 5          & 20         & 20         & 2.2                 & 3                   & 13.709                                     & 356                                                          \\
bccm.105                                                                             & 50                               & 97                               & 49                                  & 20                             & 40                                & 7                              & 7                                   & 4                                & 184                                                                               & 5          & 40         & 20         & 2                   & 3                   & 21.77                                      & 356                                                          \\
bccm.105                                                                             & 50                               & 97                               & 49                                  & 20                             & 40                                & 7                              & 7                                   & 4                                & 184                                                                               & 5          & 60         & 20         & 2.2                 & 3                   & 27.956                                     & 355                                                          \\ \hline
\end{tabular}}
\end{table}

\begin{table}[t]
\renewcommand{\arraystretch}{1.15}
\caption{\textrm{Sensitivity Analysis EGLESE instances}}
\label{sensitivity_analysis_eglese}
\resizebox{\textwidth}{!}{
\begin{tabular}{ccccccccccccccccc}
\hline
\textbf{\begin{tabular}[c]{@{}c@{}}Instance\\ Name\end{tabular}} & 
\textbf{$|N|$} & 
\textbf{$|E|$} & 
\textbf{$|E_u|$} & 
\textbf{$C$} & 
\textbf{$R_T$} & 
\textbf{$K$} & 
\textbf{$|N_d|$} & 
\textbf{$|F|$} & 
\textbf{$\beta_{SA}^{\text{init}}$}  & 
\textbf{$W$} & 
\textbf{$L$} & 
\textbf{$R$} & 
\textbf{$R_{avg}$} &
\textbf{$R_{max}$} & 
\textbf{\begin{tabular}[c]{@{}c@{}}$ET_{CA+PA}$\\ (sec)\end{tabular}} & 
\textbf{$\beta_{CA+PA}$} \\ \hline
eglese.10                                                                            & 77                               & 98                               & 49                                  & 184                            & 368                               & 12                             & 12                                  & 3                                & 655                                                                               & 1          & 5          & 20         & 2.7                 & 3                   & 2.43                                       & 1502                                                         \\
eglese.10                                                                            & 77                               & 98                               & 49                                  & 184                            & 368                               & 12                             & 12                                  & 3                                & 655                                                                               & 1          & 10         & 20         & 3                   & 4                   & 2.753                                      & 1122                                                         \\
eglese.10                                                                            & 77                               & 98                               & 49                                  & 184                            & 368                               & 12                             & 12                                  & 3                                & 655                                                                               & 1          & 20         & 20         & 3                   & 4                   & 2.767                                      & 1122                                                         \\
eglese.10                                                                            & 77                               & 98                               & 49                                  & 184                            & 368                               & 12                             & 12                                  & 3                                & 655                                                                               & 1          & 40         & 20         & 3                   & 4                   & 2.796                                      & 1122                                                         \\
eglese.10                                                                            & 77                               & 98                               & 49                                  & 184                            & 368                               & 12                             & 12                                  & 3                                & 655                                                                               & 1          & 60         & 20         & 3                   & 4                   & 2.776                                      & 1122                                                         \\
eglese.10                                                                            & 77                               & 98                               & 49                                  & 184                            & 368                               & 12                             & 12                                  & 3                                & 655                                                                               & 2          & 5          & 20         & 4                   & 5                   & 4.841                                      & 1222                                                         \\
eglese.10                                                                            & 77                               & 98                               & 49                                  & 184                            & 368                               & 12                             & 12                                  & 3                                & 655                                                                               & 2          & 10         & 20         & 4.7                 & 6                   & 8.858                                      & 1119                                                         \\
eglese.10                                                                            & 77                               & 98                               & 49                                  & 184                            & 368                               & 12                             & 12                                  & 3                                & 655                                                                               & 2          & 20         & 20         & 4.7                 & 6                   & 8.614                                      & 1119                                                         \\
eglese.10                                                                            & 77                               & 98                               & 49                                  & 184                            & 368                               & 12                             & 12                                  & 3                                & 655                                                                               & 2          & 40         & 20         & 4.7                 & 6                   & 10.883                                     & 1119                                                         \\
eglese.10                                                                            & 77                               & 98                               & 49                                  & 184                            & 368                               & 12                             & 12                                  & 3                                & 655                                                                               & 2          & 60         & 20         & 4.7                 & 6                   & 13.291                                     & 1119                                                         \\
eglese.10                                                                            & 77                               & 98                               & 49                                  & 184                            & 368                               & 12                             & 12                                  & 3                                & 655                                                                               & 3          & 5          & 20         & 4                   & 5                   & 4.955                                      & 1222                                                         \\
eglese.10                                                                            & 77                               & 98                               & 49                                  & 184                            & 368                               & 12                             & 12                                  & 3                                & 655                                                                               & 3          & 10         & 20         & 4.7                 & 6                   & 7.572                                      & 1119                                                         \\
eglese.10                                                                            & 77                               & 98                               & 49                                  & 184                            & 368                               & 12                             & 12                                  & 3                                & 655                                                                               & 3          & 20         & 20         & 4.7                 & 6                   & 9.648                                      & 1119                                                         \\
eglese.10                                                                            & 77                               & 98                               & 49                                  & 184                            & 368                               & 12                             & 12                                  & 3                                & 655                                                                               & 3          & 40         & 20         & 4.7                 & 6                   & 16.235                                     & 1119                                                         \\
eglese.10                                                                            & 77                               & 98                               & 49                                  & 184                            & 368                               & 12                             & 12                                  & 3                                & 655                                                                               & 3          & 60         & 20         & 4.7                 & 6                   & 9.246                                      & 1119                                                         \\
eglese.10                                                                            & 77                               & 98                               & 49                                  & 184                            & 368                               & 12                             & 12                                  & 3                                & 655                                                                               & 4          & 5          & 20         & 4                   & 5                   & 4.827                                      & 1222                                                         \\
eglese.10                                                                            & 77                               & 98                               & 49                                  & 184                            & 368                               & 12                             & 12                                  & 3                                & 655                                                                               & 4          & 10         & 20         & 4.7                 & 6                   & 8.505                                      & 1119                                                         \\
eglese.10                                                                            & 77                               & 98                               & 49                                  & 184                            & 368                               & 12                             & 12                                  & 3                                & 655                                                                               & 4          & 20         & 20         & 4.7                 & 6                   & 11.242                                     & 1119                                                         \\
eglese.10                                                                            & 77                               & 98                               & 49                                  & 184                            & 368                               & 12                             & 12                                  & 3                                & 655                                                                               & 4          & 40         & 20         & 4.7                 & 6                   & 10.459                                     & 1119                                                         \\
eglese.10                                                                            & 77                               & 98                               & 49                                  & 184                            & 368                               & 12                             & 12                                  & 3                                & 655                                                                               & 4          & 60         & 20         & 4.7                 & 6                   & 11.165                                     & 1119                                                         \\
eglese.10                                                                            & 77                               & 98                               & 49                                  & 184                            & 368                               & 12                             & 12                                  & 3                                & 655                                                                               & 5          & 5          & 20         & 4                   & 5                   & 5.628                                      & 1222                                                         \\
eglese.10                                                                            & 77                               & 98                               & 49                                  & 184                            & 368                               & 12                             & 12                                  & 3                                & 655                                                                               & 5          & 10         & 20         & 4.7                 & 6                   & 8.859                                      & 1119                                                         \\
eglese.10                                                                            & 77                               & 98                               & 49                                  & 184                            & 368                               & 12                             & 12                                  & 3                                & 655                                                                               & 5          & 20         & 20         & 4.7                 & 6                   & 9.972                                      & 1119                                                         \\
eglese.10                                                                            & 77                               & 98                               & 49                                  & 184                            & 368                               & 12                             & 12                                  & 3                                & 655                                                                               & 5          & 40         & 20         & 4.7                 & 6                   & 10.861                                     & 1119                                                         \\
eglese.10                                                                            & 77                               & 98                               & 49                                  & 184                            & 368                               & 12                             & 12                                  & 3                                & 655                                                                               & 5          & 60         & 20         & 4.7                 & 6                   & 10.088                                     & 1119                                                         \\
eglese.101                                                                           & 140                              & 190                              & 95                                  & 206                            & 412                               & 9                              & 9                                   & 2                                & 2508                                                                              & 1          & 5          & 20         & 2                   & 3                   & 7.058                                      & 3743                                                         \\
eglese.101                                                                           & 140                              & 190                              & 95                                  & 206                            & 412                               & 9                              & 9                                   & 2                                & 2508                                                                              & 1          & 10         & 20         & 2.5                 & 3                   & 13.139                                     & 3166                                                         \\
eglese.101                                                                           & 140                              & 190                              & 95                                  & 206                            & 412                               & 9                              & 9                                   & 2                                & 2508                                                                              & 1          & 20         & 20         & 4                   & 6                   & 19.761                                     & 3090                                                         \\
eglese.101                                                                           & 140                              & 190                              & 95                                  & 206                            & 412                               & 9                              & 9                                   & 2                                & 2508                                                                              & 1          & 40         & 20         & 4                   & 6                   & 19.448                                     & 3090                                                         \\
eglese.101                                                                           & 140                              & 190                              & 95                                  & 206                            & 412                               & 9                              & 9                                   & 2                                & 2508                                                                              & 1          & 60         & 20         & 4                   & 6                   & 20.316                                     & 3090                                                         \\
eglese.101                                                                           & 140                              & 190                              & 95                                  & 206                            & 412                               & 9                              & 9                                   & 2                                & 2508                                                                              & 2          & 5          & 20         & 2                   & 3                   & 10.597                                     & 3743                                                         \\
eglese.101                                                                           & 140                              & 190                              & 95                                  & 206                            & 412                               & 9                              & 9                                   & 2                                & 2508                                                                              & 2          & 10         & 20         & 2.5                 & 3                   & 16.866                                     & 3741                                                         \\
eglese.101                                                                           & 140                              & 190                              & 95                                  & 206                            & 412                               & 9                              & 9                                   & 2                                & 2508                                                                              & 2          & 20         & 20         & 4                   & 6                   & 38.9                                       & 3090                                                         \\
eglese.101                                                                           & 140                              & 190                              & 95                                  & 206                            & 412                               & 9                              & 9                                   & 2                                & 2508                                                                              & 2          & 40         & 20         & 4                   & 6                   & 237.391                                    & 3090                                                         \\
eglese.101                                                                           & 140                              & 190                              & 95                                  & 206                            & 412                               & 9                              & 9                                   & 2                                & 2508                                                                              & 2          & 60         & 20         & 4                   & 6                   & 52.323                                     & 3090                                                         \\
eglese.101                                                                           & 140                              & 190                              & 95                                  & 206                            & 412                               & 9                              & 9                                   & 2                                & 2508                                                                              & 3          & 5          & 20         & 2                   & 3                   & 13.279                                     & 3743                                                         \\
eglese.101                                                                           & 140                              & 190                              & 95                                  & 206                            & 412                               & 9                              & 9                                   & 2                                & 2508                                                                              & 3          & 10         & 20         & 2.5                 & 3                   & 20.437                                     & 3741                                                         \\
eglese.101                                                                           & 140                              & 190                              & 95                                  & 206                            & 412                               & 9                              & 9                                   & 2                                & 2508                                                                              & 3          & 20         & 20         & 4                   & 6                   & 44.674                                     & 3090                                                         \\
eglese.101                                                                           & 140                              & 190                              & 95                                  & 206                            & 412                               & 9                              & 9                                   & 2                                & 2508                                                                              & 3          & 40         & 20         & 3.5                 & 5                   & 62.186                                     & 3090                                                         \\
eglese.101                                                                           & 140                              & 190                              & 95                                  & 206                            & 412                               & 9                              & 9                                   & 2                                & 2508                                                                              & 3          & 60         & 20         & 3.5                 & 5                   & 69.214                                     & 3090                                                         \\
eglese.101                                                                           & 140                              & 190                              & 95                                  & 206                            & 412                               & 9                              & 9                                   & 2                                & 2508                                                                              & 4          & 5          & 20         & 2                   & 3                   & 12.56                                      & 3743                                                         \\
eglese.101                                                                           & 140                              & 190                              & 95                                  & 206                            & 412                               & 9                              & 9                                   & 2                                & 2508                                                                              & 4          & 10         & 20         & 2.5                 & 3                   & 20.971                                     & 3741                                                         \\
eglese.101                                                                           & 140                              & 190                              & 95                                  & 206                            & 412                               & 9                              & 9                                   & 2                                & 2508                                                                              & 4          & 20         & 20         & 4                   & 6                   & 45.319                                     & 3090                                                         \\
eglese.101                                                                           & 140                              & 190                              & 95                                  & 206                            & 412                               & 9                              & 9                                   & 2                                & 2508                                                                              & 4          & 40         & 20         & 3.5                 & 5                   & 62.739                                     & 3090                                                         \\
eglese.101                                                                           & 140                              & 190                              & 95                                  & 206                            & 412                               & 9                              & 9                                   & 2                                & 2508                                                                              & 4          & 60         & 20         & 3.5                 & 5                   & 66.58                                      & 3090                                                         \\
eglese.101                                                                           & 140                              & 190                              & 95                                  & 206                            & 412                               & 9                              & 9                                   & 2                                & 2508                                                                              & 5          & 5          & 20         & 2                   & 3                   & 13.467                                     & 3743                                                         \\
eglese.101                                                                           & 140                              & 190                              & 95                                  & 206                            & 412                               & 9                              & 9                                   & 2                                & 2508                                                                              & 5          & 10         & 20         & 2.5                 & 3                   & 18.476                                     & 3741                                                         \\
eglese.101                                                                           & 140                              & 190                              & 95                                  & 206                            & 412                               & 9                              & 9                                   & 2                                & 2508                                                                              & 5          & 20         & 20         & 4                   & 6                   & 43.834                                     & 3090                                                         \\
eglese.101                                                                           & 140                              & 190                              & 95                                  & 206                            & 412                               & 9                              & 9                                   & 2                                & 2508                                                                              & 5          & 40         & 20         & 3.5                 & 5                   & 59.973                                     & 3090                                                         \\
eglese.101                                                                           & 140                              & 190                              & 95                                  & 206                            & 412                               & 9                              & 9                                   & 2                                & 2508                                                                              & 5          & 60         & 20         & 3.5                 & 5                   & 70.63                                      & 3090    
\\ \hline
\end{tabular}}
\end{table}

\clearpage

\begin{algorithm}[h!]
\fontsize{11}{16.5}\selectfont
\caption{Search}
\label{search_pseudo}
\begin{algorithmic}[1]
\Procedure{Search}{$G, P_K, R_T, \tau_f, r, K, S_K, t$}
    \State Initialize $K_r = \emptyset$
    \ForAll{vehicle $k = 1, \ldots, K$} 
        \If{$S_k =$ True}                                   
            \State $i \leftarrow \Call{tripindex}{G, P_k, R_T, t}$
            \For{$j = i \to$ len($P_k$)}      
                \State $s_d \leftarrow P_k[j][1]$   
                \State $e_d \leftarrow$ end of $P_k[j]$    
                \State $D_{s_d} \leftarrow \min (\Call{distance}{G, s_d, \tau_f[1]}, \Call{distance}{G, s_d, \text{end of }\tau_f})$
                \State $D_{e_d} \leftarrow \min (\Call{distance}{G, e_d, \tau_f[1]}, \Call{distance}{G, e_d, \text{end of }\tau_f})$
                \If{$ D_{s_d} \leq r \; || \; D_{e_d} \leq r$}
                    \State Add vehicle $k$ to $K_r$
                    \State \textbf{break}
                \EndIf
            \EndFor
        \EndIf
    \EndFor
    \State \Return $K_r$
\EndProcedure
\end{algorithmic}
\end{algorithm}

\begin{algorithm}[h!]
\caption{TRIPINDEX}
\label{tripindexpseudo}
\begin{algorithmic}[1]
\Procedure{TRIPINDEX}{$G, P_k, R_T, t$}
    \State $i \leftarrow -1$   
    \State $p \leftarrow 0$     
    \While{$p < t$}
        \State $i \leftarrow i + 1$     
        \If{$i = len(P_k)$}   
            \State $p \leftarrow p + \Call{triptime}{G, P_k[i]}$
        \Else
            \State $p \leftarrow p + \Call{triptime}{G, P_k[i]} + R_T$
        \EndIf        
    \EndWhile
    \State \Return $i$
\EndProcedure
\end{algorithmic}
\end{algorithm}

\clearpage

\begin{algorithm}[t!]
\fontsize{11}{16.5}\selectfont
\caption{Calculate Bid}
\label{calcbid}
\begin{algorithmic}[1]
\Procedure{CalcBid}{$G, D_d, \tau_f, t, e_f, k, P_k, t_m$}
    \State $D_r \leftarrow \{\;\}$  
    \State $i \leftarrow \Call{tripindex}{G, P_k, R_T, t}$ 
    \For{$j = i \to$ len($P_k$)}      
        \State  $D_r[j] \leftarrow$ end of trip $P_k[j]$       
    \EndFor

    \State $P_b \leftarrow \emptyset$                   
    \State $P_t \leftarrow \infty$                      
    \ForAll{ item $(j, d_r) \in D_r$}
        \State $P_{ck} \leftarrow \Call{copy}{P_k}$   
        \State $P \leftarrow \Call{inserttrip}{G, D_d, P_{ck}, j, d_r, \tau_f, R_T}$     
   
        \If{$P_t > \Call{routetime}{G, P, R_T}$}
            \State $P_t \leftarrow \Call{routetime}{G, P, R_T}$
            \State $P_b \leftarrow P$
        \EndIf
    \EndFor
    
    \State $bid \leftarrow P_t - t_m$

    \State \Return $bid, P_b$
\EndProcedure
\end{algorithmic}
\end{algorithm}

\begin{algorithm}[b!]
\fontsize{11}{16.5}\selectfont
\caption{Insert Trip}
\label{inserttrip}
\begin{algorithmic}[1]
\Procedure{inserttrip}{$G, D_d, P_{ck}, j, d_r, \tau_f, R_T$}
    \State $P \leftarrow \emptyset$  
    \If{$j \; != len(P_{ck})$}       
        \State $P_i \leftarrow D_d[d_r, \tau_f[1]] + \tau_f + D_d[\text{end of } \tau_f, d_r]$
    \Else
        \State $s_d \leftarrow \Call{routetime}{G, D_d[d_r, \tau_f[1]], R_T}$
        \State $e_d \leftarrow \Call{routetime}{G, D_d[d_r, \text{end of } \tau_f, R_T}$
        \If{$s_d \leq e_d$}
            \State $P_i \leftarrow D_d[d_r, \tau_f[1]] + \tau_f$
        \Else
            \State $P_i \leftarrow D_d[d_r, \text{end of } \tau_f] + \text{reverse of trip }\tau_f$
        \EndIf
    \EndIf
    \State $P \leftarrow $ Insert $P_i$ in $P_{cK}$ at trip index $j$
    \State \Return $P$ 
\EndProcedure
\end{algorithmic}
\end{algorithm}

\clearpage

\section{Derivation of Theoretical Performance Bound}
\label{sec:supp_theoretical_bound}

This section provides the detailed derivation and physical interpretation of the performance bound presented in the main manuscript. We define $\beta_{CA}$ as the mission time of the centralized auction and $\beta_{OPT_f}$ as the mission time of an offline optimal solver with perfect foreknowledge.

The performance gap arises from the rescheduling cost, which captures the additional travel and recharge time incurred when active vehicles reposition to service required edges abandoned due to failures. The mission time produced by the centralized auction can be bounded additively as:
\begin{equation}
\beta_{CA} \le \beta_{OPT_f} + \sum_{j \in \mathcal{J}_{fail}} \Delta_j
\end{equation}
where $\mathcal{J}_{fail}$ denotes the set of trips that must be reassigned following vehicle failures, and $\Delta_j$ denotes the penalty incurred when reassigning trip $j$.

\subsection{Operational Assumptions for Bound Derivation}
To determine the upper bound of $\Delta_j$, we rely on the following operational assumptions:
\begin{enumerate}
    \item The depot nodes form a fully connected undirected graph such that the travel time between any two depots is within the vehicle battery capacity $C$.
    \item All required edges remain traversable by the surviving vehicles despite the occurrence of failures.
\end{enumerate}

Under these assumptions, the worst-case penalty for reassigning a single trip consists of repositioning to the start of the failed task and subsequently returning to a depot. This penalty is bounded by:
\begin{equation}
\Delta_j \le 2(C + R_T)
\end{equation}
which accounts for one full capacity traversal and one recharge to reach the failure location and, in the worst case, an additional traversal and recharge to resume the route or return to the depot.

\begin{figure}[h!]
\centering
\includegraphics[width=0.7\textwidth]{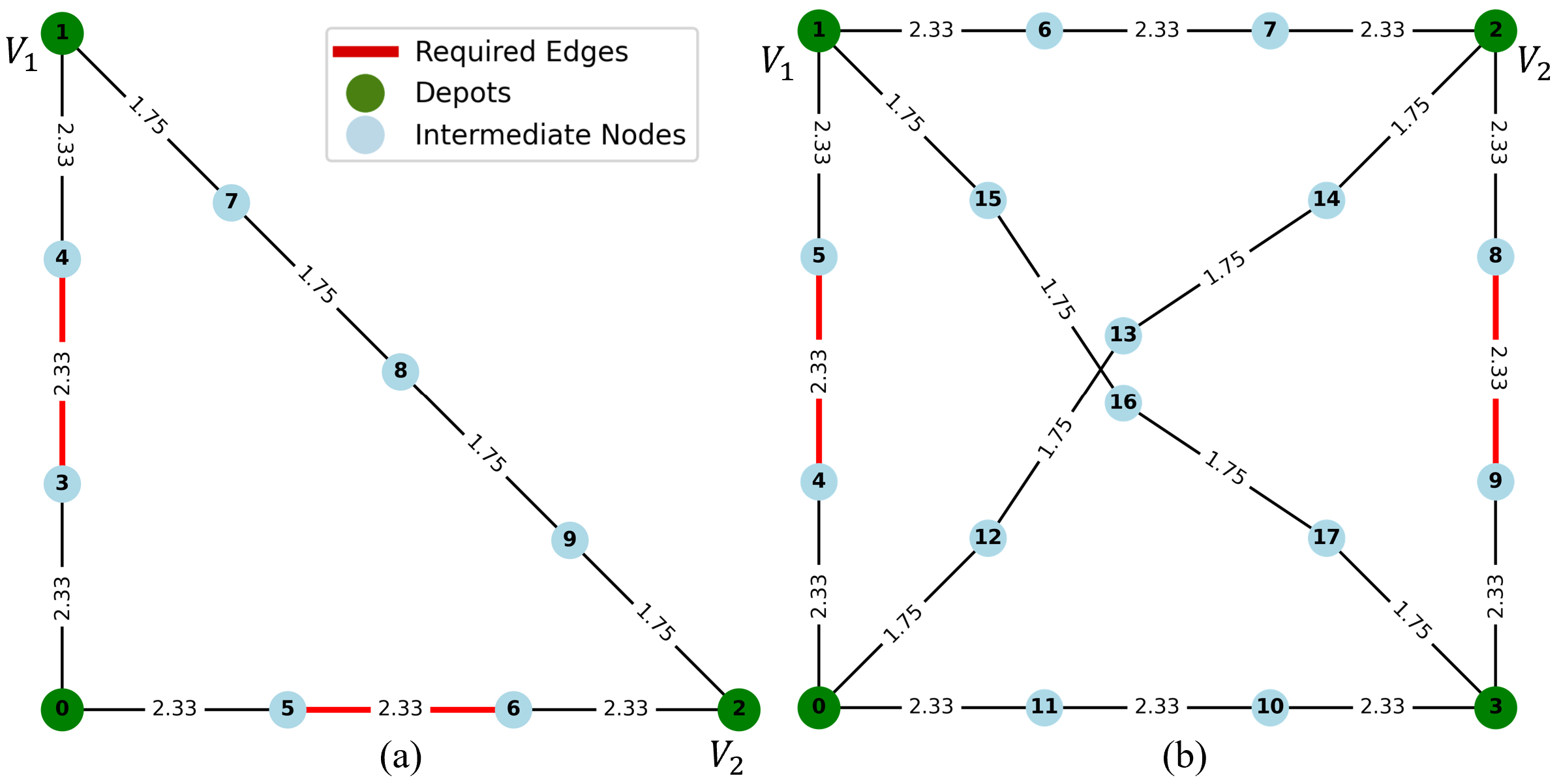}
\caption{Visualizing the rescheduling penalty in two scenarios. In Case A, the proximity of the failing vehicle's task to the survivor's route minimizes overhead. In Case B, the surviving vehicle must incur significant travel and recharge costs to reach the remote failure site, maximizing the additive penalty.}
\label{fig:resheduling_penalty_supp}
\end{figure}

\subsection{Physical Interpretation of the Bound}
To provide a physical interpretation of this bound, Figure \ref{fig:resheduling_penalty_supp} visualizes the components of the reaction penalty $\Delta$ using a scenario with capacity $C=7$ and recharge time $R_T=14$. 

\textbf{Worst-Case (Case B):} Vehicle $V_2$ fails at $t=4$ while servicing edge $(8,9)$. The surviving vehicle $V_1$ is located at the maximum distance from the failure, finishing its own task at depot 0 at $t=7$ with a depleted battery. To service the failed task, $V_1$ must incur a distinct \textit{deadhead cost}: it recharges ($14$ units) and traverses the maximum allowable distance defined by the battery capacity (path $0 \to 3$, taking $7$ units) solely to reach the failure site. This creates an unavoidable delay of $C + R_T = 21$ units. The factor of 2 in the theoretical bound accounts for the potential necessity to perform a similar traversal to return to a depot after the repair. 

\textbf{Best-Case (Case A):} Vehicle $V_1$ finishes its route (path $1 \to 0$) adjacent to the start of the failed trip. In this case, the penalty vanishes ($\Delta \to 0$), and the centralized auction achieves a mission time identical to the offline optimal solver ($\beta_{CA} \approx \beta_{OPT_f}$).

\section{Detailed Computational Complexity Analysis}
\label{sec:comp_complex}

The computational complexity of the proposed reactive framework is derived by aggregating the operational costs of the centralized auction followed by the peer auction refinement. The centralized auction, detailed in Manuscript Algorithm 2, relies on the offline precomputation of depot-to-depot routes (Manuscript Algorithm 1, line 4), which requires $O(|N_d|^2 (|E| + |V|) \log |V| + |N_d|^3)$ time using Dijkstra's and Floyd-Warshall algorithms. This precomputation enables constant-time $O(1)$ lookups during the failure response. Consequently, the real-time complexity is driven by the \textsc{Search} (Manuscript Algorithm 2, line 8) and \textsc{CalcBid} (Manuscript Algorithm 2, line 12) procedures. For a fleet of $K$ vehicles and $N_d$ depots, the search process iterates through vehicles with a complexity of $O(K \cdot D/\Delta r)$, where $D$ is the graph diameter and $\Delta r$ is the search increment. The bid calculation evaluates insertion points at depot visits, scaling linearly with the number of depots as $O(K \cdot |N_d|)$. Therefore, the total computational complexity for the first stage, which provides the initial feasible solution, is $O(K (|N_d| + D/\Delta r))$. This linear scaling with fleet size ensures that the baseline schedule is generated almost instantaneously, satisfying the immediate stability requirement of dynamic rescheduling.

The second stage, the peer auction (Manuscript Algorithm 3), refines this baseline through $R$ iterative rounds. In each round, the algorithm sorts vehicles to identify potential donors and receivers, incurring a sorting cost of $O(K \log K)$. The complexity then focuses on the generation of trip combinations and the evaluation of transactions. The \textsc{GenerateTripCombinations} procedure (Manuscript Algorithm 4) creates contiguous sub-segments of trips within a window size $W$. For a vehicle with $M$ future trips, the exact number of combinations is a linear summation of segment lengths (e.g., $2M-1$ for $W=2$), which asymptotically scales as $O(M \cdot W)$.

The subsequent \textsc{BuildTransactions} procedure (Manuscript Algorithm 5) explores the Cartesian product of these blocks, necessitating a pairwise comparison between every donor combination and every receiver combination to identify valid swaps. Without constraints, this search space scales quadratically as $O(M^2 W^2)$, creating a potential bottleneck when the number of future trips $M$ is large. To prevent this polynomial growth from overwhelming real-time responsiveness, we enforce a strict computational budget $L$. Consequently, the algorithm evaluates $\min(O(M^2 W^2), L)$ transactions. This ensures that only a constant number of the most promising moves are fully constructed and evaluated by the magnetic field router (Manuscript Algorithm 6).

The magnetic field router serves as the local repair route constructor within the peer auction loop. To reconstruct a route with $|E_{rem}|$ required edges, the router iterates through the remaining edge set and evaluates the attractive forces from adjacent neighbors defined by the maximum degree of the graph, $\text{deg}(G)$. This results in a complexity of $O(|E_{rem}| \cdot \text{deg}(G))$ for a single route reconstruction. Since the peer auction performs this reconstruction for the pair of vehicles involved in each of the limited transactions over $R$ rounds, the operational complexity of the refinement stage is dominated by $O(R \cdot L \cdot |E_{rem}| \cdot \text{deg}(G))$.

Combining both stages, the total computational complexity of the reactive framework is $O(K (|N_d| + D/\Delta r) + R (M^2 W^2 + L \cdot |E_{rem}| \cdot \text{deg}(G)))$. This analytical result highlights the significance of calibrating the window size $W$ and the computational budget $L$. By fixing these parameters, we effectively bound the search space exploration and the expensive routing calls, transforming the re-optimization problem into a polynomial-time heuristic operation. Consequently, the framework guarantees that the computational effort remains predictable, allowing the system to scale to large instances while delivering high-quality solutions within the tight time constraints of a live mission.

\section{Problem Formulation}
\label{sec:prob-for}
This section presentes the MILP formulation of the studied MD-RPP-RRV with vehicle failures. This presented formulation builds upon our previous work \citep{md-rpp-rrv}, extending it to account for vehicle failures to generate offline optimal solutions. The assumption here is that all vehicle failure times are known beforehand. For the manuscript to be self-contained, the constraints and formulation are briefly described, focusing mainly on the failure constraints. For a detailed explanation of the MILP formulation (specifically constraints \ref{constraint_1} - \ref{constraint_12}) for MD-RPP-RRV, readers are referred to \citep{md-rpp-rrv}.

The MD-RPP-RRV is modeled on an undirected weighted connected graph $G = (N, E, T)$, where $N$ represents the set of nodes, $E$ denotes the set of edges connecting these nodes, and $T$ contains the edge weights, which is the time taken by a vehicle to traverse the edge $(i,j)$. Each edge $(i,j) \in E$ corresponds to an edge with length $l(i,j)$, traversed by vehicles at a constant speed $S$. The time $t(i,j)$ required to traverse an edge is calculated as $l(i,j)/S$. The problem involves a subset of required edges $E_u \subseteq E$ that must be visited and a set of depots $N_d \subseteq N$ where vehicles can start, stop, or recharge.

The problem considers $K$ vehicles, each with a maximum operational time $C$ after charging and a recharge time $R_T$. The maximum number of trips a vehicle can make is denoted by $F$. To formulate the MD-RPP-RRV as a MILP model, three sets of binary decision variables are introduced: $x(k,f,i,j)$ indicates if vehicle $k$ traverses edge $(i,j)$ during its $f$-th trip, $y(k,f,d)$ denotes if vehicle $k$ ends its $f$-th trip at depot $d$, and $z(k,f)$ signifies if vehicle $k$ uses its $f$-th trip. The objective function $\beta$ represents the maximum total time needed by any vehicle to complete all its trips and recharge between trips.

\begin{alignat}{3}
&\min \; \beta  && \nonumber \\
\startparent
\label{constraint_1}
&\text{subject to:} \sum\limits_{(B(k),j) \in E} x(k, 1, B(k), j) = z(k, 1),  k = 1,...,K\\
\label{constraint_2}
&z(k, f) - z(k, f+1) \geq 0, \; k = 1,...,K, f = 1,...,F-1\\
\label{constraint_3}
&\sum_{\substack{(i,d) \in E, \\ d \in N_d}} x(k, f, i, d) = y(k, f, d), \; k = 1,...,K,  f = 1,...,F\\
\label{constraint_4}
\begin{split}
&y(k, f-1, d) \geq \sum_{(d,j) \in E} x(k, f, d, j), \quad k = 1,...,K, f = 2,...,F, d \in N_d \\
\end{split}
\\
\label{constraint_5}
& z(k,f) - \sum_{d \in N_d} y(k,f,d) = 0, \quad k = 1,...,K,  f = 1,...,F\\
\label{constraint_6}
\begin{split}
&\sum_{f=1}^{F} \sum_{(i.j) \in E} x(k, f, i, j) t(i,j) + (\sum_{f=1}^{F} z(k,f) - 1) \times R_T \leq \beta, \quad k = 1,...,K\\
\end{split}
\\
\label{constraint_7}
&\sum_{(i,j) \in E} x(k, f, i, j) t(i,j) \leq C, \; k = 1,...,K,  f = 1,...,F\\
\label{constraint_8}
\begin{split}
&\sum_{\substack{(i,j) \in E, \\ i \in N_d}} x(k, f, i, j) - \sum_{\substack{(i,j) \in E, \\ j \in N_d}} x(k, f, i, j) = 0, \quad k = 1,...,K,  f = 1,...,F\\    
\end{split}
\\
\label{constraint_9}
\begin{split}
&\sum_{j \in N} x(k, f, i, j) - \sum_{j \in N} x(k, f, j, i) = 0, \quad k = 1,...,K, f = 1,...,F, i \in N/\{N_d\} \\   
\end{split}
\\
\label{constraint_10}
&\sum_{k=1}^{K} \sum_{f=1}^{F} x(k,f,i,j) + \sum_{k=1}^{K} \sum_{f=1}^{F} x(k,f,j,i) \geq 1, \; \forall (i,j) \in E_u\\
\label{constraint_11}
&\sum_{(i,j) \in E} x(k,f,i,j) \leq z(k,f) \times M, \quad k = 1,...,K,  f = 1,...,F \\
\label{constraint_12}
\begin{split}
\\
&\sum_{(i,j) \in \delta(S)} x(k,f,i,j) \geq 2 \times x(k,f,p,q), \quad k = 1,...,K, f = 1,...,F, \forall S \subseteq N/\{N_d\}, (p,q) \in E(S)\\
\end{split}\\
\label{constraint_13}
\begin{split}
&\sum_{f=1}^{F} \sum_{(i.j) \in E} x(k, f, i, j) t(i,j) + (\sum_{f=1}^{F} z(k,f) - 1) \times R_T \leq f_k, \quad \forall k \in F\\
\end{split}\\
&x(k,f,i,j) \in [0,1],\quad k = 1,...,K,  f = 1,...,F, \forall (i,j) \in E\\
&y(k,f,d) \in [0,1], \quad k = 1,...,K,  f = 1,...,F, d \in N_d\\
&z(k,f) \in [0,1], \quad k = 1,...,K,  f = 1,...,F\\
&\beta \in R^{+}, \: M >> |E|
\end{alignat}

The MILP formulation includes several constraints to ensure proper routing and adherence to problem specifications. Constraints (\ref{constraint_1}-\ref{constraint_5}) manage trip initiation and termination at depots. Constraints (\ref{constraint_6}-\ref{constraint_7}) enforce maximum trip time and battery capacity limits. Constraints (\ref{constraint_8}-\ref{constraint_10}) ensure flow conservation and required edges traversal. Constraints (\ref{constraint_11}-\ref{constraint_12}) eliminate unused trips and subtours. Constraint \ref{constraint_13} forces all failure vehicles ($F \subset \{1,.., K\}$) to operate only below their respective failure times ($f_{k}, \; \forall k \in F$). This will ensure none of the failure vehicles is utilized to traverse required edges past their respective failure times. This comprehensive set of constraints allows for generating offline optimal solutions for the MD-RPP-RRV for vehicle failures.

\end{document}